\documentclass{article}

\usepackage[nonatbib, dandb, final]{neurips_2025}

\usepackage[utf8]{inputenc} 
\usepackage[T1]{fontenc}    
\usepackage{hyperref}       
\usepackage{url}            
\usepackage{booktabs}       
\usepackage{amsfonts}       
\usepackage{nicefrac}       
\usepackage{microtype}      
\usepackage{xcolor}         
\usepackage{bm}
\usepackage{amsmath}
\usepackage{amssymb}
\usepackage{arydshln}
\usepackage{enumerate}
\usepackage{enumitem}
\usepackage{pifont}
\usepackage{array}
\usepackage{soul}
\usepackage{mathtools}
\usepackage{amsthm}
\usepackage{tikz}
\usepackage{hyperref}
\usepackage{booktabs}
\usepackage{wrapfig}
\usepackage{siunitx}
\usepackage{wrapfig}
\usepackage[most]{tcolorbox}
\usepackage{url}
\usepackage{amsfonts,amssymb,amsmath,amsthm,amsopn,mathrsfs,mathtools,wasysym}	
\usepackage{hhline,booktabs,colortbl,multirow,tabularx,diagbox,threeparttable} 
\usepackage{xcolor}
\usepackage{tabularx}
\usepackage{prettyref}
\usepackage{tcolorbox}
\usepackage{xcolor}
\usepackage{titletoc}
\usepackage{makecell} 
\usepackage{longtable}
\usepackage{rotating}
\usepackage{pdflscape}
\usepackage{collcell}

\usepackage[numbers,compress]{natbib}

\usepackage{fontawesome5}

\hypersetup{
    colorlinks=true,
    linkcolor=reference,
    citecolor=reference,
    filecolor=reference,
    urlcolor=reference,
    pdfborder={0 0 0}
}

\makeatletter
\tcbset{
    myhbox/.style 2 args={%
        enhanced, 
        breakable,
        colback=white,
        colframe=black,
        attach boxed title to top left={yshift*=-\tcboxedtitleheight}, 
        title={#2},
        boxed title size=title,
        boxed title style={%
            sharp corners, 
            rounded corners=northwest, 
            colback=tcbcolframe, 
            boxrule=0pt,
        },
        underlay boxed title={%
            \path[fill=tcbcolframe] (title.south west)--(title.south east) 
                to[out=0, in=180] ([xshift=5mm]title.east)--
                (title.center-|frame.east)
                [rounded corners=\kvtcb@arc] |- 
                (frame.north) -| cycle; 
        },
        #1
    }
}   
\makeatother

\newtcolorbox{myhbox}[2][]{%
    myhbox={#1}{#2}
}

\newtheorem{definition}{Definition}[section] 
\newtheorem{example}{Example}

\definecolor{reference}{RGB}{4, 20, 110}
\definecolor{amaranth}{rgb}{0.9, 0.17, 0.31}
\definecolor{brightmaroon}{rgb}{0.76, 0.13, 0.28}
\definecolor{specialgreen}{rgb}{0.5176, 0.8275, 0.2392}
\definecolor{specialblue}{rgb}{0.3412, 0.6745, 0.9412}
\definecolor{deepblue}{rgb}{0.2118, 0.2784, 0.6039}
\definecolor{rowcolor}{RGB}{96,54,148}
\definecolor{lightpurple}{RGB}{242, 239, 246}
\definecolor{beaublue}{rgb}{0.74, 0.83, 0.9}
\definecolor{tabledeep}{RGB}{216,214,196}
\definecolor{tablelight}{RGB}{236,234,224}
\definecolor{lightgreen}{rgb}{0.8, 1.0, 0.8}
\definecolor{lightred}{rgb}{1.0, 0.8, 0.8}

\newrefformat{fig}{Fig.~\ref{#1}}
\newrefformat{tab}{Table~\ref{#1}}
\newrefformat{sec}{Section~\ref{#1}}
\newrefformat{app}{Appendix~\ref{#1}}
\newrefformat{alg}{Algorithm~\ref{#1}}

\newcommand{\colorFDC}[1]{%
  \ifdim#1pt<-0.3pt
    \cellcolor{lightgreen}#1
  \else
    \ifdim#1pt>0.0pt
      \cellcolor{lightred}#1
    \else
      #1
    \fi
  \fi
}

\newcommand{\colorRho}[1]{%
  \ifdim#1pt>0.4pt
    \cellcolor{lightgreen}#1
  \else
    \ifdim#1pt<0.0pt
      \cellcolor{lightred}#1
    \else
      #1
    \fi
  \fi
}

\newcommand{\colorNFC}[1]{%
  \ifdim#1pt>0.8pt
    \cellcolor{lightgreen}#1
  \else
    \ifdim#1pt<0.4pt
      \cellcolor{lightred}#1
    \else
      #1
    \fi
  \fi
}

\newtcolorbox{quotebox}{
    colback=lightpurple,
    colframe=black!75,
    boxrule=0pt,
    top=5pt,
    bottom=5pt,
    left=8pt,
    right=8pt,
    arc=8pt,
    boxsep=0pt,
    toptitle=2pt,
    bottomtitle=2pt,
    fonttitle=\bfseries,
}

\newcommand{\pref}{\prettyref}

\title{Augmenting Biological Fitness Prediction Benchmarks with Landscapes Features from GraphFLA}

\author{Mingyu Huang$^{1}$, Shasha Zhou$^{2}$, and Ke Li$^{2}$\\
$^1$School of Computer Science and Engineering,\\University of Electronic Science and Technology of China\\
$^2$Department of Computer Science, University of Exeter\\
Email: m.huang.gla@outlook.com; \{sz484, k.li\}@exeter.ac.uk
}

\begin{document}

\maketitle

\begin{abstract}
Machine learning models increasingly map biological sequence-fitness landscapes to predict mutational effects. Effective evaluation of these models requires benchmarks curated from empirical data. Despite their impressive scales, existing benchmarks lack topographical information regarding the underlying fitness landscapes, which hampers interpretation and comparison of model performance beyond averaged scores. Here, we introduce \texttt{GraphFLA}, a Python framework that constructs and analyzes fitness landscapes from mutagensis data in diverse modalities (e.g., DNA, RNA, protein, and beyond) with up to millions of mutants. \texttt{GraphFLA} calculates 20 biologically relevant features that characterize 4 fundamental aspects of landscape topography. By applying \texttt{GraphFLA} to over 5,300 landscapes from ProteinGym, RNAGym, and CIS-BP, we demonstrate its utility in interpreting and comparing the performance of dozens of fitness prediction models, highlighting factors influencing model accuracy and respective advantages of different models. In addition, we release 155 combinatorially complete empirical fitness landscapes, encompassing over 2.2 million sequences across various modalities. All the codes and datasets are available at \url{https://github.com/COLA-Laboratory/GraphFLA}.
\end{abstract}

\section{Introduction}
\label{sec:introduction}

The fitness landscape is a nearly century-old foundational concept rooted in evolutionary biology~\cite{Wright32} with profound implications on the understanding of biological principles in all 3 modalities of the central dogma (DNA, RNA, protein)---from drug resistance~\cite{RizviHSMKHLYWWHMMRMIIBGZMSCWSC15,ChenFT22}, enzyme activity~\cite{PapkouRM23,WeinreichDDH06,LunzerMFD05,JohnstonAWLPYA24,JudgeSHPBPP24}, protein stability and expression~\cite{ZhengGW20,OgdenKSC19,SarkisyanEtAl16}, RNA folding and function~\cite{LiQMZ16,GreenburyLA22,PuchtaCCTSK16,JimenezXCTC13,PittF10}, to transcription factor binding~\cite{PayneW14,AguilarRodriguezPW17}. Efficiently and accurately mapping these fitness landscape surfaces is critical to enable various downstream tasks~\cite{NotinRGGSM24,HayesRASNOZVTWDWBSGDMTKMKBNHSC25,JiangYDBSVKKCHNGGA24,VaishnavBMYFATLCR22,YuYSYZYDLK24}, and has been recently advanced by machine learning (ML) models that can capture complex and high-dimensional patterns of the sequence-fitness map~\cite{TanWWZH24,LiTMZYZOZTH24,GrothKOSB24,HollmannMPKKHSH25,ChenHSTWYZYHXSFKL22,WangBLLLMKX24,BrixiDKPBCCKLM25}. 

A critical step in developing these models is their proper performance evaluation to understand limitations and enable comparisons with existing methods. For this purpose, large-scale benchmarks have been established across different modalities. For example, ProteinGym~\cite{NotinKRNPSRSOWF23} offers more than $250$ tasks curated from deep mutational scanning (DMS) assays for proteins, while RNAGym~\cite{AroraMACKQSWGGMXN25} incorporates over $30$ standardized RNA DMS assays. Considering their impressive scales and the famous ``no free lunch'' theorem~\cite{DolpertM97}, it is often unrealistic to expect one single model to dominate on \textit{all} tasks. For instance, although the VenusREM model~\cite{TanWWZH24} yields the highest \textit{average} score across all $217$ DMS substitution tasks in ProteinGym, it leads in only $14$ ($6.5\%$) individual tasks. Meanwhile, $44$ out of the evaluated $89$ models leads in \textit{at least one} task. This reality prompts critical questions: \textbf{Q1:} \textit{``Why did one model perform well on one set of tasks but poorly on another?''}, \textbf{Q2:} \textit{``why did one model outperform baseline on one task, but not on the other?''} 

Answering these questions necessitates informative features that characterize each task. Unfortunately, existing benchmarks typically provide only basic labels (e.g., taxon) or statistics (e.g., sequence length), which are insufficient to fully elucidate the 2 questions above. Consequently, users often rely on average scores for decision-making and comparison, which can lead to biased conclusions. 

For decades, evolutionary biologists have developed various features to quantitatively characterize topographical aspects of fitness landscapes, including ruggedness~\cite{PapkouRM23,WestmannGW24,VanCleveW15,AguilarRodriguezPW17}, navigability~\cite{AguilarRodriguezPW17,PapkouRM23,WestmannGW24,PayneW14,PoelwijkKWT07}, epistasis~\cite{Bank22,ChouCDSM11,BakerleeNSRD22,OtwinowskiMP18,KhanDSLC11,DomingoDL18}, and neutrality~\cite{LauringFA13,DraghiPWP10,FelixB15,VaishnavBMYFATLCR22}. These features have been extensively applied to unveil fundamental principles governing evolutionary dynamics~\cite{RizviHSMKHLYWWHMMRMIIBGZMSCWSC15,ChenFT22,PapkouRM23,WeinreichDDH06,LunzerMFD05,JohnstonAWLPYA24,JudgeSHPBPP24,ZhengGW20,OgdenKSC19,SarkisyanEtAl16,LiQMZ16,GreenburyLA22,PuchtaCCTSK16,JimenezXCTC13,PittF10,PayneW14,AguilarRodriguezPW17}. As biological sequence models essentially aim to learn these landscape surfaces, we hypothesize that these same features can explain why models perform differently across tasks and address the previous 2 questions.

\begin{figure*}[t!]
    \centering
    \includegraphics[width=\linewidth]{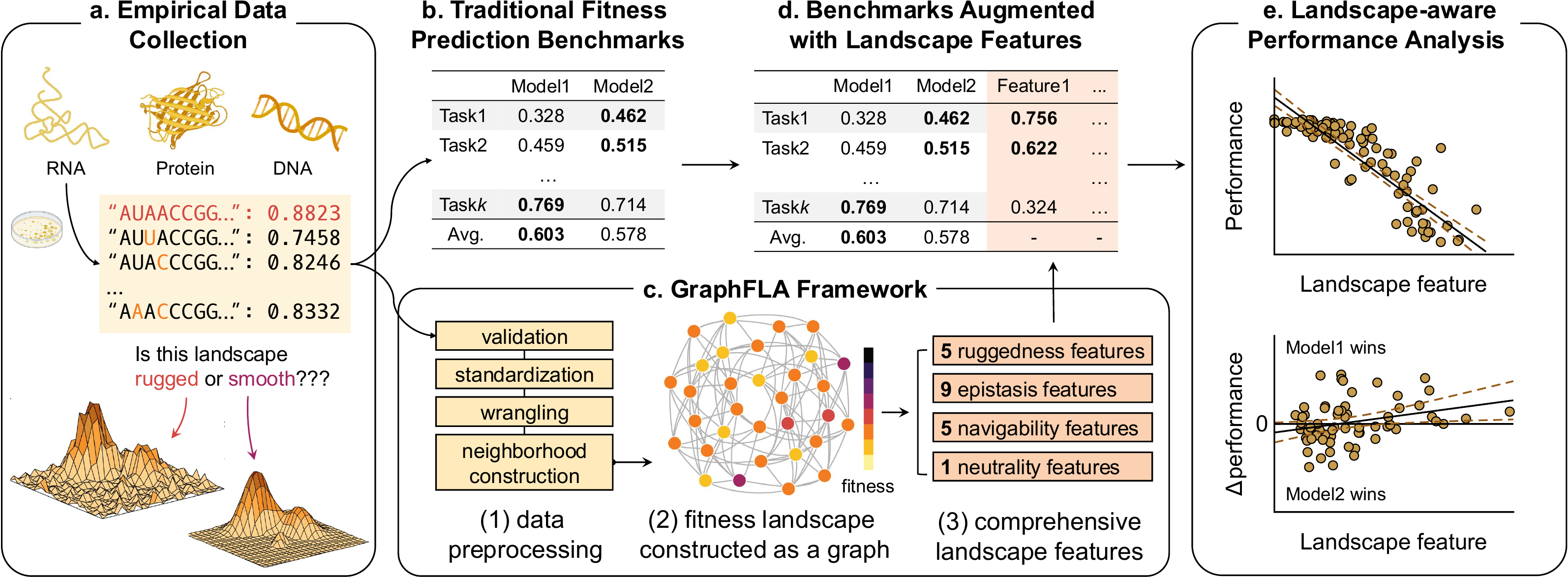}
    \caption{\small \textbf{Overview of how GraphFLA contributes to the performance benchmarking of fitness prediction models.} Existing biological fitness prediction benchmarks \textbf{(b)} are often curated from empirical fitness landscape datasets without interrogating landscape topography \textbf{(a)}. \texttt{GraphFLA} constructs these landscapes and offers a comprehensive suite of features characterizing their topography \textbf{(c)}. Such landscape features can then augment existing benchmarks \textbf{(d)} and thus assist performance interpretation \textbf{(e, upper)} and comparison \textbf{(e, lower)}.}
    \label{fig:intro}
\end{figure*}

Yet, despite decades of study, landscape analysis remains a highly specialized biological field. As a result, standard open-source implementations for calculating many landscape features are rarely available. Also, existing research often targets specific landscape aspects, leaving no consensus on a comprehensive feature set. In addition, empirical landscapes span diverse biological modalities and scales, further complicating the development of unified analysis tools. The rapid growth in empirical data also
demands highly scalable methods. Consequently, researchers currently lack accessible, broadly applicable tools for characterizing fitness landscape features in common benchmarking tasks.

To address this lack of analysis tooling, we present \texttt{GraphFLA}, a versatile, comprehensive, scalable and end-to-end Python framework for streamlining fitness landscape analysis. \texttt{GraphFLA} constructs fitness landscapes from biological sequence-fitness data in diverse modalities (including, but not limited to DNA, RNA, and protein)  and is heavily optimized to scale to datasets with even millions of mutants. It is interoperable with established fitness prediction benchmarks by using an API and data format similar to that used for model training, and is essentially applicable to empirical data in other databases and in the literature. Once a landscape is constructed, \texttt{GraphFLA} offers a rich suite of $20$ features compiled from thousands of papers characterizing $4$ fundamental aspects of landscape topography: ruggedness, navigability, epistasis, and neutrality, which can serve as biologically meaningful meta-features for each benchmark task to better interpret model performance. 

We extensively compared \texttt{GraphFLA}'s scalability to existing tools and validated its reliability via a large-scale replication study using 155 \textit{combinatorially complete} empirical landscapes collected from 61 works (\pref{sec:versatility}), which are released as part of \texttt{GraphFLA}. We then demonstrated \texttt{GraphFLA}'s robustness to data missing, biased sampling, as well as noise in~\pref{sec:robustness} with synthetic landscapes. To further demonstrate its versatility, we applied \texttt{GraphFLA} to analyze 5,000+ empirical landscapes from ProteinGym, RNAGym, and the CIS-BP database~\cite{Weirauch2014}. By employing landscape features from \texttt{GraphFLA} to interpret the performance of dozens of established models on these landscapes, we illustrate that: $\blacktriangleright$ Model performance strongly depends on landscape topography; Landscapes that are more rugged, epistatic, neutral, while less navigable, are harder for models to predict accurately (\textbf{Q1}; \pref{sec:influence}); $\blacktriangleright$ Different models, even with similar overall performance, can excel at different types of landscapes; Performance gaps between them can change with specific landscape characteristics (\textbf{Q2}; \pref{sec:comparison}). Finally, we showcase the wider utility of \texttt{GraphFLA} by applying it to analyze results of ML-guided directed evolution (MLDE) and phenotype landscapes in~\pref{sec:broader_app}.




\section{Background and Related Work}
\label{sec:related}

\textbf{Fitness landscapes.} In his pioneering work in 1932, \citeauthor{Wright32} first described the concept of a fitness landscape by analogy to a physical landscape, where each spatial location represents a genotype, and the elevation indicates its fitness. Though this landscape metaphor is initially used to describe the genotype-fitness map, its influence quickly extended to other biological modalities and scales, e.g., molecules like RNA~\cite{LiQMZ16,GreenburyLA22,PuchtaCCTSK16,JimenezXCTC13,PittF10}, proteins~\cite{PodgornaiaL15,OgdenKSC19,SarkisyanEtAl16}, genes~\cite{FriedlanderPBT17,YangS22}, and even communities~\cite{KeheKOATSGKGFB19,DiazColungaCSRAS24,SkwaraGYDRRTKS23,Diaz-ColungaSVBS24,Sanchez-GorostiagaBOPS19}. 

\textbf{Landscape topography.} Since adaptation can be viewed as navigating fitness landscapes towards their highest peaks, their topography is essential for understanding the course of evolution. The most intuitive and widely studied aspect is \textbf{ruggedness}~\cite{PapkouRM23,WestmannGW24,VanCleveW15,AguilarRodriguezPW17}, often characterized by the presence of multiple local optima (peaks). A necessary condition for landscape ruggedness is \textbf{epistasis}~\cite{ChouCDSM11,BakerleeNSRD22,OtwinowskiMP18,KhanDSLC11,DomingoDL18}, which occurs when one or more mutations interact. In contrast, a purely additive landscape, where mutational effects are independent, would be fairly smooth with a single global optimum. Ruggedness along with pervasive epistasis can pose a fundamental challenge to an evolving population's ability to find the highest peak, thus reducing the landscape's \textbf{navigability}~\cite{AguilarRodriguezPW17,PapkouRM23,WestmannGW24,PayneW14,PoelwijkKWT07}, another important topography aspect. Finally, many studies also interrogate \textbf{neutrality}~\cite{LauringFA13,DraghiPWP10,FelixB15,VaishnavBMYFATLCR22}, which describes the presence of ``plateaus'' consisting of genotypes sharing the same fitness. 

\textbf{Software packages for landscape analysis.} The only biological landscape analysis package known to us, \texttt{MAGALLEN}~\cite{BrouilletAFA15}, offers several quantitative metrics but is limited in scope. In contrast, \texttt{GraphFLA} offers a holistic suite of $20$ features covering all $4$ fundamental aspects above. Also, while \texttt{MAGALLEN} is written in \texttt{C}, it can only rapidly handle landscapes at the scale of $10^5$ variants. \texttt{GraphFLA}, however, easily scales to landscapes of $10^7$. Furthermore, \texttt{MAGALLEN}'s pure-\texttt{C} implementation also hinders interoperability with modern ML ecosystems, unlike \texttt{GraphFLA}'s native Python API.

\textbf{Empirical fitness landscapes.} While early studies of landscape topography often relied on theoretical models (e.g., the NK model~\cite{KauffmanL87}), advancements in experimental methodologies have enabled the empirical assessment of increasingly large fitness landscapes~\cite{NotinKRNPSRSOWF23,RaoBTDCCAS19,DallagoMJWBGMY21}. These empirical landscapes are usually constructed by either $\blacktriangleright$ randomly sampling a vast number of single- or multi-mutants for a wild-type (WT) sequence (e.g., \cite{OgdenKSC19,SarkisyanEtAl16,LiQMZ16}), or $\blacktriangleright$ systematically assaying all possible sequences in a predefined space (e.g., \cite{PapkouRM23,JohnstonAWLPYA24,DomingoDL18,WuDOLS16}). The first approach probes a fairly large area of the sequence space, but the resulting landscape is of narrow depth by containing only immediate neighbors of the WT. In contrast, the second approach generates \textit{combinatorially complete} landscapes that allow exact analysis of topography and enable testing of model predictions on combined effects of mutations.

\textbf{Fitness prediction benchmarks.} Apart from driving biological insights, empirical landscape data also give rise to the wealth of benchmarking tasks for fitness prediction in different modalities~\cite{NotinKRNPSRSOWF23,AroraMACKQSWGGMXN25,DallagoMJWBGMY21,HopfIPS3SM17,RiesselmanIM18,SumiHS24,NguyenPDKLLTTKBSNLALERBHH24,RenCQJCSYMSYYOYL24}. While earlier benchmarks like FLIP~\cite{DallagoMJWBGMY21}, \cite{SumiHS24} and \cite{NguyenPDKLLTTKBSNLALERBHH24} comprise only a handful of tasks, recent ones like ProteinGym~\cite{NotinKRNPSRSOWF23} and RNAGym~\cite{AroraMACKQSWGGMXN25} now offer dozens to hundreds of tasks to enable more robust evaluation. Yet this scale also makes it harder to interpret the results, and users often abandon task-level scores and resort to averages~\cite{BrixiDKPBCCKLM25,NguyenPDKLLTTKBSNLALERBHH24}. Though grouping scores based on basic task features (e.g., mutational type, taxon) or analyzing performance distribution can offer additional information~\cite{NotinKRNPSRSOWF23,AroraMACKQSWGGMXN25}, they are not sufficient to fully address \textbf{Q1} and \textbf{Q2} that we previously posed. As a result, these benchmarks have not yet been fully leveraged. \texttt{GraphFLA} contributes augmenting them with fitness landscape features that enable biologically meaningful task-level analysis.

\textbf{Landscape analysis in other domains.} Landscape features have also been widely used to describe problem characteristics in black-box optimization (BBO). For example, the classic $24$ BBO benchmarking functions included the \texttt{COCO} platform~\cite{HansenARMTCB21} are classified into $5$ groups (from easy to hard) based on features like separability, modality, etc. Another \texttt{R}-package, \texttt{flacco}~\cite{KerschkeT19}, offers 17 sets of features describing diverse characteristics of the optimization landscape. Similar features also exist for multi-objective optimization problems~\cite{LiefoogheDVDAT20}, and they can enable more informed algorithm testing, comparison, selection~\cite{KerschkeT19b}, and configuration~\cite{KerschkeHNT19}. Yet, all these features are designed for general continuous BBO problems. In contrast, \texttt{GraphFLA} is rooted in evolutionary biology for analyzing sequence-fitness landscapes, and goes beyond simple statistics to biologically meaningful ones.

\section{GraphFLA: A Framework for Fitness Landscape Analysis}
\label{sec:graphfla}

The \texttt{GraphFLA} framework mainly consists of $3$ parts (\pref{fig:intro}c): (1) data preprocessing, (2) landscape construction, and (3) landscape analysis. We purpose-built \texttt{GraphFLA} to meet $4$ desiderata: $\blacktriangleright$ \textbf{Applicability} across empirical landscapes from diverse biological modalities and scales. $\blacktriangleright$ \textbf{ Interoperability} with existing ML-ready data. $\blacktriangleright$ \textbf{Scalability} to efficiently handle landscapes containing millions of genetic variants. $\blacktriangleright$ \textbf{Extendability} to include new analysis methods via an unified API. 

\begin{table*}[t]
    \centering
    \caption{Collection of $20$ essential landscapes features in \texttt{GraphFLA}}
    \fontsize{8}{9}\selectfont
    \renewcommand{\arraystretch}{1.1}
    \begin{tabular}{lllllll}
    \hline
    \textbf{Class} & \textbf{Index} & \textbf{Feature} & \textbf{Ref.} & \textbf{Notation} & \textbf{Range} & \textbf{Higher value indicates} \\
    \hline
    \multirow{6}{*}{Ruggedness}
      & F1 & Fraction of local optima & \cite{PapkouRM23} & $\phi_{\text{lo}}$ & $[0,1]$ & ↑ more peaks \\
      & F2 & Roughness-slope ratio & \cite{AitaIH01} & $r/s$ & $[0,\infty)$ & ↑ ruggedness \\
      & F3 & Autocorrelation & \cite{Weinberger90} & $\rho_a$ & $[-1,1]$ & ↓ ruggedness \\
      & F4 & Gamma statistic & \cite{FerrettiSWYKTA16} & $\gamma$ & $[-1,1]$ & ↑ ruggedness \\
      & F5 & Neighbor-fitness correlation & \cite{Wagner23} & NFC & $[-1,1]$ & ↓ ruggedness \\
    \hline
    \multirow{9}{*}{Epistasis}
      & F6 & Magnitude epistasis & \cite{PoelwijkKWT07} & $\epsilon_{\text{mag}}$ & $[0,1)$ & ↓ evolutionary constraints \\
      & F7 & Sign epistasis & \cite{PoelwijkKWT07} & $\epsilon_{\text{sign}}$ & $[0,1]$ & ↑ evolutionary constraints \\
      & F8 & Reciprocal sign epistasis & \cite{PoelwijkKWT07} & $\epsilon_{\text{reci}}$ & $[0,1]$ & ↑ evolutionary constraints \\
      & F9 & Positive epistasis & \cite{Phillips08} & $\epsilon_{\text{pos}}$ & $[0,1]$ & ↑ synergistic effects \\
      & F10 & Negative epistasis & \cite{Phillips08} & $\epsilon_{\text{neg}}$ & $[0,1]$ & ↑ antagonistic effects \\
      & F11 & Global idiosyncratic index & \cite{BakerleeNSRD22} & $I_{\text{id}}$ & $[0,1]$ & ↑ specific interactions \\
      & F12 & Diminishing return epistasis & \cite{ChouCDSM11} & $\epsilon_{\text{DR}}$ & $[0,1]$ & ↑ flat peaks \\
      & F13 & Increasing cost epistasis & \cite{JohnsonMKD19} & $\epsilon_{\text{IC}}$ & $[0,1]$ & ↑ steep descents \\
      & F14 & Pairwise epistasis & \cite{DomingoDL18} & $\epsilon_{\text{(2)}}$ & $[0,1]$ & ↓ higher-order interactions \\
    \hline
    \multirow{5}{*}{Navigability}
      & F15 & Fitness-distance correlation & \cite{JonesF95} & FDC & $[-1,1]$ & ↑ navigation \\
      & F16 & Glocal optima accessibility & \cite{AguilarRodriguezPW17} & $\alpha_{\text{go}}$ & $[0,1]$ & ↑ access to global peaks \\
      & F17 & Basin-fitness corr. (accessible) & \cite{PapkouRM23} & $\text{BFC}_{\text{acc}}$ & $[-1,1]$ & ↑ access to fitter peaks \\
      & F18 & Basin-fitness corr. (greedy) & \cite{PapkouRM23} & $\text{BFC}_{\text{greedy}}$ & $[-1,1]$ & ↑ access to fitter peaks \\
      & F19 & Evol-enhancing mutation & \cite{Wagner23} & $\phi_{\text{EE}}$ & $[0,1]$ & ↑ evolvability \\
    \hline
    Neutrality & F20 & Neutrality & \cite{PayneW15} & $\eta$ & $[0,1]$ & ↑ neutrality \\
    \hline
    \end{tabular}
\label{tab:landscape_features}
\end{table*}

\textbf{Data input.} To ensure compatibility with existing fitness prediction benchmarks, \texttt{GraphFLA}'s API accepts the standard inputs used by typical ML models. It takes a list of biological sequences (\texttt{X}) and their corresponding fitness values (\texttt{f}), which can be obtained from either random, site-saturation, or combinatorial mutagenesis, or other analogous design. \texttt{GraphFLA} supports sequences of length $n$ where each locus $i\in\{1,\dots,n\}$ can take distinct values from a predefined set $\mathcal{A}_i$ ($|\mathcal{A}_i| \ge 2$; for DNA, $\mathcal{A}_i = \{\texttt{A}, \texttt{C}, \texttt{G}, \texttt{T}\}$). This flexible representation enables \texttt{GraphFLA} to handle various biological modalities, such as DNA, RNA, proteins, genes, and even complex microbial communities. We also include built-in classes optimized for common sequence types (e.g., DNA, RNA, proteins, and binary data) to enhance performance. Additionally, \texttt{GraphFLA}'s preprocessing pipeline automatically detects the composition of the sequence space, standardizes the input data, and identifies duplicates or missing values. This careful preprocessing ensures reliable results in subsequent analyses.

\textbf{Neighborhood identification.} Next, \texttt{GraphFLA} determines a neighborhood structure, which specifies which variants are genetically adjacent to each other in the sequence space. To this end, traditional methods calculate genetic distances between all possible pairs of variants to find one-mutant neighbors~\cite{PapkouRM23,BrouilletAFA15}. However, this pairwise calculation quickly becomes impractical because it requires quadratic time and memory resources. To overcome this scalability issue, \texttt{GraphFLA} employs a more efficient approach: Instead of comparing every pair, it directly generates all potential single-mutation neighbors for each variant. This new strategy achieves nearly linear complexity and significantly outperforms existing implementations in both runtime and memory efficiency (\pref{sec:versatility}). 

\textbf{Landscape as a variant network.} Once the neighborhood is identified, \texttt{GraphFLA} constructs the fitness landscape as a directed graph (\pref{fig:intro}c), wherein each node represents a variant and is associated with its fitness; any two variants that are neighbors to each other are connected by an directed edge, which represents a single mutational step towards the fitter variant. This graph representation, backend by the \texttt{igraph} package in \texttt{C}, allows many landscape analysis to be implemented via efficient graph mining algorithms for significant speed up. For example, locating local optima can be done by finding all \textit{sinks} in this graph, while classifying different types of epistasis is equivalent to finding specific types of $4$-node motifs (\pref{app:graphfla}).

\textbf{Landscape analysis.} For each constructed landscape, \texttt{GraphFLA} offers a comprehensive suite of $20$ features characterizing their $4$ fundamental topographical aspects: ruggedness, navigability, epistasis, and neutrality (\pref{tab:landscape_features}). In selection of these features, we conducted an large language model-assisted, data-driven survey of $1,673$ papers on landscape analysis and evolutionary biology (\pref{app:datasets}), which aims to identify all prevalemt quantitative indicators of landscape topography in literature. From more than $100$ initial candidates, we compiled this final collection of $20$ essential features based on their (1) frequency of appearance in literature, (2) biological significance, (3) coverage across different aspects, (4) computational feasibility, and (5) compatibility with data modalities, sizes, and structures (see details in~\pref{app:datasets}). A full introduction to them is available in~\pref{app:graphfla}

\textbf{Empirical and theoretical landscapes.} Beyond the main modules depicted in~\pref{fig:intro}c, \texttt{GraphFLA} provides a data module featuring 155 empirical fitness landscapes that are combinatorially complete, covering more than 2.2M total variants (\pref{tab:datasets_summary}; \pref{tab:datasets}). These landscapes, gathered via another extensive literature survey (\pref{app:datasets}), span multiple modalities (DNA, RNA, protein) and taxa with diverse fitness metrics. The comprehensive combinatorial nature of this collection distinguishes it from current benchmarks, which mainly consist of randomly generated mutagenesis libraries. This structure enables systematic evaluation of model predictions on combined mutations of varying orders and aids in interpreting results through landscape topography. Although some landscapes in this suite overlap with benchmarks like ProteinGym and RNAGym, \texttt{GraphFLA} is more exhaustive regarding combinatorial libraries. For instance, it contains several 4-site saturation protein landscapes ($20^4=160,000$ total variants) not found in ProteinGym~\cite{JalalTSCLTNLL20,TuSE22,PodgornaiaL15}. In addition to these empirical landscapes, \texttt{GraphFLA} provides 5 theoretical models for generating synthetic landscapes with tunable characteristics (e.g., size, ruggedness; \pref{app:landscape_models})

\begin{table*}[t]
    \centering
    \caption{Summary of combinatorially complete datasets in \texttt{GraphFLA}}
    \fontsize{8}{9}\selectfont
    \renewcommand{\arraystretch}{1.1}
    \begin{tabular}{llll}
    \hline
    \textbf{Modality} & \textbf{Space} & \textbf{No. of Datasets} & \textbf{No. of Mutants} \\
    \hline

    DNA    & Genomic sequence & $55$ & $724$k \\
    \hline
    Protein    & Transcript sequence & $63$ & $1.1$M \\
    \hline
    RNA    & Amino acid sequence & $37$ & $348$k \\
    \hline
    Total &  & $155$ & $2.2$M \\
    \hline
    \end{tabular}
    \label{tab:datasets_summary}
\end{table*}

\section{Results}
\label{sec:results}

\subsection{GraphFLA Enables Efficient and Accurate Landscape Analysis Across Modalities}
\label{sec:versatility}
\begin{figure*}[t!]
    \centering
    \includegraphics[width=\linewidth]{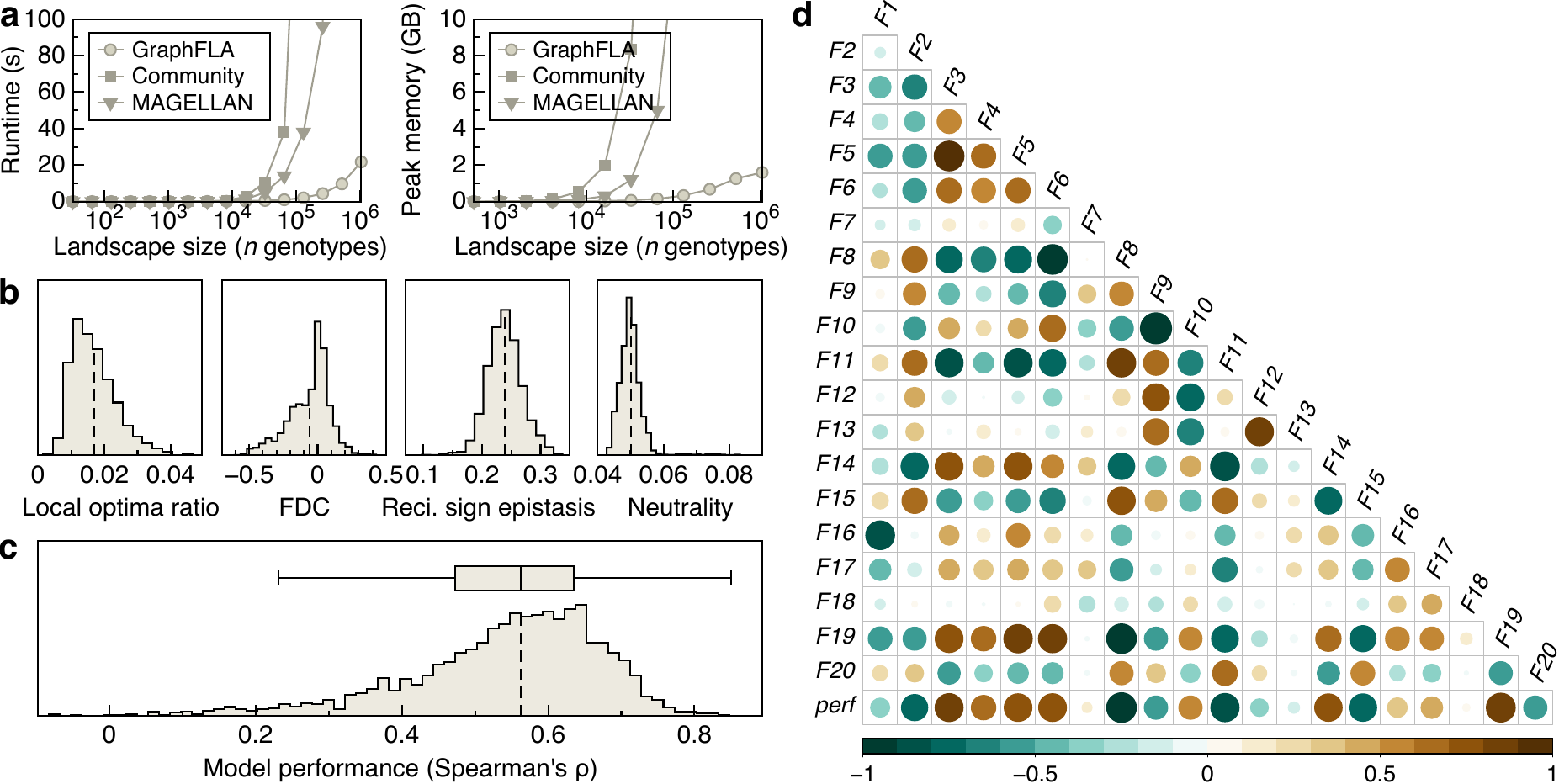}
    \caption{\small
    \textbf{GraphFLA scales efficiently and captures influential landscape features for model performance.}
    \textbf{(a)} Runtime (\textit{left}) and peak memory usage (\textit{right}) during fitness landscape construction for \texttt{GraphFLA}, \texttt{MAGELLAN}, and a community implementation~\cite{PapkouRM23}, as a function of landscape size. Landscapes were generated using the \textit{NK} model~\cite{Kauffman93} by varying the number of loci \textit{N} from $5$ to $20$ ($\to$ landscape sizes from $2^5$ to $2^{20}$). Results shown are averages across $10$ replicates.
    \textbf{(b)} Distribution of $3$ representative landscape features across $155$ combinatorially complete landscapes collected in \texttt{GraphFLA}.
    \textbf{(c)} Distribution of model performance, measured by Spearman's $\rho$, for Evo2 predictions across the same $155$ landscapes.
    \textbf{(d)} Correlation matrix showing Spearman's $\rho$ between $20$ landscape features derived from \texttt{GraphFLA} and Evo2 performance across all $155$ combinatorial landscapes.}
    \label{fig:heatmap}
\end{figure*}

\textbf{Runtime and memory scalability.} To evaluate the performance of \texttt{GraphFLA}, we generated synthetic fitness landscapes of varying sizes using the \textit{NK} model by \citeauthor{Kauffman93}. We then compared the scalability of landscape construction with that of the \texttt{MAGELLAN} package and the community implementation used in~\cite{PapkouRM23} The benchmarks were conducted using a single core of an Intel Xeon Platinum 8260 CPU with 256GB RAM. Our findings in~\pref{fig:heatmap}a show that \texttt{GraphFLA} is significantly faster in landscape construction compared to existing implementations. For example, the two baselines took more than 5h and 3h respectively to construct the landscape at the scale of 1 million mutants, whereas \texttt{GraphFLA} used only 20s. Regarding memory efficiency, \texttt{GraphFLA} stood out by requiring only 2GB memory to process 1 million mutants. In contrast, the community implementation encountered out-of-memory errors when handling over 100,000 mutants. In sum, \texttt{GraphFLA} scales almost linearly with landscape size, allowing it to process even the largest empirical fitness landscapes with millions of mutants.

\textbf{Application to diverse real-world data.} To demonstrate the versatility of \texttt{GraphFLA} in processing real-world data for fitness predictions, we applied it to 4 large-scale benchmarks and databases to construct 5,300+ landscapes in different modalities and sizes for different species. We then calculated the 20 landscape features introduced in~\pref{tab:landscape_features} for each of them (\pref{tab:datasets}, \pref{tab:proteingym}, \pref{tab:rnagym}).
\begin{itemize}[topsep=0pt, leftmargin=1.5em, itemsep=0pt]
    \item \textbf{ProteinGym.} We used its 217 DMS substitution tasks. Since the data is highly curated, \texttt{GraphFLA} can directly construct landscape for each task using the ``\texttt{sequences}'' and ``\texttt{DMS\_score}'' columns, with the only preprocessing being removing the common genetic backgrounds. This results in 217 landscapes (2.2M total mutants) describing protein activity, binding, expression, stability, etc. We note that 168 of these landscapes only contain single mutants, and were excluded from subsequent analysis as there is little information beyond the local neighborhood of the WT.
    \item \textbf{RNAGym.} We used the 33 fitness prediction tasks from it, following the same procedure above. This leads to 31 landscapes (358k total mutants) for mRNAs, tRNAs, aptamers, and ribozymes.
    \item \textbf{Combinatorially complete landscapes.} We constructed 155 combinatorially complete using the datasets introduced in~\pref{tab:datasets_summary}, which contain a mixture of DNA, RNA, and protein landscapes. 
    \item \textbf{CIS-BP database.} It harbors protein-binding microarray (PBM) data for 5,016 TFs of 329 eukaryotic species and 162 DNA-binding domain structural classes from 78 studies~\cite{Weirauch2014}. For each TF, the fitness is its binding affinity to all 32,896 possible 8-nucleotide, double-stranded DNA sequences. \texttt{GraphFLA} constructed these 5,016 TF binding landscapes with 174M total mutants.
\end{itemize}

\textbf{Validation with existing literature.} We proceeded to assess the precision of \texttt{GraphFLA}'s landscape analysis by performing a large-scale replication study across the 61 papers from which our 155 combinatorially complete datasets originate. For each publication, we identified its reported qualitative (e.g., ``highly navigable''~\cite{PapkouRM23}) and quantitative (e.g., ``514 peaks''~\cite{PapkouRM23}) landscape characteristics. We then utilized the relevant metrics within \texttt{GraphFLA} to re-analyze these landscapes, and compared our outputs against the original findings. \texttt{GraphFLA} successfully replicated the qualitative conclusions from all 61 studies (full results in~\pref{tab:datasets}). Notably, for features with unique definitions such as $\phi_\text{lo}$ and $\epsilon_\text{reci}$, \texttt{GraphFLA} precisely reproduced the published values if data processing details were sufficiently described. For features with more generalized definitions for which implementations can vary (e.g., $\epsilon_\text{DR}$), \texttt{GraphFLA}'s analysis consistently supported the conclusions drawn in the original studies. These results demonstrated that \texttt{GraphFLA} is a reliable framework for landscape analysis. 

\subsection{GraphFLA is Robust to Incomplete, Noisy, and Biasedly Sampled Data}
\label{sec:robustness}
A crucial validation for any analysis framework is quantifying its robustness to imperfect data. Real-world empirical landscapes often suffer from (a) missing variants, (b) experimental noise in fitness measurement, or (c) biased sampling in generating the mutant library. In order to validate \texttt{GraphFLA}'s rosbutness to these, we conducted experiments on a \textit{complete} \textit{NK} landscape with moderate size and ruggedness ($n=15, k=7$), which can serve as a reference that represents the most ideal data.

\begin{table*}[t]
    \centering
    \caption{Four essential landscape metrics across reference, incomplete, noisy, and biasedly sampled $NK$ landscapes.}
    \fontsize{8}{9}\selectfont
    \renewcommand{\arraystretch}{1.1}
    \begin{tabular}{lcccc}
        \hline
        \textbf{Setting} & \textbf{Reciprocal sign epistasis} & \textbf{Global optima accessibility} & \textbf{Autocorrelation} & \textbf{FDC} \\
        \hline
        \textbf{reference (complete)} & \textbf{0.1885} & \textbf{0.6729} & \textbf{0.1151} & \textbf{-0.0313} \\
        \hline
        incomplete (10\%)        & 0.1883 & 0.6103 & 0.0927 & -0.0420 \\
        incomplete (20\%)        & 0.1884 & 0.5276 & 0.0771 & -0.0337 \\
        incomplete (50\%)        & 0.1774 & 0.3223 & 0.0553 & -0.0313 \\
        \hline
        noisy (0.01$\sigma$)            & 0.1889 & 0.6492 & 0.0965 & -0.0314 \\
        noisy (0.05$\sigma$)            & 0.1896 & 0.6542 & 0.0927 & -0.0317 \\
        noisy (0.1$\sigma$)             & 0.1921 & 0.6362 & 0.0966 & -0.0319 \\
        noisy (0.2$\sigma$)             & 0.1984 & 0.6339 & 0.0867 & -0.0414 \\
        \hline
        biased (random mutagenesis) & 0.1823 & 0.7246 & 0.1208 & -0.0837 \\
        \hline
    \end{tabular}
    \label{tab:robustness}
\end{table*}

\textbf{Robustness to incomplete data.}
We created \textit{incomplete} landscapes by randomly removing a fraction $\alpha=\{10\%, 20\%, 50\%\}$ of total variants. We then calculated four representative landscape features for these and the reference landscape in~\pref{tab:robustness}. The results shows that most key features are highly robust to data incompleteness. The one exception is global optima accessibility, which, as expected, decreases as more data is removed, since this will destroy paths leading to the global optima regardless of their evolutionary accessibility. Yet, this effect is predictable and can be partially corrected by scaling the measured accessibility by the fraction of remaining data ($1-\alpha$).

\textbf{Robustness to noisy data.} To simulate experimental noise, we perturbed the fitness score of each variant by adding random noise drawn from a Gaussian distribution $\mathcal{N}(0, (\beta\sigma)^2)$, where $\sigma$ is the standard deviation of the original fitness values and the noise level $\beta$ was set to $\{0.01, 0.05, 0.1, 02\}$. The results in~\pref{tab:robustness} demonstrate that all four landscape features remain remarkably stable. Even with noise equivalent to $0.2\sigma$, the calculated values are consistent with the reference. This highlights that \texttt{GraphFLA}'s feature calculations are resilient to typical levels of experimental noise.

\textbf{Robustness to biased sampling.} We then simulated a more realistic scenario of random mutagenesis, which often creates a library that is densely sampled near a wild-type sequence but sparse elsewhere. We created a sparse, \textit{biased} library of 1,804 variants (from 32,768 total) by applying a 10\% per-site mutation rate to the global optimum. As shown in~\pref{tab:robustness}, the key landscape features remain highly consistent with the reference landscape, even when calculated on this much smaller, non-uniform subset. This implies that \texttt{GraphFLA} can still provide reliable approximation of the overall landscape topography even with only biasedly and sparsly sampled data.

\begin{figure*}[t!]
    \centering
    \includegraphics[width=\linewidth]{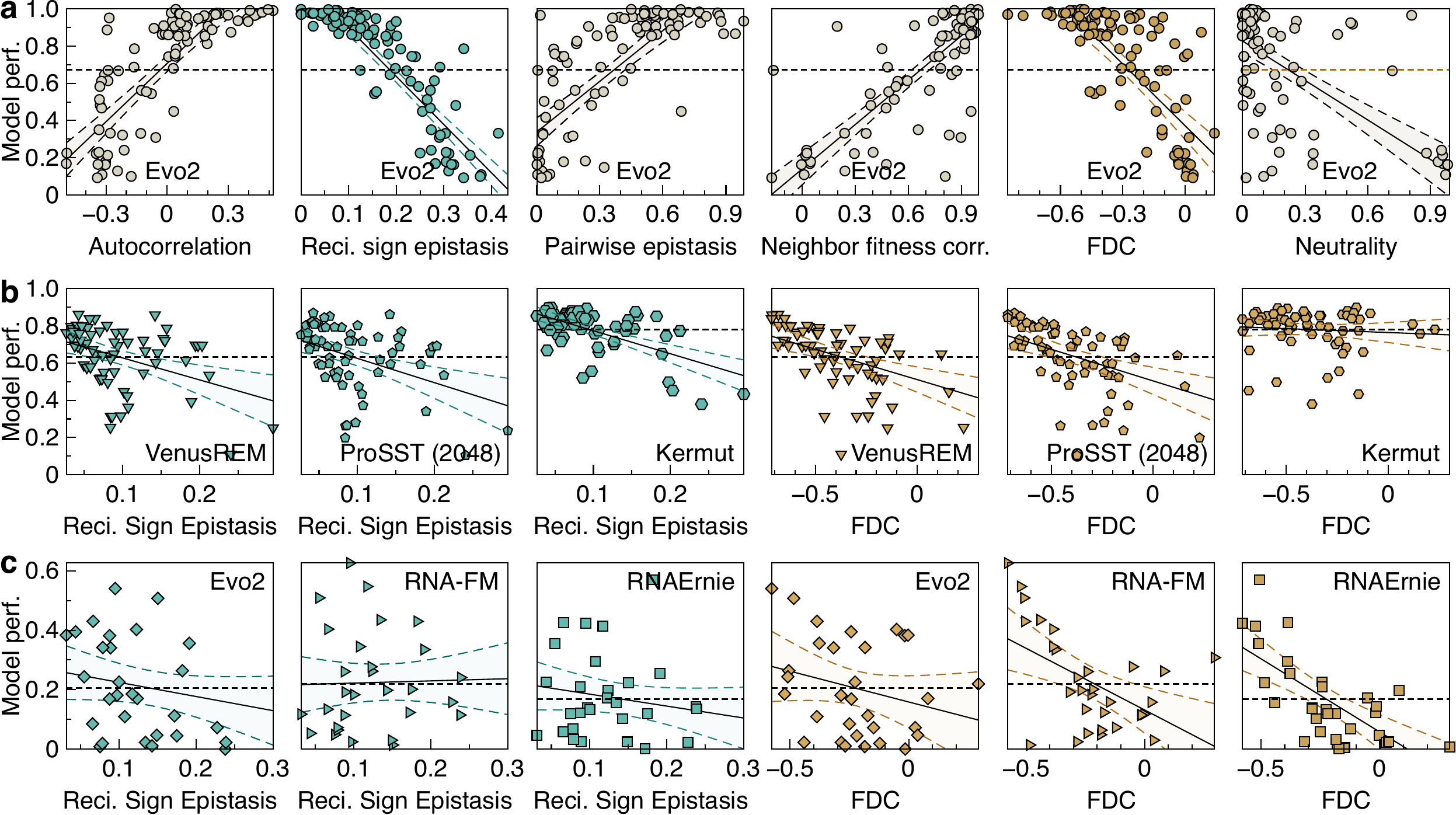}
    \caption{\small\textbf{GraphFLA identifies influencing factors for model performance.} For \textbf{(a)} our $155$ combinatorial landscapes, \textbf{(b)} ProteinGym, and \textbf{(c)} RNAGym, we plot the distribution of model (name specified in each plot) performance ($y$-axis; measured as Spearman's $\rho$) against landscape features ($x$-axis). Straight lines show a fit of the linear regression model, and shaded regions depict the $95\%$ confidence intervals. Dashed horizontal lines indicate the average performance across all landscapes.}
    \label{fig:cc}
\end{figure*}

\begin{figure*}[t!]
    \centering
    \includegraphics[width=\linewidth]{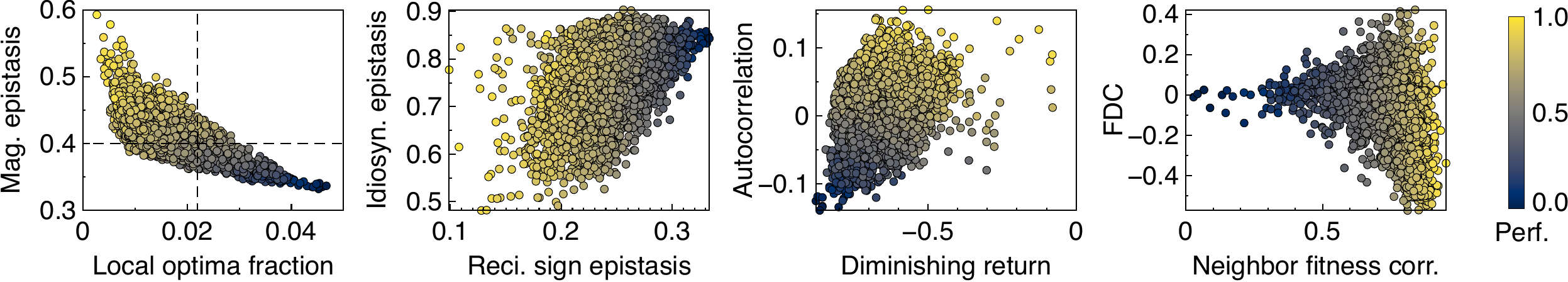}
    \caption{\small\textbf{Visualizing the distribution of model performance in landscape feature space.} We map each of the $5,016$ landscapes constructed from the CIS-BP data in the space composed of landscape features and color-coded with the performance (Spearman's $\rho$) of Evo2-7b to visualize its distribution in the feature space.}
    \label{fig:dna}
\end{figure*}

\subsection{GraphFLA Identifies Key Influence Factors and Bottlenecks in Fitness Prediction}
\label{sec:influence}
The $5,300$ empirical landscapes we constructed in \pref{sec:versatility} exhibit significant variation in their topography. For instance, across the $155$ combinatorially complete landscapes, the percentage of local optima ranges from around $0\%$ to $5\%$ (\pref{fig:heatmap}b). While the former value corresponds to a fairly smooth, unimodal landscape, the latter is on par with that of the most rugged \textit{NK} landscapes~\cite{Weinberger91}. Similar observations can be made for other landscape features (\pref{fig:heatmap}b; \pref{fig:dist_feature}) and for ProteinGym (\pref{fig:dist_feature_proteingym}) as well as RNAGym (\pref{fig:dist_feature_rnagym}). From a performance benchmarking perspective, this is a good sign since it implies that the included tasks are diverse enough to ``stress-test'' models~\cite{SmithMilesM23}.

To illustrate how landscape features can shed light on fitness prediction performance, and thus our \textbf{Q1} in~\pref{sec:introduction}: \textit{``Why did one model perform well on one set of tasks but poorly on another?''}, we used Evo2-7b~\cite{HollmannMPKKHSH25}, the successor of Evo~\cite{NguyenPDKLLTTKBSNLALERBHH24}, as an example. It is trained on $9.3$ trillion DNA base pairs and applicable to diverse modalities including DNA, RNA, and protein. We applied Evo2 to each of our $155$ combinatorial landscapes, and assessed its performance in fitness prediction using Spearman's $\rho$ as in prior works~\cite{HsuNFL22,HopfIPS3SM17,RussFSSBKHHCWR20,RiesselmanIM18,RivesMSGGLLGOZMF21}. We found that Evo2's performance varies significantly across landscapes (\pref{fig:heatmap}c), and is highly dependent on landscape features (\pref{fig:heatmap}d). Specifically, half of the landscape features yielded Spearman's $|\rho|$ higher than $0.6$ with Evo2's performance, and $6$ revealed moderate correlation ($0.3 < |\rho| < 0.6$). Similar results can be obtained by considering partial correlations controlling for landscape size. By taking a closer look at how Evo2's performance varies with landscape features~\pref{fig:cc}a, we found that it struggles on landscapes that are:

\begin{itemize}[leftmargin=1.5em,topsep=0pt,itemsep=0pt]
    \item \textbf{More rugged} and \textbf{more epistatic}. In such landscapes, fitness values often fluctuate dramatically even in local regions within a small genetic distance (low $\rho_a$ and NFC). Extrapolation of such landscapes, even across only a single mutation, may fail due to the existence of local epistatic hotspots (often local optima) resulting from high-order (indicated by low $\epsilon_\text{(2)}$), non-magnitude epistasis (indicated by high $\epsilon_\text{reci}$) between sites that are unique to the current landscape. 
    \item \textbf{Less navigable.} Benign landscapes that are easy to navigate and predict have strongly negative fitness distance correlation (FDC)---variants closer to the global peak in genetic distance tend to have higher fitness. In contrast, when FDC becomes closer to zero or positive, this information diminishes and even becomes ``deceptive'' (e.g., fitness declines as approaching global peak).
    \item \textbf{Highly neutral.} These landscapes feature abundant ``plateau'' regions where mutations have zero fitness effects (i.e., neutral), which can hardly be predicted by models trained on non-neutral data.
\end{itemize}

While such findings were drawn from landscapes with heterogeneous modalities and a general model, similar patterns can be observed in more specific settings. For example, current leading zero-shot models on ProteinGym's $217$ DMS substitution tasks, VenusREM~\cite{TanWWZH24}, ProSST ($k=2,048$)~\cite{LiTMZYZOZTH24}, and the leading supervised model, Kermut~\cite{GrothKOSB24}, tend to excel at fitness prediction for benign protein landscapes with FDC $<-0.5$ and $\epsilon_\text{reci}<0.1$, yet still struggles for more complex ones (\pref{fig:cc}b; more models and features in~\pref{fig:app_proteingym_q1}, \pref{fig:app_proteingym_q1_2}). Established models for RNA fitness prediction like RNA-FM~\cite{ChenHSTWYZYHXSFKL22} and RNAErine~\cite{WangBLLLMKX24} exhibited the same behavior (\pref{fig:cc}c; more in~\pref{fig:app_rnagym_q1}, \pref{fig:app_rnagym_q1_2}). 

To see this at a larger scale, we evaluated the performance of Evo2-7b on the $5,016$ CIS-BP TF binding landscapes described in~\pref{sec:versatility}. We plotted each landscape instance in the feature space shown in~\pref{fig:dna}, and mapped Evo2's performance to this space. From the results, we can observe a clear trend that landscapes located in certain regions of the feature space are in general harder to predict. For example, in the first panel, Quadrant II is occupied by landscapes featuring both a large number of local optima ($\phi_\text{lo}$) and abundant non-magnitude epistasis ($\epsilon_\text{reci}$), and Evo2 can hardly identify the true fitness rank in them despite extensive pre-training. On the other hand, landscapes belonging to Quadrant IV are much more benign with low $\phi_\text{lo}$ and $\epsilon_\text{reci}$. For such landscapes, Evo2 typically achieved Spearman's $\rho > 0.5$. As for Quadrant I and III, landscapes in these regions contain a combination of complex and benign landscapes, which gave rise to mixed performance outcomes. The same trend can be observed for other combinations of features in other panels of~\pref{fig:dna}.

\subsection{GraphFLA Facilitates Landscape-aware Model Comparison}
\label{sec:comparison}
\begin{figure*}[t!]
    \centering
    \includegraphics[width=\linewidth]{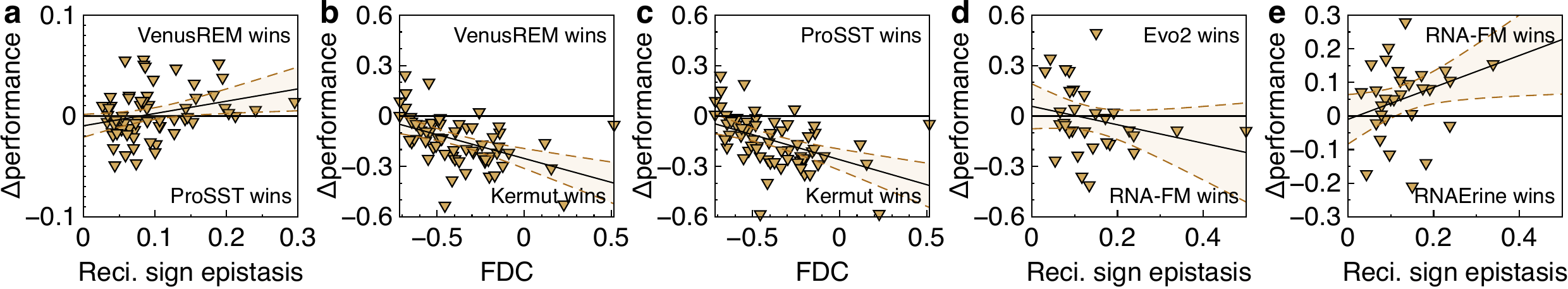}
    \caption{\small\textbf{GraphFLA facilitates landscape-aware model comparison.} Difference in performance ($y$-axis) between $5$ pairs of baselines in ProteinGym (\textbf{a, b, c}) and RNAGym (\textbf{d, e}) is plotted against landscape features on the $x$-axis. Line regression fit lines and $95\%$ confidence intervals are depicted.}
    \label{fig:comparison}
\end{figure*}

Another important question we asked is \textbf{Q2:} ``\textit{why does one model outperform the baseline on one task, but not on the other?}'' For example, though VenusREM and ProSST ($k=2,048$) have similar (Spearman's $\rho$: $0.518$ vs $0.507$) zero-shot performance on ProteinGym, they lead on $53\%$ and $47\%$ of tasks, respectively. Without further information on each task it is hard to strictly distinguish them.

Here we demonstrate that \texttt{GraphFLA}'s landscape features can shed light on their respective advantages. \pref{fig:comparison}a plots the $\Delta$performance between VenusREM and ProSST on each task against the $\epsilon_\text{reci}$ of the corresponding landscape. We found that on benign landscapes with little reciprocal sign epistasis ($\epsilon_{\text{reci}} < 0.1$), ProSST tends to outperform VenusREM (Wilcoxon signed-rank test, $w = 123, p = 0.003$). Yet this advantage diminishes as $\epsilon_{\text{reci}}$ increases, as indicated by a positive slope of the linear regression line in~\pref{fig:comparison}a. Eventually, for landscapes with $\epsilon_{\text{reci}} > 0.15$, VenusREM consistently outperforms ProSST, which implies it is better at capturing complex epistatic interactions.

More intriguingly, both these zero-shot models can only outperform Kermut, the leading supervised baseline, on highly navigable landscapes ($\text{FDC} \approx -0.7$; \pref{fig:comparison}b, c). As landscapes become less navigable (i.e., FDC increases), the performance gap between VenusREM (or ProSST) and Kermut increases drastically (\pref{fig:comparison}b, c). Notably, for the ODP2 landscape from~\cite{TsuboyamaDCLMMWMO23}, which has an $\text{FDC}=0.23$, Kermut outperforms VenusREM by a Spearman's $\rho$ of $0.53$. The same pattern can be observed if we replace Kermut with other supervised baselines like ProteinNPT~\cite{NotinWMG23} (\pref{fig:app_comparison}). These highlight that supervised training is still necessary to better extrapolate on complex landscapes.  

As for RNAGym, though Evo2 and RNA-FM achieved comparable prediction performance across all analyzed landscapes (\pref{fig:cc}c), the former falls short on landscapes with high incidence of reciprocal sign epistasis ($\epsilon_{\text{reci}} > 0.2$; \pref{fig:comparison}d). Also, the performance gap between RNA-FM and RNAErine increases as the landscape becomes more epistatic ($\epsilon_{\text{reci}}$ increases; \pref{fig:comparison}e). These results shed new light on the respective of different models beyond simple averaged scores.

\subsection{GraphFLA is Applicable to a Broader Range of Tasks and Data}
\label{sec:broader_app}

\begin{figure*}[t!]
    \centering
    \includegraphics[width=\linewidth]{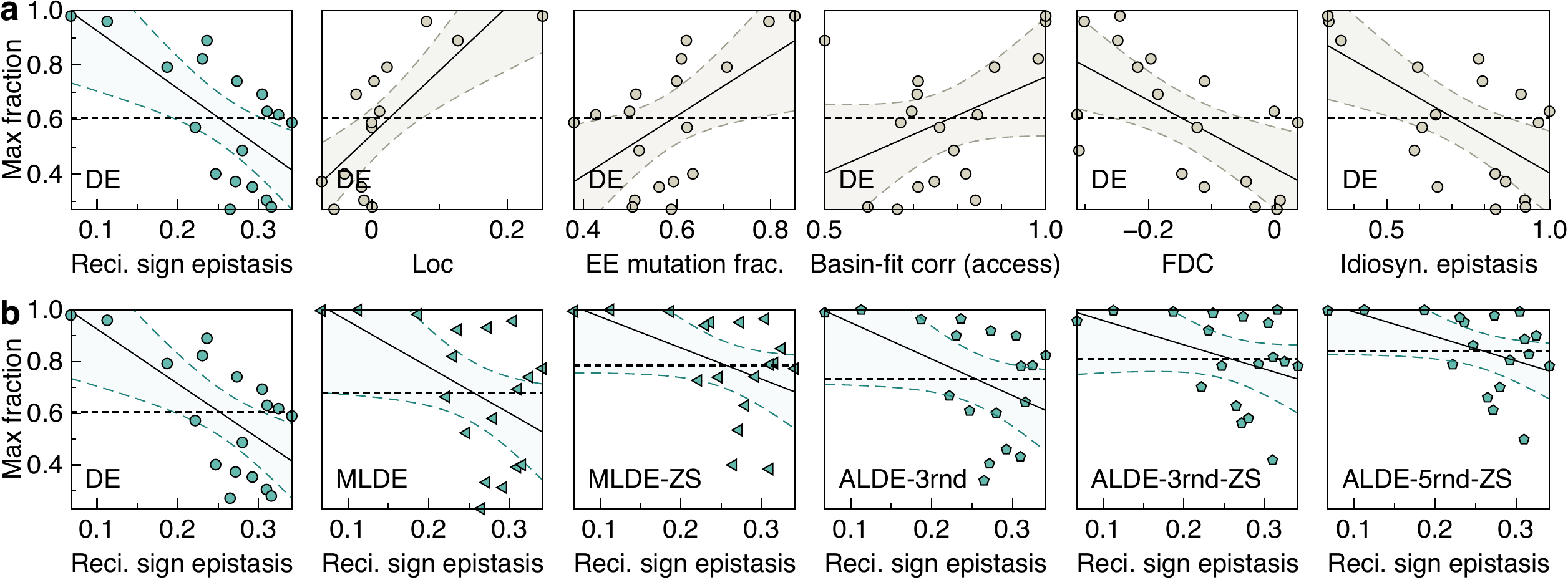}
    \caption{\small\textbf{GraphFLA can be employed to interpret directed evolution (DE) outcomes.} \textbf{(a)} The maximum fitness achieved (in percentile) by DE on each selected landscape against landscape features. \textbf{(b)} We fixed the $x$-axis to be reciprocal sign epistasis and replaced DE with 5 other ML-guided methods. Dashed horizontal lines indicate the average performance across all landscapes.}
    \label{fig:de}
\end{figure*}

\textbf{Application to directed evolution.} Beyond fitness prediction, \texttt{GraphFLA} can also be employed to shed light on other tasks related to fitness landscapes. For example, as directed evolution (DE)~\cite{PackerL15,WangXCYLZ21,WittmannYA21,RomeroKA13}, a central technique in protein engineering, is essentially an adaptive walk on the protein fitness landscape to find high fitness variants, landscape topography can have fundamental impact on its success. For instance, DE on rugged landscapes is notoriously difficult~\cite{YangWA19}. However, a comprehensive understanding of the impact of ruggedness, and other landscape features, on DE, has been missing due to the lack of (1) holistic landscape analysis frameworks like \texttt{GraphFLA}, and (2) large collection of combinatorially complete empirical landscapes like in~\pref{tab:datasets_summary}. 

To demonstrate how \texttt{GraphFLA} can provide insights into DE, we used 20 protein landscapes from our combinatorial library that are 3- or 4-site-saturated (i.e., total variants being $20^3$ or $20^4$). For each of them, we evaluated the performance of 5 DE classic approaches: (1) the simplest DE, implemented as a greedy adaptive walk; (2) ML-guided DE (MLDE)~\cite{YangWA19,WuKLWA19}; (3) MLDE warm-started with a zero short model~\cite{WittmannYA21}; (4) Active learning-guided directed evolution (ALDE)~\cite{QiuHW21,YangLBHAKHYA25}; (5) ALDE with zero-shot warm start. Detailed implementations are available in~\pref{app:de}. 
For each approach, we measured performance using the fitness percentile (where 1 indicates finding the global optimum) of the best variant it found, and aggregated across $100$ randomly initialized runs. For the two ALDE methods, we additionally set the number of iterations to $3$ or $5$ rounds. 

From~\pref{fig:de}a, we can see that while the basic DE method can easily find variants with fitness close to the global optimum on benign landscapes, it struggles on ones that are more rugged and epistatic, and less navigable. The $5$ ML- or active learning-guided approaches are also susceptible to the incidence of epistasis, but more advanced approaches---such as MLDE with a zero-shot warm start and the ALDE variants---demonstrated greater robustness and are less adversely affected (\pref{fig:de}b).

\textbf{Application to other data.} Beyond molecular sequences such as DNA, RNA, and proteins, \texttt{GraphFLA} can be applied to fitness landscapes at various other biological scales. For instance, it can be utilized to analyze evolutionary landscapes of gene regulatory networks~\cite{FriedlanderPBT17,YangS22} or metabolic landscapes~\cite{PinheiroWAL21} at the cellular level. It can also analyze how alterations in community composition impact collective functions~\cite{KeheKOATSGKGFB19,DiazColungaCSRAS24,SkwaraGYDRRTKS23,Diaz-ColungaSVBS24,Sanchez-GorostiagaBOPS19}. As a demo, we analyzed $6$ microbial community-function landscapes in~\pref{tab:microbiem}. In addition to traditional fitness landscapes, \texttt{GraphFLA} can be adapted to study a broad array of phenotype landscapes, a.k.a, genotype-phenotype (GP) maps, for RNA secondary structure~\cite{HofackerEFSTBS94}, protein tertiary structure~\cite{Dill85,LauD89}, and protein complexes~\cite{GreenburyJLA14}, etc. We provide demonstrations for these using computational models and 3 phenotype landscape features in~\pref{app:phenotype}.

\section{Conclusion}
\label{sec:discussion}

\texttt{GraphFLA} addresses the critical lack of meaningful features for interpreting performance benchmarking results in sequence fitness prediction. Using its comprehensive suite of 20 features describing the underlying landscape topography, we are now able to answer questions like \textit{``why model performance varies across tasks?''}, \textit{``when and why will a model outperform the other?''}. In this way, \texttt{GraphFLA} augments current benchmarks like ProteinGym and RNAGym to fully take advantage of their impressive scales, and assists in obtaining granular understanding of the capabilities and limitations of existing genomic models that were previously impossible. In addition, since \texttt{GraphFLA} itself is designed for arbitrary combinatorial landscapes, we expect it will be a useful resource for advancing our understanding on a broader range of tasks and data that is related to combinatorial optimization. 

\textbf{Acknowledgements.} We sincerely thank all the reviewers for their encouraging and constructive feedback. This work was supported by the UKRI Future Leaders Fellowship under Grant MR/S017062/1 and MR/X011135/1; in part by NSFC under Grant 62376056 and 62076056; in part by the Royal Society Faraday Discovery Fellowship (FDF/S2/251014), BBSRC Transformative Research Technologies (UKRI1875), Royal Society International Exchanges Award (IES/R3/243136), Kan Tong Po Fellowship (KTP/R1/231017); and the Amazon Research Award and Alan Turing Fellowship.




\newpage
\bibliographystyle{unsrtnat}
\bibliography{icml2024}

\begin{thebibliography}{193}
\providecommand{\natexlab}[1]{#1}
\providecommand{\url}[1]{\texttt{#1}}
\expandafter\ifx\csname urlstyle\endcsname\relax
  \providecommand{\doi}[1]{doi: #1}\else
  \providecommand{\doi}{doi: \begingroup \urlstyle{rm}\Url}\fi

\bibitem[Wright(1932)]{Wright32}
S.~Wright.
\newblock The roles of mutation, inbreeding, crossbreeding and selection in
  evolution.
\newblock \emph{Proc. XI Int. Congr. Genet.}, 8:\penalty0 209--222, 1932.

\bibitem[Rizvi et~al.(2015)Rizvi, Hellmann, Snyder, Kvistborg, Makarov, Havel,
  Lee, Yuan, Wong, Ho, Miller, Rekhtman, Moreira, Ibrahim, Bruggeman, Gasmi,
  Zappasodi, Maeda, Sander, Garon, Merghoub, Wolchok, Schumacher, and
  Chan]{RizviHSMKHLYWWHMMRMIIBGZMSCWSC15}
N.~A. Rizvi, M.~D. Hellmann, A.~Snyder, P.~Kvistborg, V.~Makarov, J.~J. Havel,
  W.~Lee, J.~Yuan, P.~Wong, T.~S. Ho, M.~L. Miller, N.~Rekhtman, A.~L. Moreira,
  F.~Ibrahim, C.~Bruggeman, B.~Gasmi, R.~Zappasodi, Y.~Maeda, C.~Sander, E.~B.
  Garon, T.~Merghoub, J.~D. Wolchok, T.~N. Schumacher, and T.~A. Chan.
\newblock Cancer immunology. mutational landscape determines sensitivity to
  {PD}-1 blockade in non-small cell lung cancer.
\newblock \emph{Science}, 348\penalty0 (6230):\penalty0 124--128, 2015.

\bibitem[JZ et~al.(2022)JZ, DM, and N]{ChenFT22}
Chen JZ, Fowler DM, and Tokuriki N.
\newblock Environmental selection and epistasis in an empirical
  phenotype-environment-fitness landscape.
\newblock \emph{Nat. Ecol. Evol.}, 6\penalty0 (4):\penalty0 427--438, 2022.

\bibitem[Papkou et~al.(2023)Papkou, Regev, and Masel]{PapkouRM23}
Andrei Papkou, Aviv Regev, and Joanna Masel.
\newblock A rugged yet easily navigable fitness landscape.
\newblock \emph{Science}, 382:\penalty0 eadh3860, 2023.

\bibitem[Weinreich et~al.(2006)Weinreich, Delaney, DePristo, and
  Hartl]{WeinreichDDH06}
Daniel~M. Weinreich, Nigel Delaney, Mark~A. DePristo, and Daniel~L. Hartl.
\newblock Darwinian evolution can follow only very few mutational paths to
  fitter proteins.
\newblock \emph{Science}, 312\penalty0 (5770):\penalty0 111--114, 2006.

\bibitem[Lunzer et~al.(2005)Lunzer, Miller, Felsheim, and Dean]{LunzerMFD05}
M.~Lunzer, S.~P. Miller, R.~Felsheim, and A.~M. Dean.
\newblock The biochemical architecture of an ancient adaptive landscape.
\newblock \emph{Science}, 310\penalty0 (5747):\penalty0 499--501, 2005.

\bibitem[Johnston et~al.(2024)Johnston, Almhjell, Watkins-Dulaney, Liu, Porter,
  Yang, and Arnold]{JohnstonAWLPYA24}
K.~E. Johnston, P.~J. Almhjell, E.~J. Watkins-Dulaney, G.~Liu, N.~J. Porter,
  J.~Yang, and F.~H. Arnold.
\newblock A combinatorially complete epistatic fitness landscape in an enzyme
  active site.
\newblock \emph{Proc. Natl. Acad. Sci. U. S. A.}, 121\penalty0 (32):\penalty0
  e2400439121, 2024.

\bibitem[Judge et~al.(2024)Judge, Sankaran, Hu, Palaniappan, Birgy, Prasad, and
  Palzkill]{JudgeSHPBPP24}
A.~Judge, B.~Sankaran, L.~Hu, M.~Palaniappan, A.~Birgy, B.~V.~V. Prasad, and
  T.~Palzkill.
\newblock Network of epistatic interactions in an enzyme active site revealed
  by large-scale deep mutational scanning.
\newblock \emph{Proc. Natl. Acad. Sci. U. S. A.}, 121\penalty0 (12):\penalty0
  e2313513121, 2024.

\bibitem[Zheng et~al.(2020)Zheng, Guo, and Wagner]{ZhengGW20}
J.~Zheng, N.~Guo, and A.~Wagner.
\newblock Selection enhances protein evolvability by increasing mutational
  robustness and foldability.
\newblock \emph{Science}, 370\penalty0 (6521):\penalty0 eabb5962, 2020.

\bibitem[Ogden et~al.(2019)Ogden, Kelsic, Sinai, and Church]{OgdenKSC19}
P.~J. Ogden, E.~D. Kelsic, S.~Sinai, and G.~M. Church.
\newblock Comprehensive {AAV} capsid fitness landscape reveals a viral gene and
  enables machine-guided design.
\newblock \emph{Science}, 366\penalty0 (6469):\penalty0 1139--1143, 2019.

\bibitem[Sarkisyan et~al.(2016)]{SarkisyanEtAl16}
K.~S. Sarkisyan et~al.
\newblock Local fitness landscape of the green fluorescent protein.
\newblock \emph{Nature}, 533:\penalty0 397--401, 2016.

\bibitem[Li et~al.(2016)Li, Qian, Maclean, and Zhang]{LiQMZ16}
C.~Li, W.~Qian, M.~Maclean, and J.~Zhang.
\newblock The fitness landscape of a {tRNA} gene.
\newblock \emph{Science}, 352:\penalty0 837--840, 2016.

\bibitem[Greenbury et~al.(2022)Greenbury, Louis, and Ahnert]{GreenburyLA22}
Sam~F. Greenbury, Ard~A. Louis, and Sebastian~E. Ahnert.
\newblock The structure of genotype-phenotype maps makes fitness landscapes
  navigable.
\newblock \emph{Nat. Ecol. Evol.}, 6\penalty0 (11):\penalty0 1742--1752, 2022.

\bibitem[Puchta et~al.(2016)Puchta, Cseke, Czaja, Tollervey, Sanguinetti, and
  Kudla]{PuchtaCCTSK16}
O.~Puchta, B.~Cseke, H.~Czaja, D.~Tollervey, G.~Sanguinetti, and G.~Kudla.
\newblock Network of epistatic interactions within a yeast snorna.
\newblock \emph{Science}, 352\penalty0 (6287):\penalty0 840--844, 2016.

\bibitem[Jiménez et~al.(2013)Jiménez, Xulvi-Brunet, Campbell, Turk-MacLeod,
  and Chen]{JimenezXCTC13}
J.~I. Jiménez, R.~Xulvi-Brunet, G.~W. Campbell, R.~Turk-MacLeod, and I.~A.
  Chen.
\newblock Comprehensive experimental fitness landscape and evolutionary network
  for small {RNA}.
\newblock \emph{Proc. Natl. Acad. Sci. U. S. A.}, 110\penalty0 (37):\penalty0
  14984--14989, 2013.

\bibitem[Pitt and Ferré-D'Amaré(2010)]{PittF10}
Joel~N. Pitt and Adrian~R. Ferré-D'Amaré.
\newblock Rapid construction of empirical {RNA} fitness landscapes.
\newblock \emph{Science}, 330\penalty0 (6002):\penalty0 376--379, 2010.

\bibitem[Payne and Wagner(2014)]{PayneW14}
Joshua~L. Payne and Andreas Wagner.
\newblock The robustness and evolvability of transcription factor binding
  sites.
\newblock \emph{Science}, 343\penalty0 (6173):\penalty0 875--877, 2014.

\bibitem[Aguilar-Rodríguez et~al.(2017)Aguilar-Rodríguez, Payne, and
  Wagner]{AguilarRodriguezPW17}
Juan Aguilar-Rodríguez, Joshua~L. Payne, and Andreas Wagner.
\newblock A thousand empirical adaptive landscapes and their navigability.
\newblock \emph{Nat. Ecol. Evol.}, 1\penalty0 (2):\penalty0 45, 2017.

\bibitem[Notin et~al.(2024)Notin, Rollins, Gal, Sander, and
  Marks]{NotinRGGSM24}
P.~Notin, N.~Rollins, Y.~Gal, C.~Sander, and D.~Marks.
\newblock Machine learning for functional protein design.
\newblock \emph{Nat. Biotechnol.}, 42\penalty0 (2):\penalty0 216--228, 2024.

\bibitem[Hayes et~al.(2025{\natexlab{a}})Hayes, Rao, Akin, Sofroniew, Oktay,
  Lin, Verkuil, Tran, Deaton, Wiggert, Badkundri, Shafkat, Gong, Derry, Molina,
  Thomas, Khan, Mishra, Kim, Bartie, Nemeth, Hsu, Sercu, Candido, and
  Rives]{HayesRASNOZVTWDWBSGDMTKMKBNHSC25}
T.~Hayes, R.~Rao, H.~Akin, N.~J. Sofroniew, D.~Oktay, Z.~Lin, R.~Verkuil, V.~Q.
  Tran, J.~Deaton, M.~Wiggert, R.~Badkundri, I.~Shafkat, J.~Gong, A.~Derry,
  R.~S. Molina, N.~Thomas, Y.~A. Khan, C.~Mishra, C.~Kim, L.~J. Bartie,
  M.~Nemeth, P.~D. Hsu, T.~Sercu, S.~Candido, and A.~Rives.
\newblock Simulating 500 million years of evolution with a language model.
\newblock \emph{Science}, 387\penalty0 (6736):\penalty0 850--858,
  2025{\natexlab{a}}.

\bibitem[Jiang et~al.(2024)Jiang, Yan, Bernardo, Sgrizzi, Villiger, Kayabolen,
  Kim, Carscadden, Hiraizumi, Nishimasu, Gootenberg, and
  Abudayyeh]{JiangYDBSVKKCHNGGA24}
K.~Jiang, Z.~Yan, M.~Di Bernardo, S.~R. Sgrizzi, L.~Villiger, A.~Kayabolen,
  B.~Kim, J.~K. Carscadden, M.~Hiraizumi, H.~Nishimasu, J.~S. Gootenberg, and
  O.~O. Abudayyeh.
\newblock Rapid protein evolution by few-shot learning with a protein language
  model.
\newblock \emph{bioRxiv}, 2024:\penalty0 2024.07.17.604015, 2024.

\bibitem[Vaishnav et~al.(2022)Vaishnav, de~Boer, Molinet, Yassour, Fan,
  Adiconis, Thompson, Levin, Cubillos, and Regev]{VaishnavBMYFATLCR22}
E.~D. Vaishnav, C.~G. de~Boer, J.~Molinet, M.~Yassour, L.~Fan, X.~Adiconis,
  D.~A. Thompson, J.~Z. Levin, F.~A. Cubillos, and A.~Regev.
\newblock The evolution, evolvability and engineering of gene regulatory {DNA}.
\newblock \emph{Nature}, 603\penalty0 (7901):\penalty0 455--463, 2022.

\bibitem[Yu et~al.(2024)Yu, Yang, Sun, Yan, Yang, Zhang, Ding, and
  Li]{YuYSYZYDLK24}
H.~Yu, H.~Yang, W.~Sun, Z.~Yan, X.~Yang, H.~Zhang, Y.~Ding, and K.~Li.
\newblock An interpretable {RNA} foundation model for exploring functional
  {RNA} motifs in plants.
\newblock \emph{Nat. Mach. Intell.}, 6\penalty0 (12):\penalty0 1616--1625,
  2024.

\bibitem[Tan et~al.(2024)Tan, Wang, Wu, Hong, and Zhou]{TanWWZH24}
Yang Tan, Ruilin Wang, Banghao Wu, Liang Hong, and Bingxin Zhou.
\newblock Retrieval-enhanced mutation mastery: Augmenting zero-shot prediction
  of protein language model.
\newblock \emph{CoRR}, abs/2410.21127, 2024.

\bibitem[Li et~al.(2024)Li, Tan, Ma, Zhong, Yu, Zhou, Ouyang, Zhou, Tan, and
  Hong]{LiTMZYZOZTH24}
Li, Tan, Ma, Zhong, Yu, Zhou, Ouyang, Zhou, Tan, and Hong.
\newblock Prosst: protein language modeling with quantized structure and
  disentangled attention.
\newblock \emph{{NeurIPS}'24: Proc. of Advances in Neural Information
  Processing Systems}, 38, 2024.

\bibitem[Groth et~al.(2024)Groth, Kerrn, Olsen, Salomon, and
  Boomsma]{GrothKOSB24}
Groth, Kerrn, Olsen, Salomon, and Boomsma.
\newblock Kermut: composite kernel regression for protein variant effects.
\newblock \emph{{NeurIPS}'24: Proc. of Advances in Neural Information
  Processing Systems}, 38, 2024.

\bibitem[Hollmann et~al.(2025)Hollmann, Müller, Purucker, Krishnakumar,
  Körfer, Hoo, Schirrmeister, and Hutter]{HollmannMPKKHSH25}
N.~Hollmann, S.~Müller, L.~Purucker, A.~Krishnakumar, M.~Körfer, S.~B. Hoo,
  R.~T. Schirrmeister, and F.~Hutter.
\newblock Accurate predictions on small data with a tabular foundation model.
\newblock \emph{Nature}, 637\penalty0 (8045):\penalty0 319--326, 2025.

\bibitem[Chen et~al.(2022)Chen, Hu, Sun, Tan, Wang, Yu, Zong, Hong, Xiao, Shen,
  King, and Li]{ChenHSTWYZYHXSFKL22}
Jiayang Chen, Zhihang Hu, Siqi Sun, Qingxiong Tan, Yixuan Wang, Qinze Yu,
  Licheng Zong, Liang Hong, Jin Xiao, Tao Shen, Irwin King, and Yu~Li.
\newblock Interpretable {RNA} foundation model from unannotated data for highly
  accurate {RNA} structure and function predictions.
\newblock \emph{arXiv}, 2204:\penalty0 00300, 2022.

\bibitem[Wang et~al.(2024)Wang, Bian, Li, Li, Mumtaz, Kong, and
  Xiong]{WangBLLLMKX24}
Ning Wang, Jiang Bian, Yuchen Li, Xuhong Li, Shahid Mumtaz, Linghe Kong, and
  Haoyi Xiong.
\newblock Multi-purpose {RNA} language modelling with motif-aware pretraining
  and type-guided fine-tuning.
\newblock \emph{Nat. Mach. Intell.}, 6\penalty0 (5):\penalty0 548--557, 2024.

\bibitem[Brixi et~al.(2025)Brixi, Durrant, Ku, Poli, Brockman, Chang, Gonzalez,
  King, Li, Merchant, Naghipourfar, Nguyen, Ricci-Tam, Romero, Sun,
  Taghibakshi, Vorontsov, Yang, Deng, Gorton, Nguyen, Wang, Adams, Baccus,
  Dillmann, Ermon, Guo, Ilango, Janik, Lu, Mehta, Mofrad, Ng, Pannu, Ré,
  Schmok, John, Sullivan, Zhu, Zynda, Balsam, Collison, Costa,
  Hernandez-Boussard, Ho, Liu, McGrath, Powell, Burke, Goodarzi, Hsu, and
  Hie]{BrixiDKPBCCKLM25}
Garyk Brixi, Matthew~G. Durrant, Jerome Ku, Michael Poli, Greg Brockman, Daniel
  Chang, Gabriel~A. Gonzalez, Samuel~H. King, David~B. Li, Aditi~T. Merchant,
  Mohsen Naghipourfar, Eric Nguyen, Chiara Ricci-Tam, David~W. Romero, Gwanggyu
  Sun, Ali Taghibakshi, Anton Vorontsov, Brandon Yang, Myra Deng, Liv Gorton,
  Nam Nguyen, Nicholas~K. Wang, Etowah Adams, Stephen~A. Baccus, Steven
  Dillmann, Stefano Ermon, Daniel Guo, Rajesh Ilango, Ken Janik, Amy~X. Lu,
  Reshma Mehta, Mohammad~R.K. Mofrad, Madelena~Y. Ng, Jaspreet Pannu,
  Christopher Ré, Jonathan~C. Schmok, John~St. John, Jeremy Sullivan, Kevin
  Zhu, Greg Zynda, Daniel Balsam, Patrick Collison, Anthony~B. Costa, Tina
  Hernandez-Boussard, Eric Ho, Ming-Yu Liu, Thomas McGrath, Kimberly Powell,
  Dave~P. Burke, Hani Goodarzi, Patrick~D. Hsu, and Brian~L. Hie.
\newblock Genome modeling and design across all domains of life with {Evo 2}.
\newblock \emph{bioRxiv}, 2025:\penalty0 2025.02.18.638918, 2025.

\bibitem[Notin et~al.(2023{\natexlab{a}})Notin, Kollasch, Ritter, van Niekerk,
  Paul, Spinner, Rollins, Shaw, Orenbuch, Weitzman, Frazer, Dias, Franceschi,
  Gal, and Marks]{NotinKRNPSRSOWF23}
Pascal Notin, Aaron Kollasch, Daniel Ritter, Lood van Niekerk, Steffanie Paul,
  Han Spinner, Nathan~J. Rollins, Ada Shaw, Rose Orenbuch, Ruben Weitzman,
  Jonathan Frazer, Mafalda Dias, Dinko Franceschi, Yarin Gal, and Debora~S.
  Marks.
\newblock Proteingym: Large-scale benchmarks for protein fitness prediction and
  design.
\newblock In \emph{{NeurIPS}'23: Proc. of Advances in Neural Information
  Processing Systems 36}, 2023{\natexlab{a}}.

\bibitem[Arora et~al.(2025)Arora, Angelo, Choe, Kollasch, Qu, Shearer,
  Weitzman, Gazizov, Gurev, Xie, Marks, and Notin]{AroraMACKQSWGGMXN25}
Rohit Arora, Murphy Angelo, Christian~Andrew Choe, Aaron~W Kollasch, Fiona Qu,
  Courtney~A. Shearer, Ruben Weitzman, Artem Gazizov, Sarah Gurev, Erik Xie,
  Debora~Susan Marks, and Pascal Notin.
\newblock {RNAG}ym: benchmarks for {RNA} fitness and structure prediction.
\newblock In \emph{ICLR 2025 Workshop on Machine Learning for Genomics
  Explorations}, 2025.

\bibitem[Wolpert and Macready(1997)]{DolpertM97}
David~H. Wolpert and William~G. Macready.
\newblock No free lunch theorems for optimization.
\newblock \emph{{IEEE} Trans. Evol. Comput.}, 1\penalty0 (1):\penalty0 67--82,
  1997.

\bibitem[Westmann et~al.(2024)Westmann, Goldbach, and Wagner]{WestmannGW24}
C.~A. Westmann, L.~Goldbach, and A.~Wagner.
\newblock The highly rugged yet navigable regulatory landscape of the bacterial
  transcription factor {TetR}.
\newblock \emph{Nat. Commun.}, 15\penalty0 (1):\penalty0 10745, 2024.

\bibitem[J and DB(2015)]{VanCleveW15}
Van~Cleve J and Weissman DB.
\newblock Measuring ruggedness in fitness landscapes.
\newblock \emph{Proc. Natl. Acad. Sci. U. S. A.}, 112\penalty0 (24):\penalty0
  7345--7346, 2015.

\bibitem[Poelwijk et~al.(2007)Poelwijk, Kiviet, Weinreich, and
  Tans]{PoelwijkKWT07}
Frank~J. Poelwijk, Daan~J. Kiviet, Daniel~M. Weinreich, and Sander~J. Tans.
\newblock Empirical fitness landscapes reveal accessible evolutionary paths.
\newblock \emph{Nature}, 445\penalty0 (7126):\penalty0 383--386, 2007.

\bibitem[Bank(2022)]{Bank22}
Claudia Bank.
\newblock Epistasis and adaptation on fitness landscapes.
\newblock \emph{Annu. Rev. Ecol. Evol. Syst.}, 53:\penalty0 457--479, 2022.

\bibitem[Chou et~al.(2011)Chou, Chiu, Delaney, Segr{\`e}, and Marx]{ChouCDSM11}
HH~Chou, HC~Chiu, NF~Delaney, D~Segr{\`e}, and CJ~Marx.
\newblock Diminishing returns epistasis among beneficial mutations decelerates
  adaptation.
\newblock \emph{Science}, 332\penalty0 (6034):\penalty0 1190--1192, 2011.

\bibitem[Bakerlee et~al.(2022)Bakerlee, Ba, Shulgina, Echenique, and
  Desai]{BakerleeNSRD22}
CW~Bakerlee, AN~Nguyen Ba, Y~Shulgina, JI~Rojas Echenique, and MM~Desai.
\newblock Idiosyncratic epistasis leads to global fitness-correlated trends.
\newblock \emph{Science}, 376\penalty0 (6593):\penalty0 630--635, 2022.

\bibitem[Otwinowski et~al.(2018)Otwinowski, McCandlish, and
  Plotkin]{OtwinowskiMP18}
Jakub Otwinowski, David~M. McCandlish, and Joshua~B. Plotkin.
\newblock Inferring the shape of global epistasis.
\newblock \emph{Proc. Natl. Acad. Sci. U. S. A.}, 115\penalty0 (32):\penalty0
  E7550--E7558, 2018.

\bibitem[Khan et~al.(2011)Khan, Dinh, Schneider, Lenski, and
  Cooper]{KhanDSLC11}
A.~I. Khan, D.~M. Dinh, D.~Schneider, R.~E. Lenski, and T.~F. Cooper.
\newblock {N}egative epistasis between beneficial mutations in an evolving
  bacterial population.
\newblock \emph{Science}, 332\penalty0 (6034):\penalty0 1193--1196, 2011.

\bibitem[Domingo et~al.(2018)Domingo, Diss, and Lehner]{DomingoDL18}
J.~Domingo, G.~Diss, and B.~Lehner.
\newblock Pairwise and higher-order genetic interactions during the evolution
  of a {tRNA}.
\newblock \emph{Nature}, 558\penalty0 (7708):\penalty0 117--121, 2018.

\bibitem[Lauring et~al.(2013)Lauring, Frydman, and Andino]{LauringFA13}
Andrew~S. Lauring, Judith Frydman, and Raul Andino.
\newblock The role of mutational robustness in {RNA} virus evolution.
\newblock \emph{Nat. Rev. Microbiol.}, 11\penalty0 (5):\penalty0 327--336,
  2013.

\bibitem[Draghi et~al.(2010)Draghi, Parsons, Wagner, and Plotkin]{DraghiPWP10}
Jeremy~A. Draghi, Todd~L. Parsons, Günter~P. Wagner, and Joshua~B. Plotkin.
\newblock Mutational robustness can facilitate adaptation.
\newblock \emph{Nature}, 463\penalty0 (7279):\penalty0 353--355, 2010.

\bibitem[Félix and Barkoulas(2015)]{FelixB15}
M.~A. Félix and M.~Barkoulas.
\newblock Pervasive robustness in biological systems.
\newblock \emph{Nat. Rev. Genet.}, 16\penalty0 (8):\penalty0 483--496, 2015.

\bibitem[Weirauch et~al.(2014)Weirauch, Yang, Albu, Cote, Montenegro-Montero,
  Drewe, Najafabadi, Lambert, Mann, Cook, Zheng, Goity, van Bakel, Lozano,
  Galli, Lewsey, Huang, Mukherjee, Chen, Reece-Hoyes, Govindarajan, Shaulsky,
  Walhout, Bouget, R{\"a}tsch, Larrondo, Ecker, and Hughes]{Weirauch2014}
Matthew~T. Weirauch, Alina Yang, Mihai Albu, Andre~G. Cote, Andres
  Montenegro-Montero, Philipp Drewe, Hamed~S. Najafabadi, Samuel~A. Lambert,
  Ian Mann, Katie Cook, Hao Zheng, Andrea Goity, Harm van Bakel, Juan~C.
  Lozano, Mary Galli, Matthew~G. Lewsey, Eric Huang, Tathagata Mukherjee,
  Xianjun Chen, John~S. Reece-Hoyes, Suganthi Govindarajan, Gad Shaulsky,
  Albertha J.~M. Walhout, Fran{\c c}ois-Yves Bouget, Gunnar R{\"a}tsch, Luis~F.
  Larrondo, Joseph~R. Ecker, and Timothy~R. Hughes.
\newblock Determination and inference of eukaryotic transcription factor
  sequence specificity.
\newblock \emph{Cell}, 158\penalty0 (6):\penalty0 1431--1443, September 2014.

\bibitem[Podgornaia and Laub(2015)]{PodgornaiaL15}
Anna~I. Podgornaia and Michael~T. Laub.
\newblock Pervasive degeneracy and epistasis in a protein-protein interface.
\newblock \emph{Science}, 347\penalty0 (6222):\penalty0 673--677, 2015.

\bibitem[Friedlander et~al.(2017)Friedlander, Prizak, Barton, and
  Tkačik]{FriedlanderPBT17}
T.~Friedlander, R.~Prizak, N.~H. Barton, and G.~Tkačik.
\newblock Evolution of new regulatory functions on biophysically realistic
  fitness landscapes.
\newblock \emph{Nat. Commun.}, 8\penalty0 (1):\penalty0 216, 2017.

\bibitem[Yang and Scarpino(2022)]{YangS22}
C.~H. Yang and S.~V. Scarpino.
\newblock A family of fitness landscapes modeled through gene regulatory
  networks.
\newblock \emph{Entropy}, 24\penalty0 (5):\penalty0 622, 2022.

\bibitem[Kehe et~al.(2019)Kehe, Kulesa, Ortiz, Ackerman, Thakku, Sellers,
  Kuehn, Gore, Friedman, and Blainey]{KeheKOATSGKGFB19}
J.~Kehe, A.~Kulesa, A.~Ortiz, C.M. Ackerman, S.G. Thakku, D.~Sellers, S.~Kuehn,
  J.~Gore, J.~Friedman, and P.C. Blainey.
\newblock Massively parallel screening of synthetic microbial communities.
\newblock \emph{Proc. Natl. Acad. Sci. U.S.A.}, 116\penalty0 (26):\penalty0
  12804--12809, 2019.

\bibitem[Diaz-Colunga et~al.(2024{\natexlab{a}})Diaz-Colunga, Catalan, Roman,
  Arrabal, and Sanchez]{DiazColungaCSRAS24}
Juan Diaz-Colunga, Pablo Catalan, Magdalena~San Roman, Andrea Arrabal, and
  Alvaro Sanchez.
\newblock Full factorial construction of synthetic microbial communities.
\newblock \emph{eLife}, 13:\penalty0 RP101906, 2024{\natexlab{a}}.

\bibitem[Skwara et~al.(2023)Skwara, Gowda, Yousef, Diaz-Colunga, Raman,
  Sanchez, Tikhonov, and Kuehn]{SkwaraGYDRRTKS23}
Abigail Skwara, Karna Gowda, Mahmoud Yousef, Juan Diaz-Colunga, Arjun~S. Raman,
  Alvaro Sanchez, Mikhail Tikhonov, and Seppe Kuehn.
\newblock Statistically learning the functional landscape of microbial
  communities.
\newblock \emph{Nat. Ecol. Evol.}, 7:\penalty0 1823--1833, 2023.

\bibitem[Diaz-Colunga et~al.(2024{\natexlab{b}})Diaz-Colunga, Skwara, Vila,
  Bajic, and Sanchez]{Diaz-ColungaSVBS24}
Juan Diaz-Colunga, Abigail Skwara, Jean~C.C. Vila, Djordje Bajic, and Alvaro
  Sanchez.
\newblock Global epistasis and the emergence of function in microbial
  consortia.
\newblock \emph{Cell}, 187\penalty0 (12):\penalty0 3108--3119.e30,
  2024{\natexlab{b}}.

\bibitem[Sanchez-Gorostiaga et~al.(2019)Sanchez-Gorostiaga, Baji{\'c}, Osborne,
  Poyatos, and Sanchez]{Sanchez-GorostiagaBOPS19}
A.~Sanchez-Gorostiaga, D.~Baji{\'c}, M.L. Osborne, J.F. Poyatos, and
  A.~Sanchez.
\newblock High-order interactions distort the functional landscape of microbial
  consortia.
\newblock \emph{PLoS Biol.}, 17\penalty0 (12):\penalty0 e3000550, 2019.

\bibitem[Brouillet et~al.(2015)Brouillet, Annoni, Ferretti, and
  Achaz]{BrouilletAFA15}
S~Brouillet, H~Annoni, L~Ferretti, and G~Achaz.
\newblock {MAGELLAN}: a tool to explore small fitness landscapes.
\newblock \emph{bioRxiv}, 2015:\penalty0 031583, 2015.

\bibitem[Kauffman and Levin(1987)]{KauffmanL87}
Stuart Kauffman and Simon Levin.
\newblock Towards a general theory of adaptive walks on rugged landscapes.
\newblock \emph{J. Theor. Biol.}, 128:\penalty0 11--45, 1987.

\bibitem[Rao et~al.(2019)Rao, Bhattacharya, Thomas, Duan, Chen, Canny, Abbeel,
  and Song]{RaoBTDCCAS19}
Roshan Rao, Nicholas Bhattacharya, Neil Thomas, Yan Duan, Xi~Chen, John~F.
  Canny, Pieter Abbeel, and Yun~S. Song.
\newblock Evaluating protein transfer learning with {TAPE}.
\newblock In \emph{{NeurIPS}'19: Proc. of Advances in Neural Information
  Processing Systems 32}, pages 9686--9698, 2019.

\bibitem[Dallago et~al.(2021)Dallago, Mou, Johnston, Wittmann, Bhattacharya,
  Goldman, Madani, and Yang]{DallagoMJWBGMY21}
Christian Dallago, Jody Mou, Kadina~E. Johnston, Bruce~J. Wittmann, Nicholas
  Bhattacharya, Samuel Goldman, Ali Madani, and Kevin~K. Yang.
\newblock {FLIP:} benchmark tasks in fitness landscape inference for proteins.
\newblock In \emph{Proc. of the Neural Information Processing Systems Track on
  Datasets and Benchmarks 1}, 2021.

\bibitem[Wu et~al.(2016)Wu, Dai, Olson, Lloyd-Smith, and Sun]{WuDOLS16}
Nicholas~C Wu, Lei Dai, C~Anders Olson, James~O Lloyd-Smith, and Ren Sun.
\newblock Adaptation in protein fitness landscapes is facilitated by indirect
  paths.
\newblock \emph{eLife}, 5:\penalty0 e16965, 2016.

\bibitem[Hopf et~al.(2017)Hopf, Ingraham, Poelwijk, Schärfe, Springer, Sander,
  and Marks]{HopfIPS3SM17}
T.~A. Hopf, J.~B. Ingraham, F.~J. Poelwijk, C.~P. Schärfe, M.~Springer,
  C.~Sander, and D.~S. Marks.
\newblock Mutation effects predicted from sequence co-variation.
\newblock \emph{Nat. Biotechnol.}, 35\penalty0 (2):\penalty0 128--135, 2017.

\bibitem[Riesselman et~al.(2018)Riesselman, Ingraham, and
  Marks]{RiesselmanIM18}
A.~J. Riesselman, J.~B. Ingraham, and D.~S. Marks.
\newblock Deep generative models of genetic variation capture the effects of
  mutations.
\newblock \emph{Nat. Methods}, 15\penalty0 (10):\penalty0 816--822, 2018.

\bibitem[Sumi et~al.(2024)Sumi, Hamada, and Saito]{SumiHS24}
S.~Sumi, M.~Hamada, and H.~Saito.
\newblock Deep generative design of {RNA} family sequences.
\newblock \emph{Nat. Methods}, 21\penalty0 (3):\penalty0 435--443, 2024.

\bibitem[Nguyen et~al.(2024)Nguyen, Poli, Durrant, Kang, Katrekar, Li, Bartie,
  Thomas, King, Brixi, Sullivan, Ng, Lewis, Lou, Ermon, Baccus,
  Hernandez-Boussard, Ré, Hsu, and Hie]{NguyenPDKLLTTKBSNLALERBHH24}
E.~Nguyen, M.~Poli, M.~G. Durrant, B.~Kang, D.~Katrekar, D.~B. Li, L.~J.
  Bartie, A.~W. Thomas, S.~H. King, G.~Brixi, J.~Sullivan, M.~Y. Ng, A.~Lewis,
  A.~Lou, S.~Ermon, S.~A. Baccus, T.~Hernandez-Boussard, C.~Ré, P.~D. Hsu, and
  B.~L. Hie.
\newblock Sequence modeling and design from molecular to genome scale with
  {Evo}.
\newblock \emph{Science}, 386\penalty0 (6723):\penalty0 eado9336, 2024.

\bibitem[Ren et~al.(2024)Ren, Chen, Qiao, Jing, Cai, Xu, Ye, Ma, Sun, Yan,
  Yuan, Ouyang, and Liu]{RenCQJCSYMSYYOYL24}
Yuchen Ren, Zhiyuan Chen, Lifeng Qiao, Hongtai Jing, Yuchen Cai, Sheng Xu, Peng
  Ye, Xinzhu Ma, Siqi Sun, Hongliang Yan, Dong Yuan, Wanli Ouyang, and Xihui
  Liu.
\newblock {BEACON}: benchmark for comprehensive {RNA} tasks and language
  models.
\newblock \emph{Proc. of the Neural Information Processing Systems Track on
  Datasets and Benchmarks}, 2024.

\bibitem[Hansen et~al.(2021)Hansen, Auger, Ros, Mersmann, Tu{\v s}ar, and
  Brockhoff]{HansenARMTCB21}
N.~Hansen, A.~Auger, R.~Ros, O.~Mersmann, T.~Tu{\v s}ar, and D.~Brockhoff.
\newblock {COCO}: a platform for comparing continuous optimizers in a black-box
  setting.
\newblock \emph{Optim. Methods Softw.}, 36\penalty0 (1):\penalty0 114--144,
  2021.

\bibitem[Kerschke and Trautmann(2019{\natexlab{a}})]{KerschkeT19}
Pascal Kerschke and Heike Trautmann.
\newblock Comprehensive feature-based landscape analysis of continuous and
  constrained optimization problems using the {R}-package {flacco}.
\newblock pages 93--123, 2019{\natexlab{a}}.

\bibitem[Liefooghe et~al.(2020)Liefooghe, Daolio, Verel, Derbel, Aguirre, and
  Tanaka]{LiefoogheDVDAT20}
A.~Liefooghe, F.~Daolio, S.~Verel, B.~Derbel, H.~Aguirre, and K.~Tanaka.
\newblock Landscape-aware performance prediction for evolutionary
  multiobjective optimization.
\newblock \emph{{IEEE} Trans. Evol. Comput.}, 24\penalty0 (6):\penalty0
  1063--1077, 2020.

\bibitem[Kerschke and Trautmann(2019{\natexlab{b}})]{KerschkeT19b}
Pascal Kerschke and Heike Trautmann.
\newblock Automated algorithm selection on continuous black-box problems by
  combining exploratory landscape analysis and machine learning.
\newblock \emph{Evol. Comput.}, 27\penalty0 (1):\penalty0 99--127,
  2019{\natexlab{b}}.

\bibitem[Kerschke et~al.(2019)Kerschke, Hoos, Neumann, and
  Trautmann]{KerschkeHNT19}
Pascal Kerschke, Holger~H. Hoos, Frank Neumann, and Heike Trautmann.
\newblock Automated algorithm selection: Survey and perspectives.
\newblock \emph{Evol. Comput.}, 27\penalty0 (1):\penalty0 3--45, 2019.

\bibitem[Aita et~al.(2001)Aita, Iwakura, and Husimi]{AitaIH01}
T.~Aita, M.~Iwakura, and Y.~Husimi.
\newblock A cross-section of the fitness landscape of dihydrofolate reductase.
\newblock \emph{Protein Eng.}, 14\penalty0 (9):\penalty0 633--638, 2001.

\bibitem[Weinberger(1990)]{Weinberger90}
Edward Weinberger.
\newblock Correlated and uncorrelated fitness landscapes and how to tell the
  difference.
\newblock \emph{Biol. Cybern.}, 63\penalty0 (5):\penalty0 325--336, 1990.

\bibitem[Ferretti et~al.(2016)Ferretti, Schmiegelt, Weinreich, Yamauchi,
  Kobayashi, Tajima, and Achaz]{FerrettiSWYKTA16}
Luca Ferretti, Bastian Schmiegelt, Daniel Weinreich, Atsushi Yamauchi, Yutaka
  Kobayashi, Fumio Tajima, and Guillaume Achaz.
\newblock Measuring epistasis in fitness landscapes: the correlation of fitness
  effects of mutations.
\newblock \emph{J. Theor. Biol.}, 396:\penalty0 132--143, 2016.

\bibitem[Wagner(2023)]{Wagner23}
Andreas Wagner.
\newblock Evolvability-enhancing mutations in the fitness landscapes of an
  {RNA} and a protein.
\newblock \emph{Nat. Commun.}, 14\penalty0 (1):\penalty0 3624, 2023.

\bibitem[Phillips(2008)]{Phillips08}
P.~C. Phillips.
\newblock Epistasis--the essential role of gene interactions in the structure
  and evolution of genetic systems.
\newblock \emph{Nat. Rev. Genet.}, 9\penalty0 (11):\penalty0 855--867, 2008.

\bibitem[Johnson et~al.(2019)Johnson, Martsul, Kryazhimskiy, and
  Desai]{JohnsonMKD19}
Matthew~S. Johnson, Aleksej Martsul, Sergey Kryazhimskiy, and Michael~M. Desai.
\newblock Higher-fitness yeast genotypes are less robust to deleterious
  mutations.
\newblock \emph{Science}, 366\penalty0 (6464):\penalty0 490--493, 2019.

\bibitem[Jones and Forrest(1995)]{JonesF95}
T.~Jones and S.~Forrest.
\newblock Fitness distance correlation as a measure of problem difficulty for
  genetic algorithms.
\newblock \emph{Proc. Int. Conf. Genet. Algorithms}, 6:\penalty0 184--192,
  1995.

\bibitem[Payne and Wagner(2015)]{PayneW15}
Joshua~L. Payne and Andreas Wagner.
\newblock Mechanisms of mutational robustness in transcriptional regulation.
\newblock \emph{Front. Genet.}, 6:\penalty0 322, 2015.

\bibitem[Jalal et~al.(2020)Jalal, Tran, Stevenson, Chan, Lo, Tan, Noy, Lawson,
  and Le]{JalalTSCLTNLL20}
A.~S.~B. Jalal, N.~T. Tran, C.~E. Stevenson, E.~W. Chan, R.~Lo, X.~Tan, A.~Noy,
  D.~M. Lawson, and T.~B.~K. Le.
\newblock Diversification of {DNA}-binding specificity by permissive and
  specificity-switching mutations in the {ParB/Noc} protein family.
\newblock \emph{Cell Rep.}, 32\penalty0 (3):\penalty0 107928, 2020.

\bibitem[Tu et~al.(2022)Tu, Sundar, and Esvelt]{TuSE22}
Boqiang Tu, Vikram Sundar, and Kevin~M. Esvelt.
\newblock An ultra-high-throughput method for measuring biomolecular
  activities.
\newblock \emph{bioRxiv}, 2022.

\bibitem[Kauffman(1993)]{Kauffman93}
Stuart~A. Kauffman.
\newblock \emph{The origins of order: self organization and selection in
  evolution}.
\newblock Oxford Univ Press, New York, 1993.

\bibitem[Weinberger(1991)]{Weinberger91}
Edward~D. Weinberger.
\newblock Local properties of {Kauffman}'s {N-k} model: a tunably rugged energy
  landscape.
\newblock \emph{Phys. Rev. A}, 44\penalty0 (10):\penalty0 6399--6413, 1991.

\bibitem[Smith{-}Miles and Mu{\~{n}}oz(2023)]{SmithMilesM23}
Kate Smith{-}Miles and Mario~Andr{\'{e}}s Mu{\~{n}}oz.
\newblock Instance space analysis for algorithm testing: Methodology and
  software tools.
\newblock \emph{{ACM} Comput. Surv.}, 55\penalty0 (12):\penalty0 255:1--255:31,
  2023.

\bibitem[Hsu et~al.(2022)Hsu, Nisonoff, Fannjiang, and Listgarten]{HsuNFL22}
C.~Hsu, H.~Nisonoff, C.~Fannjiang, and J.~Listgarten.
\newblock Learning protein fitness models from evolutionary and assay-labeled
  data.
\newblock \emph{Nat. Biotechnol.}, 40\penalty0 (7):\penalty0 1114--1122, 2022.

\bibitem[P. et~al.(2020)P., M., C., P., M., P., D., R., S., M., and
  R.]{RussFSSBKHHCWR20}
Russ~W. P., Figliuzzi M., Stocker C., Barrat-Charlaix P., Socolich M., Kast P.,
  Hilvert D., Monasson R., Cocco S., Weigt M., and Ranganathan R.
\newblock An evolution-based model for designing chorismate mutase enzymes.
\newblock \emph{Science}, 369\penalty0 (6502):\penalty0 440--445, 2020.

\bibitem[Rives et~al.(2021)Rives, Meier, Sercu, Goyal, Lin, Liu, Guo, Ott,
  Zitnick, Ma, and Fergus]{RivesMSGGLLGOZMF21}
A~Rives, J~Meier, T~Sercu, S~Goyal, Z~Lin, J~Liu, D~Guo, M~Ott, C~L Zitnick,
  J~Ma, and R~Fergus.
\newblock Biological structure and function emerge from scaling unsupervised
  learning to 250 million protein sequences.
\newblock \emph{Proc. Natl. Acad. Sci. U. S. A.}, 118\penalty0 (15):\penalty0
  e2016239118, 2021.

\bibitem[Tsuboyama et~al.(2023)Tsuboyama, Dauparas, Chen, Laine, Behbahani,
  Weinstein, Mangan, Ovchinnikov, and Rocklin]{TsuboyamaDCLMMWMO23}
K.~Tsuboyama, J.~Dauparas, J.~Chen, E.~Laine, Y.~Mohseni Behbahani, J.~J.
  Weinstein, N.~M. Mangan, S.~Ovchinnikov, and G.~J. Rocklin.
\newblock Mega-scale experimental analysis of protein folding stability in
  biology and design.
\newblock \emph{Nature}, 620\penalty0 (7973):\penalty0 434--444, 2023.

\bibitem[Notin et~al.(2023{\natexlab{b}})Notin, Weitzman, Marks, and
  Gal]{NotinWMG23}
P.~Notin, R.~Weitzman, D.~Marks, and Y.~Gal.
\newblock Proteinnpt: improving protein property prediction and design with
  non-parametric transformers.
\newblock \emph{Adv. Neural Inf. Process. Syst.}, 36:\penalty0 33529--33563,
  2023{\natexlab{b}}.

\bibitem[Packer and Liu(2015)]{PackerL15}
M.~S. Packer and D.~R. Liu.
\newblock Methods for the directed evolution of proteins.
\newblock \emph{Nat. Rev. Genet.}, 16\penalty0 (7):\penalty0 379--394, 2015.

\bibitem[Wang et~al.(2021)Wang, Xue, Cao, Yu, Lane, and Zhao]{WangXCYLZ21}
Y.~Wang, P.~Xue, M.~Cao, T.~Yu, S.~T. Lane, and H.~Zhao.
\newblock Directed evolution: methodologies and applications.
\newblock \emph{Chem. Rev.}, 121\penalty0 (20):\penalty0 12384--12444, 2021.

\bibitem[Wittmann et~al.(2021)Wittmann, Yue, and Arnold]{WittmannYA21}
B.~J. Wittmann, Y.~Yue, and F.~H. Arnold.
\newblock Informed training set design enables efficient machine
  learning-assisted directed protein evolution.
\newblock \emph{Cell Syst.}, 12\penalty0 (11):\penalty0 1026--1045.e7, 2021.

\bibitem[Romero et~al.(2013)Romero, Krause, and Arnold]{RomeroKA13}
P.~A. Romero, A.~Krause, and F.~H. Arnold.
\newblock Navigating the protein fitness landscape with gaussian processes.
\newblock \emph{Proc. Natl. Acad. Sci. U. S. A.}, 110\penalty0 (3):\penalty0
  E193--E201, 2013.

\bibitem[Yang et~al.(2019)Yang, Wu, and Arnold]{YangWA19}
K.~K. Yang, Z.~Wu, and F.~H. Arnold.
\newblock Machine-learning-guided directed evolution for protein engineering.
\newblock \emph{Nat. Methods}, 16\penalty0 (8):\penalty0 687--694, 2019.

\bibitem[Wu et~al.(2019)Wu, Kan, Lewis, Wittmann, and Arnold]{WuKLWA19}
Z.~Wu, S.~B.~J. Kan, R.~D. Lewis, B.~J. Wittmann, and F.~H. Arnold.
\newblock Machine learning-assisted directed protein evolution with
  combinatorial libraries.
\newblock \emph{Proc. Natl. Acad. Sci. U. S. A.}, 116\penalty0 (18):\penalty0
  8852--8858, 2019.

\bibitem[Qiu et~al.(2021)Qiu, Hu, and Wei]{QiuHW21}
Y.~Qiu, J.~Hu, and G.~W. Wei.
\newblock Cluster learning-assisted directed evolution.
\newblock \emph{Nat. Comput. Sci.}, 1\penalty0 (12):\penalty0 809--818, 2021.

\bibitem[Yang et~al.(2025)Yang, Lal, Bowden, Astudillo, Hameedi, Kaur, Hill,
  Yue, and Arnold]{YangLBHAKHYA25}
J.~Yang, R.~G. Lal, J.~C. Bowden, R.~Astudillo, M.~A. Hameedi, S.~Kaur,
  M.~Hill, Y.~Yue, and F.~H. Arnold.
\newblock Active learning-assisted directed evolution.
\newblock \emph{Nat. Commun.}, 16\penalty0 (1):\penalty0 714, 2025.

\bibitem[Pinheiro et~al.(2021)Pinheiro, Warsi, Andersson, and
  Lässig]{PinheiroWAL21}
F.~Pinheiro, O.~Warsi, D.~I. Andersson, and M.~Lässig.
\newblock Metabolic fitness landscapes predict the evolution of antibiotic
  resistance.
\newblock \emph{Nat. Ecol. Evol.}, 5\penalty0 (5):\penalty0 677--687, 2021.

\bibitem[Hofacker et~al.(1994)Hofacker, Fontana, Stadler, Bonhoeffer, Tacker,
  and Schuster]{HofackerEFSTBS94}
I.~L. Hofacker, W.~Fontana, P.~F. Stadler, L.~S. Bonhoeffer, M.~Tacker, and
  P.~Schuster.
\newblock Fast folding and comparison of {RNA} secondary structures.
\newblock \emph{Monatsh. Chem.}, 125:\penalty0 167--188, 1994.

\bibitem[Dill(1985)]{Dill85}
K.~A. Dill.
\newblock Theory for the folding and stability of globular proteins.
\newblock \emph{Biochemistry}, 24\penalty0 (6):\penalty0 1501--1509, 1985.

\bibitem[Lau and Dill(1989)]{LauD89}
K.~F. Lau and K.~A. Dill.
\newblock A lattice statistical mechanics model of the conformational and
  sequence spaces of proteins.
\newblock \emph{Macromol.}, 22\penalty0 (10):\penalty0 3986--3997, 1989.

\bibitem[Greenbury et~al.(2014)Greenbury, Johnston, Louis, and
  Ahnert]{GreenburyJLA14}
S.~F. Greenbury, I.~G. Johnston, A.~A. Louis, and S.~E. Ahnert.
\newblock A tractable genotype-phenotype map modelling the self-assembly of
  protein quaternary structure.
\newblock \emph{J. R. Soc. Interface}, 11\penalty0 (95):\penalty0 20140249,
  2014.

\bibitem[Altman and Krzywinski(2018)]{AltmanK18}
N.~Altman and M.~Krzywinski.
\newblock The curse(s) of dimensionality.
\newblock \emph{Nat. Methods}, 15\penalty0 (6):\penalty0 399--400, 2018.

\bibitem[Jolliffe and Cadima(2016)]{JolliffeC16}
I.~T. Jolliffe and J.~Cadima.
\newblock Principal component analysis: a review and recent developments.
\newblock \emph{Philos. Trans. A Math. Phys. Eng. Sci.}, 374\penalty0
  (2065):\penalty0 20150202, 2016.

\bibitem[van~der Maaten and Hinton(2008)]{vandermaatenH08}
Laurens van~der Maaten and Geoffrey Hinton.
\newblock Visualizing data using t-{SNE}.
\newblock \emph{J. Mach. Learn. Res.}, 9\penalty0 (86):\penalty0 2579--2605,
  2008.

\bibitem[McInnes and Healy(2018)]{McInnesH18}
Leland McInnes and John Healy.
\newblock {UMAP:} uniform manifold approximation and projection for dimension
  reduction.
\newblock \emph{CoRR}, abs/1802.03426, 2018.

\bibitem[Zhang et~al.(2024{\natexlab{a}})Zhang, Zemke, Armand, and
  Ren]{ZhangZAR24}
K.~Zhang, N.~R. Zemke, E.~J. Armand, and B.~Ren.
\newblock A fast, scalable and versatile tool for analysis of single-cell omics
  data.
\newblock \emph{Nat. Methods}, 21\penalty0 (2):\penalty0 217--227,
  2024{\natexlab{a}}.

\bibitem[Zhou and Troyanskaya(2021)]{ZhouT21}
J.~Zhou and O.~G. Troyanskaya.
\newblock An analytical framework for interpretable and generalizable
  single-cell data analysis.
\newblock \emph{Nat. Methods}, 18\penalty0 (11):\penalty0 1317--1321, 2021.

\bibitem[Heumos et~al.(2024)Heumos, Ehmele, Treis, Belzen, Roellin, May,
  Namsaraeva, Horlava, Shitov, Zhang, Zappia, Knoll, Lang, Hetzel, Virshup,
  Sikkema, Curion, Eils, Schiller, Hilgendorff, and
  Theis]{HeumosETUZBRMNHSZZKLVSCHEST24}
L.~Heumos, P.~Ehmele, T.~Treis, J.~Upmeier~Zu Belzen, E.~Roellin, L.~May,
  A.~Namsaraeva, N.~Horlava, V.~A. Shitov, X.~Zhang, L.~Zappia, R.~Knoll, N.~J.
  Lang, L.~Hetzel, I.~Virshup, L.~Sikkema, F.~Curion, R.~Eils, H.~B. Schiller,
  A.~Hilgendorff, and F.~J. Theis.
\newblock An open-source framework for end-to-end analysis of electronic health
  record data.
\newblock \emph{Nat. Med.}, 30\penalty0 (11):\penalty0 3369--3380, 2024.

\bibitem[Keller et~al.(2025)Keller, Gold, McCallum, Manz, Kharchenko, and
  Gehlenborg]{KellerGMMKGN25}
M.~S. Keller, I.~Gold, C.~McCallum, T.~Manz, P.~V. Kharchenko, and
  N.~Gehlenborg.
\newblock Vitessce: integrative visualization of multimodal and spatially
  resolved single-cell data.
\newblock \emph{Nat. Methods}, 22\penalty0 (1):\penalty0 63--67, 2025.

\bibitem[Huang et~al.(2025)Huang, Zhou, Chen, and Li]{HuangZCL25}
Mingyu Huang, Shasha Zhou, Yuxin Chen, and Ke~Li.
\newblock Conversational exploration of literature landscape with litchat.
\newblock In \emph{{IJCAI}'25: Proc. of the 34th International Joint Conference
  on Artificial Intelligence}, volume in press, 2025.

\bibitem[Huang and Li(2024)]{HuangL24}
Mingyu Huang and Ke~Li.
\newblock A survey of decomposition-based evolutionary multi-objective
  optimization: Part {II} - {A} data science perspective.
\newblock \emph{CoRR}, abs/2404.14228, 2024.

\bibitem[Li and Li(2024)]{LiL24}
Ke~Li and Fan Li.
\newblock Multi-fidelity methods for optimization: {A} survey.
\newblock \emph{CoRR}, abs/2402.09638, 2024.

\bibitem[Wales(2018)]{Wales18}
D.~J. Wales.
\newblock Exploring energy landscapes.
\newblock \emph{Annu. Rev. Phys. Chem.}, 69:\penalty0 401--425, 2018.

\bibitem[Huang and Li(2023)]{HuangL23}
Mingyu Huang and Ke~Li.
\newblock Exploring structural similarity in fitness landscapes via graph data
  mining: {A} case study on number partitioning problems.
\newblock In \emph{{IJCAI}'23: Proc. of the 32nd International Joint Conference
  on Artificial Intelligence}, pages 5595--5603, 2023.

\bibitem[Chitra et~al.(2025)Chitra, Arnold, and Raphael]{ChitraAR25}
U.~Chitra, B.~Arnold, and B.~J. Raphael.
\newblock Resolving discrepancies between chimeric and multiplicative measures
  of higher-order epistasis.
\newblock \emph{Nat. Commun.}, 16\penalty0 (1):\penalty0 1711, 2025.

\bibitem[Bank et~al.(2016)Bank, Matuszewski, Hietpas, and Jensen]{BankMHJ16}
C.~Bank, S.~Matuszewski, R.~T. Hietpas, and J.~D. Jensen.
\newblock On the (un)predictability of a large intragenic fitness landscape.
\newblock \emph{Proc. Natl. Acad. Sci. U. S. A.}, 113\penalty0 (49):\penalty0
  14085--14090, 2016.

\bibitem[Fahlberg et~al.(2024)Fahlberg, Freschlin, Heinzelman, and
  Romero]{FahlbergFHR24}
S.~A. Fahlberg, C.~R. Freschlin, P.~Heinzelman, and P.~A. Romero.
\newblock Neural network extrapolation to distant regions of the protein
  fitness landscape.
\newblock \emph{Nat. Commun.}, 15\penalty0 (1):\penalty0 6405, 2024.

\bibitem[Phillips et~al.(2021)Phillips, Lawrence, Moulana, Dupic, Chang,
  Johnson, Cvijovic, Mora, Walczak, and Desai]{PhillipsLMDCJCMWD21}
Angela~M. Phillips, Katherine~R. Lawrence, Alief Moulana, Thomas Dupic, Jeffrey
  Chang, Milo~S. Johnson, Ivana Cvijovic, Thierry Mora, Aleksandra~M. Walczak,
  and Michael~M. Desai.
\newblock Binding affinity landscapes constrain the evolution of broadly
  neutralizing anti-influenza antibodies.
\newblock \emph{eLife}, 10:\penalty0 e71393, 2021.

\bibitem[Soo et~al.(2021)Soo, Swadling, Faure, and Warnecke]{SooSFW21}
V.~W.~C. Soo, J.~B. Swadling, A.~J. Faure, and T.~Warnecke.
\newblock Fitness landscape of a dynamic {RNA} structure.
\newblock \emph{PLoS Genet.}, 17\penalty0 (2):\penalty0 e1009353, 2021.

\bibitem[Clark et~al.(2021)Clark, Connors, and Stevenson]{ClarkCS21}
R.~L. Clark, B.~M. Connors, and D.~M. Stevenson.
\newblock Design of synthetic human gut microbiome assembly and butyrate
  production.
\newblock \emph{Nat. Commun.}, 12:\penalty0 3254, 2021.

\bibitem[Kauffman and Weinberger(1989)]{KauffmanW89}
Stuart~A. Kauffman and Edward~D. Weinberger.
\newblock The {NK} model of rugged fitness landscapes and its application to
  maturation of the immune response.
\newblock \emph{J. Theor. Biol.}, 141\penalty0 (2):\penalty0 211--245, 1989.

\bibitem[de~Visser and Krug(2014)]{VisserK14}
J.~Arjan G.~M. de~Visser and Joachim Krug.
\newblock Empirical fitness landscapes and the predictability of evolution.
\newblock \emph{Nat. Rev. Genet.}, 15\penalty0 (7):\penalty0 480--490, 2014.

\bibitem[Szendro et~al.(2013)Szendro, Schenk, Franke, Krug, and
  de~Visser]{SzendroSFEV13}
Ivan~G. Szendro, Martijn~F. Schenk, Jasper Franke, Joachim Krug, and J.~Arjan
  G.~M. de~Visser.
\newblock Quantitative analyses of empirical fitness landscapes.
\newblock \emph{J. Stat. Mech. Theory Exp.}, 2013\penalty0 (01):\penalty0
  P01005, 2013.

\bibitem[Carneiro and Hartl(2010)]{CarneiroH10}
Maur{\'i}cio Carneiro and Daniel~L. Hartl.
\newblock Adaptive landscapes and protein evolution.
\newblock \emph{Proc. Natl. Acad. Sci. U. S. A.}, 107\penalty0 (Suppl.
  1):\penalty0 1747--1751, 2010.

\bibitem[Gillespie(1984)]{Gillespie84}
John~H. Gillespie.
\newblock Molecular evolution over the mutational landscape.
\newblock \emph{Evolution}, 38\penalty0 (5):\penalty0 1116--1129, 1984.

\bibitem[Blum and Roli(2003)]{BlumR03}
Christian Blum and Andrea Roli.
\newblock Metaheuristics in combinatorial optimization: overview and conceptual
  comparison.
\newblock \emph{ACM Comput. Surv.}, 35\penalty0 (3):\penalty0 268--308, 2003.

\bibitem[Whitley et~al.(2013)Whitley, Howe, and Hains]{WhitleyHH13}
L.~Darrell Whitley, Adele~E. Howe, and Doug Hains.
\newblock Greedy or not? best improving versus first improving stochastic local
  search for {MAXSAT}.
\newblock In \emph{{AAAI}'13: Proc. of the 27th {AAAI} Conference on Artificial
  Intelligence}, pages 940--946, 2013.

\bibitem[Payne and Wagner(2019)]{PayneW19}
Joshua~L. Payne and Andreas Wagner.
\newblock The causes of evolvability and their evolution.
\newblock \emph{Nat. Rev. Genet.}, 20\penalty0 (1):\penalty0 24--38, 2019.

\bibitem[Cariani(2002)]{Cariani02}
Peter~A. Cariani.
\newblock Extradimensional bypass.
\newblock \emph{Biosyst.}, 64\penalty0 (1--3):\penalty0 47--53, 2002.

\bibitem[Kryazhimskiy et~al.(2014)Kryazhimskiy, Rice, Jerison, and
  Desai]{KryazhimskiyRJD14}
Sergey Kryazhimskiy, Daniel~P. Rice, Eric~R. Jerison, and Michael~M. Desai.
\newblock Global epistasis makes adaptation predictable despite sequence-level
  stochasticity.
\newblock \emph{Science}, 344\penalty0 (6191):\penalty0 1519--1522, 2014.

\bibitem[Romero and Arnold(2009)]{RomeroA09}
Philip~A. Romero and Frances~H. Arnold.
\newblock Exploring protein fitness landscapes by directed evolution.
\newblock \emph{Nat. Rev. Mol. Cell Biol.}, 10\penalty0 (12):\penalty0
  866--876, 2009.

\bibitem[Lyons et~al.(2020)Lyons, Zou, Xu, and Zhang]{LyonsZXZ20}
Daniel~M. Lyons, Zhengting Zou, Haiqing Xu, and Jianzhi Zhang.
\newblock Idiosyncratic epistasis creates universals in mutational effects and
  evolutionary trajectories.
\newblock \emph{Nat. Ecol. Evol.}, 4:\penalty0 1685--1693, 2020.

\bibitem[Aguirre et~al.(2023)Aguirre, Hendelman, Hutton, McCandlish, and
  Lippman]{AguirreHHML23}
Leandro Aguirre, Arielle Hendelman, Samantha~F. Hutton, David~M. McCandlish,
  and Zachary~B. Lippman.
\newblock Idiosyncratic and dose-dependent epistasis drives variation in tomato
  fruit size.
\newblock \emph{Science}, 382\penalty0 (6668):\penalty0 315--320, 2023.

\bibitem[Hansen and Wagner(2001)]{HansenW01}
T.~F. Hansen and G.~P. Wagner.
\newblock Modeling genetic architecture: a multilinear theory of gene
  interaction.
\newblock \emph{Theor. Popul. Biol.}, 59\penalty0 (1):\penalty0 61--86, 2001.

\bibitem[Weinreich et~al.(2013)Weinreich, Lan, Wylie, and
  Heckendorn]{WeinreichLWH13}
D.~M. Weinreich, Y.~Lan, C.~S. Wylie, and R.~B. Heckendorn.
\newblock Should evolutionary geneticists worry about higher-order epistasis?
\newblock \emph{Curr. Opin. Genet. Dev.}, 23\penalty0 (6):\penalty0 700--707,
  2013.

\bibitem[Neidhart et~al.(2014)Neidhart, Szendro, and Krug]{NeidhartSK14}
J{\"o}rg Neidhart, Istv{\'a}n~G. Szendro, and Joachim Krug.
\newblock Adaptation in tunably rugged fitness landscapes: the rough mount
  {Fuji} model.
\newblock \emph{Genetics}, 198\penalty0 (2):\penalty0 699--721, 2014.

\bibitem[Kingman(1978)]{Kingman78}
J.~F.~C. Kingman.
\newblock A simple model for the balance between selection and mutation.
\newblock \emph{J. Appl. Probab.}, 15\penalty0 (1):\penalty0 1--12, 1978.

\bibitem[Aita et~al.(2000)Aita, Uchiyama, Inaoka, Nakajima, Kokubo, and
  Husimi]{AitaUINKKHY00}
Toshio Aita, Hiroshi Uchiyama, Takashi Inaoka, Masahiro Nakajima, Tatsuya
  Kokubo, and Yoshiyuki Husimi.
\newblock Analysis of a local fitness landscape with a model of the rough {Mt.
  Fuji}-type landscape: application to prolyl endopeptidase and thermolysin.
\newblock \emph{Biopolymers}, 54\penalty0 (1):\penalty0 64--79, 2000.

\bibitem[Schmiegelt and Krug(2014)]{SchmiegeltK14}
Benedikt Schmiegelt and Joachim Krug.
\newblock Evolutionary accessibility of modular fitness landscapes.
\newblock \emph{J. Stat. Phys.}, 154\penalty0 (1):\penalty0 334--355, 2014.

\bibitem[Schuster et~al.(1994)Schuster, Fontana, Stadler, and
  Hofacker]{SchusterFSH94}
P.~Schuster, W.~Fontana, P.~F. Stadler, and I.~L. Hofacker.
\newblock From sequences to shapes and back: a case study in {RNA} secondary
  structures.
\newblock \emph{Proc. Biol. Sci.}, 255\penalty0 (1344):\penalty0 279--284,
  1994.

\bibitem[Aguirre et~al.(2011)Aguirre, Buldú, Stich, and
  Manrubia]{AguirreBSM11}
J.~Aguirre, J.~M. Buldú, M.~Stich, and S.~C. Manrubia.
\newblock Topological structure of the space of phenotypes: the case of {RNA}
  neutral networks.
\newblock \emph{PLoS One}, 6\penalty0 (10):\penalty0 e26324, 2011.

\bibitem[Greenbury et~al.(2016)Greenbury, Schaper, Ahnert, and
  Louis]{GreenburySAL16}
S.~F. Greenbury, S.~Schaper, S.~E. Ahnert, and A.~A. Louis.
\newblock Genetic correlations greatly increase mutational robustness and can
  both reduce and enhance evolvability.
\newblock \emph{PLoS Comput. Biol.}, 12\penalty0 (3):\penalty0 e1004773, 2016.

\bibitem[Johnston et~al.(2022)Johnston, Dingle, Greenbury, Camargo, Doye,
  Ahnert, and Louis]{JohnstonDGCDAL22}
I.~G. Johnston, K.~Dingle, S.~F. Greenbury, C.~Q. Camargo, J.~P.~K. Doye, S.~E.
  Ahnert, and A.~A. Louis.
\newblock Symmetry and simplicity spontaneously emerge from the algorithmic
  nature of evolution.
\newblock \emph{Proc. Natl. Acad. Sci. U. S. A.}, 119\penalty0 (11):\penalty0
  e2113883119, 2022.

\bibitem[Irbäck and Troein(2002)]{IrbackT02}
A.~Irbäck and C.~Troein.
\newblock Enumerating designing sequences in the {HP} model.
\newblock \emph{J. Biol. Phys.}, 28\penalty0 (1):\penalty0 1--15, 2002.

\bibitem[Ferrada and Wagner(2012)]{FerradaW12}
E.~Ferrada and A.~Wagner.
\newblock A comparison of genotype-phenotype maps for {RNA} and proteins.
\newblock \emph{Biophys. J.}, 102\penalty0 (8):\penalty0 1916--1925, 2012.

\bibitem[Li et~al.(1996)Li, Helling, Tang, and Wingreen]{LiHTW96}
H.~Li, R.~Helling, C.~Tang, and N.~Wingreen.
\newblock Emergence of preferred structures in a simple model of protein
  folding.
\newblock \emph{Science}, 273\penalty0 (5275):\penalty0 666--669, 1996.

\bibitem[Chen and Guestrin(2016)]{ChenG16}
Tianqi Chen and Carlos Guestrin.
\newblock Xgboost: {A} scalable tree boosting system.
\newblock In \emph{{KDD}'16: Proc. of the 22nd {ACM} {SIGKDD} International
  Conference on Knowledge Discovery and Data Mining}, pages 785--794. {ACM},
  2016.

\bibitem[Biswas et~al.(2021)Biswas, Khimulya, Alley, Esvelt, and
  Church]{BiswasKAEEC21}
S.~Biswas, G.~Khimulya, E.~C. Alley, K.~M. Esvelt, and G.~M. Church.
\newblock Low-n protein engineering with data-efficient deep learning.
\newblock \emph{Nat. Methods}, 18\penalty0 (4):\penalty0 389--396, 2021.

\bibitem[Wang et~al.(2023)Wang, Tang, Huang, Pan, Yang, Yang, Mu, and
  Yang]{WangTHPLYMY23}
Yi~Wang, Hui Tang, Lichao Huang, Lulu Pan, Lixiang Yang, Huanming Yang, Feng
  Mu, and Meng Yang.
\newblock Self-play reinforcement learning guides protein engineering.
\newblock \emph{Nat. Mach. Intell.}, 5:\penalty0 845--860, 2023.

\bibitem[Meier et~al.(2021)Meier, Rao, Verkuil, Liu, Sercu, and
  Rives]{MeierRVLSR21}
Joshua Meier, Roshan Rao, Robert Verkuil, Jason Liu, Tom Sercu, and Alexander
  Rives.
\newblock Language models enable zero-shot prediction of the effects of
  mutations on protein function.
\newblock In \emph{{NeurIPS}'21: Proc. of Advances in Neural Information
  Processing Systems 34}, pages 29287--29303, 2021.

\bibitem[Zhang et~al.(2024{\natexlab{b}})Zhang, Notin, Huang, Lozano,
  Chenthamarakshan, Marks, Das, and Tang]{ZhangNHLCMD024}
Zuobai Zhang, Pascal Notin, Yining Huang, Aur{\'{e}}lie~C. Lozano, Vijil
  Chenthamarakshan, Debora~S. Marks, Payel Das, and Jian Tang.
\newblock Multi-scale representation learning for protein fitness prediction.
\newblock In \emph{NeurIPS'24: Advances in Neural Information Processing
  Systems 38}, 2024{\natexlab{b}}.

\bibitem[Tekpinar et~al.(2024)]{Tekpinar24}
Mustafa Tekpinar et~al.
\newblock Prescott: a population aware, epistatic and structural model
  accurately 575 predicts missense effect. medrxiv.
\newblock 2024.

\bibitem[Jr. and Bepler(2023)]{TruongB23}
Timothy F.~Truong Jr. and Tristan Bepler.
\newblock Poet: {A} generative model of protein families as
  sequences-of-sequences.
\newblock In \emph{NeurIPS'23: Advances in Neural Information Processing
  Systems 36}, 2023.

\bibitem[Hayes et~al.(2025{\natexlab{b}})Hayes, Rao, Akin, Sofroniew, Oktay,
  Lin, Verkuil, Tran, Deaton, Wiggert, et~al.]{HayesRA+25}
Thomas Hayes, Roshan Rao, Halil Akin, Nicholas~J Sofroniew, Deniz Oktay, Zeming
  Lin, Robert Verkuil, Vincent~Q Tran, Jonathan Deaton, Marius Wiggert, et~al.
\newblock Simulating 500 million years of evolution with a language model.
\newblock \emph{Science}, page eads0018, 2025{\natexlab{b}}.

\bibitem[Tsishyn et~al.(2025)Tsishyn, Hermans, Pucci, and Rooman]{TsishynHP+25}
Matsvei Tsishyn, Pauline Hermans, Fabrizio Pucci, and Marianne Rooman.
\newblock Residue conservation and solvent accessibility are (almost) all you
  need for predicting mutational effects in proteins.
\newblock \emph{bioRxiv}, pages 2025--02, 2025.

\bibitem[Marquet et~al.(2022)Marquet, Heinzinger, Olenyi, Dallago, Erckert,
  Bernhofer, Nechaev, and Rost]{MarquetHO+22}
C{\'e}line Marquet, Michael Heinzinger, Tobias Olenyi, Christian Dallago, Kyra
  Erckert, Michael Bernhofer, Dmitrii Nechaev, and Burkhard Rost.
\newblock Embeddings from protein language models predict conservation and
  variant effects.
\newblock \emph{Human genetics}, 141\penalty0 (10):\penalty0 1629--1647, 2022.

\bibitem[Su et~al.(2024)Su, Han, Zhou, Shan, Zhou, and Yuan]{SuHZSZY24}
Jin Su, Chenchen Han, Yuyang Zhou, Junjie Shan, Xibin Zhou, and Fajie Yuan.
\newblock Saprot: Protein language modeling with structure-aware vocabulary.
\newblock In \emph{ICLR'24: Proc. of the Twelfth International Conference on
  Learning Representations}. OpenReview.net, 2024.

\bibitem[Notin et~al.(2022{\natexlab{a}})Notin, Van~Niekerk, Kollasch, Ritter,
  Gal, and Marks]{NotinVL22}
Pascal Notin, Lood Van~Niekerk, Aaron~W Kollasch, Daniel Ritter, Yarin Gal, and
  Debora~S Marks.
\newblock Trancepteve: Combining family-specific and family-agnostic models of
  protein sequences for improved fitness prediction.
\newblock \emph{bioRxiv}, pages 2022--12, 2022{\natexlab{a}}.

\bibitem[Zvyagin et~al.(2023)Zvyagin, Brace, Hippe, Deng, Zhang, Bohorquez,
  Clyde, Kale, Perez-Rivera, Ma, et~al.]{ZvyaginBH23}
Maxim Zvyagin, Alexander Brace, Kyle Hippe, Yuntian Deng, Bin Zhang,
  Cindy~Orozco Bohorquez, Austin Clyde, Bharat Kale, Danilo Perez-Rivera, Heng
  Ma, et~al.
\newblock Genslms: Genome-scale language models reveal sars-cov-2 evolutionary
  dynamics.
\newblock \emph{The International Journal of High Performance Computing
  Applications}, 37\penalty0 (6):\penalty0 683--705, 2023.

\bibitem[Dalla-Torre et~al.(2025)Dalla-Torre, Gonzalez, Mendoza-Revilla,
  Lopez~Carranza, Grzywaczewski, Oteri, Dallago, Trop, de~Almeida, Sirelkhatim,
  et~al.]{DallaGM23}
Hugo Dalla-Torre, Liam Gonzalez, Javier Mendoza-Revilla, Nicolas
  Lopez~Carranza, Adam~Henryk Grzywaczewski, Francesco Oteri, Christian
  Dallago, Evan Trop, Bernardo~P de~Almeida, Hassan Sirelkhatim, et~al.
\newblock Nucleotide transformer: building and evaluating robust foundation
  models for human genomics.
\newblock \emph{Nature Methods}, 22\penalty0 (2):\penalty0 287--297, 2025.

\bibitem[Rao et~al.(2021)Rao, Liu, Verkuil, Meier, Canny, Abbeel, Sercu, and
  Rives]{RaoLVMCASR21}
Roshan Rao, Jason Liu, Robert Verkuil, Joshua Meier, John~F. Canny, Pieter
  Abbeel, Tom Sercu, and Alexander Rives.
\newblock {MSA} transformer.
\newblock In \emph{ICML'21: Proc. of the 38th International Conference on
  Machine Learning}, volume 139 of \emph{Proceedings of Machine Learning
  Research}, pages 8844--8856. {PMLR}, 2021.

\bibitem[Notin et~al.(2022{\natexlab{b}})Notin, Dias, Frazer,
  Marchena{-}Hurtado, Gomez, Marks, and Gal]{NotinDFMGMG22}
Pascal Notin, Mafalda Dias, Jonathan Frazer, Javier Marchena{-}Hurtado,
  Aidan~N. Gomez, Debora~S. Marks, and Yarin Gal.
\newblock Tranception: Protein fitness prediction with autoregressive
  transformers and inference-time retrieval.
\newblock In \emph{ICML'22: Proc. of the International Conference on Machine
  Learning}, volume 162 of \emph{Proceedings of Machine Learning Research},
  pages 16990--17017. {PMLR}, 2022{\natexlab{b}}.

\bibitem[Langenheder et~al.(2010)Langenheder, Bulling, Solan, and
  Prosser]{LangenhederBSP10}
S.~Langenheder, M.T. Bulling, M.~Solan, and J.I. Prosser.
\newblock Bacterial biodiversity-ecosystem functioning relations are modified
  by environmental complexity.
\newblock \emph{PLoS One}, 5\penalty0 (5):\penalty0 e10834, 2010.

\bibitem[Kuo et~al.(2020)Kuo, Jahn, Cheng, Chen, Lee, Hollfelder, Wen, and
  Chou]{KuoJCCLHWC20}
Shih-Tsung Kuo, Regina~L. Jahn, Yu-Ju Cheng, Yi-Lin Chen, Yen-Ju Lee, Florian
  Hollfelder, Jen-Der Wen, and Han-Dung Chou.
\newblock Global fitness landscapes of the {Shine-Dalgarno} sequence.
\newblock \emph{Genome Res.}, 30\penalty0 (5):\penalty0 711--723, 2020.

\bibitem[Phillips et~al.(2023)Phillips, Maurer, Brooks, Dupic, Schmidt, and
  Desai]{PhillipsMBDSD23}
Angela~M Phillips, Daniel~P Maurer, Caelan Brooks, Thomas Dupic, Aaron~G
  Schmidt, and Michael~M Desai.
\newblock Hierarchical sequence-affinity landscapes shape the evolution of
  breadth in an anti-influenza receptor binding site antibody.
\newblock \emph{eLife}, 12:\penalty0 e83628, 2023.

\bibitem[Wong et~al.(2018)Wong, Kinney, and Krainer]{WongKK18}
Mandy~S. Wong, Justin~B. Kinney, and Adrian~R. Krainer.
\newblock {Q}uantitative activity profile and context dependence of all human
  5' splice sites.
\newblock \emph{Mol. Cell}, 71\penalty0 (6):\penalty0 1012--1026, 2018.

\bibitem[Moulana et~al.(2022)Moulana, Dupic, and Phillips]{MoulanaDP22}
Alief Moulana, Thomas Dupic, and Angela~M. Phillips.
\newblock Compensatory epistasis maintains {ACE2} affinity in {SARS-CoV-2}
  {Omicron} {BA.1}.
\newblock \emph{Nat. Commun.}, 13:\penalty0 7011, 2022.

\bibitem[Moulana et~al.(2023)Moulana, Dupic, Phillips, Chang, Roffler, Greaney,
  Starr, Bloom, and Desai]{MoulanaDPCRGSBD23}
Alief Moulana, Thomas Dupic, Angela~M. Phillips, Jeffrey Chang, Anne~A.
  Roffler, Allison~J. Greaney, Tyler~N. Starr, Jesse~D. Bloom, and Michael~M.
  Desai.
\newblock The landscape of antibody binding affinity in {SARS-CoV-2} {Omicron}
  {BA.1} evolution.
\newblock \emph{eLife}, 12:\penalty0 e83442, 2023.

\bibitem[Bendixsen et~al.(2019)Bendixsen, Collet, {\O}stman, and
  Hayden]{BendixsenCOH19}
D.P. Bendixsen, J.~Collet, B.~{\O}stman, and E.J. Hayden.
\newblock Genotype network intersections promote evolutionary innovation.
\newblock \emph{PLoS Biol.}, 17\penalty0 (5):\penalty0 e3000300, 2019.

\bibitem[Poelwijk et~al.(2019)Poelwijk, Socolich, and
  Ranganathan]{PoelwijkSR19}
F.J. Poelwijk, M.~Socolich, and R.~Ranganathan.
\newblock Learning the pattern of epistasis linking genotype and phenotype in a
  protein.
\newblock \emph{Nat. Commun.}, 10:\penalty0 4213, 2019.

\bibitem[Lite et~al.(2020)Lite, Grant, Nocedal, Littlehale, Guo, and
  Laub]{LiteGNLGL20}
Thuy-Lan~V Lite, Robert~A Grant, Isabel Nocedal, Megan~L Littlehale, Monica~S
  Guo, and Michael~T Laub.
\newblock Uncovering the basis of protein-protein interaction specificity with
  a combinatorially complete library.
\newblock \emph{eLife}, 9:\penalty0 e60924, 2020.

\bibitem[Baeza-Centurion et~al.(2019)Baeza-Centurion, Mi{\~n}ana, Schmiedel,
  Valc{\'a}rcel, and Lehner]{BaezaCenturionMSVL19}
Pablo Baeza-Centurion, Bel{\'e}n Mi{\~n}ana, J{\"o}rn~M. Schmiedel, Juan
  Valc{\'a}rcel, and Ben Lehner.
\newblock Combinatorial genetics reveals a scaling law for the effects of
  mutations on splicing.
\newblock \emph{Cell}, 176\penalty0 (3):\penalty0 549--563, 2019.

\bibitem[Schulz et~al.(2025)Schulz, Tan, Wu, and Wang]{SchulzTWW25}
S.~Schulz, T.~J.~C. Tan, N.~C. Wu, and S.~Wang.
\newblock Epistatic hotspots organize antibody fitness landscape and boost
  evolvability.
\newblock \emph{Proc. Natl. Acad. Sci. U. S. A.}, 122\penalty0 (2):\penalty0
  e2413884122, 2025.

\bibitem[Wu et~al.(2020)Wu, Otwinowski, Thompson, Nycholat, Nourmohammad, and
  Wilson]{WuOTNNW20}
N.~C. Wu, J.~Otwinowski, A.~J. Thompson, C.~M. Nycholat, A.~Nourmohammad, and
  I.~A. Wilson.
\newblock Major antigenic site {B} of human influenza {H3N2} viruses has an
  evolving local fitness landscape.
\newblock \emph{Nat. Commun.}, 11\penalty0 (1):\penalty0 1233, 2020.

\bibitem[Doud et~al.(2024)Doud, Gupta, Li, Medina, Fuente, and
  Meyer]{DoudGLMDFM24}
Michael~B. Doud, Animesh Gupta, Victor Li, Sarah~J. Medina, Caesar A. De~La
  Fuente, and Justin~R. Meyer.
\newblock Competition-driven eco-evolutionary feedback reshapes bacteriophage
  lambda’s fitness landscape and enables speciation.
\newblock \emph{Nat. Commun.}, 15\penalty0 (1):\penalty0 863, 2024.

\bibitem[Hall et~al.(2020)Hall, Robins, Williams, Rich, Calcott, Copp, Little,
  Schwörer, Evans, Patrick, and Ackerley]{HallRWRCCLSPEA20}
Kristina~R. Hall, Katherine~J. Robins, Emily~M. Williams, Matthew~H. Rich,
  Matthew~J. Calcott, Jason~N. Copp, Robert~F. Little, Raphael Schwörer,
  Gregory~B. Evans, Wayne~M. Patrick, and David~F. Ackerley.
\newblock Intracellular complexities of acquiring a new enzymatic function
  revealed by mass-randomisation of active-site residues.
\newblock \emph{Elife}, 9:\penalty0 e59081, 2020.

\bibitem[Fröhlich(2021)]{Frohlich21}
Christian Fröhlich.
\newblock \emph{On the evolvability of {OXA-48}}.
\newblock PhD thesis, UiT The Arctic University of Norway, 2021.

\bibitem[Hall et~al.(2010)Hall, Agan, and Pope]{HallAP10}
D.~W. Hall, M.~Agan, and S.~C. Pope.
\newblock Fitness epistasis among 6 biosynthetic loci in the budding yeast
  {Saccharomyces cerevisiae}.
\newblock \emph{J. Hered.}, 101\penalty0 (Suppl 1):\penalty0 S75--S84, 2010.

\bibitem[Tamer et~al.(2019)Tamer, Gaszek, Abdizadeh, Batur, Reynolds, Atilgan,
  Atilgan, and Toprak]{TamerGARARAT19}
Yusuf~T. Tamer, Izabela~K. Gaszek, Hamed Abdizadeh, Tugce~A. Batur, Kelly~A.
  Reynolds, Ali~R. Atilgan, Canan Atilgan, and Erdal Toprak.
\newblock High-order epistasis in catalytic power of dihydrofolate reductase
  gives rise to a rugged fitness landscape in the presence of trimethoprim
  selection.
\newblock \emph{Mol. Biol. Evol.}, 36\penalty0 (7):\penalty0 1533--1550, 2019.

\bibitem[Lozovsky et~al.(2021)Lozovsky, Daniels, Heffernan, Jacobus, and
  Hartl]{LozovskyDHJH21}
Eric~R. Lozovsky, R.~Frank Daniels, Gabriel~D. Heffernan, David~P. Jacobus, and
  Daniel~L. Hartl.
\newblock Relevance of higher-order epistasis in drug resistance.
\newblock \emph{Mol. Biol. Evol.}, 38\penalty0 (1):\penalty0 142--151, 2021.

\bibitem[Hall et~al.(2019)Hall, Karkare, Cooper, Bank, Cooper, and
  Moore]{HallKCBCM19}
A.~E. Hall, K.~Karkare, V.~S. Cooper, C.~Bank, T.~F. Cooper, and F.~B. Moore.
\newblock Environment changes epistasis to alter trade-offs along alternative
  evolutionary paths.
\newblock \emph{Evolution}, 73\penalty0 (10):\penalty0 2094--2105, 2019.

\bibitem[Whitlock and Bourguet(2000)]{WhitlockB00}
M.~C. Whitlock and D.~Bourguet.
\newblock Factors affecting the genetic load in {D}rosophila: synergistic
  epistasis and correlations among fitness components.
\newblock \emph{Evolution}, 54\penalty0 (5):\penalty0 1654--1660, 2000.

\bibitem[{de Visser} et~al.(2009){de Visser}, Park, and Krug]{deVisserPK09}
J.~A. {de Visser}, S.~C. Park, and J.~Krug.
\newblock Exploring the effect of sex on empirical fitness landscapes.
\newblock \emph{Am. Nat.}, 174\penalty0 (Suppl 1):\penalty0 S15--S30, 2009.

\bibitem[da~Silva et~al.(2010)da~Silva, Coetzer, Nedellec, Pastore, and
  Mosier]{daSilvaCNPM10}
J.~da~Silva, M.~Coetzer, R.~Nedellec, C.~Pastore, and D.~E. Mosier.
\newblock Fitness epistasis and constraints on adaptation in a human
  immunodeficiency virus type 1 protein region.
\newblock \emph{Genetics}, 185\penalty0 (1):\penalty0 293--303, 2010.

\bibitem[Sunden et~al.(2015)Sunden, Peck, Salzman, Ressl, and
  Herschlag]{SundenPSRH15}
Fredrik Sunden, Austin Peck, Julia Salzman, Simone Ressl, and Daniel Herschlag.
\newblock Extensive site-directed mutagenesis reveals interconnected functional
  units in the alkaline phosphatase active site.
\newblock \emph{Elife}, 4:\penalty0 e06181, 2015.

\bibitem[Anderson et~al.(2021)Anderson, Baier, Yang, and
  Tokuriki]{AndersonBYT21}
Dane~W. Anderson, Felix Baier, Guangfeng Yang, and Nobuhiko Tokuriki.
\newblock The adaptive landscape of a metallo-enzyme is shaped by
  environment-dependent epistasis.
\newblock \emph{Nat. Commun.}, 12\penalty0 (1):\penalty0 3867, 2021.

\bibitem[Mira et~al.(2015)Mira, Crona, Greene, Meza, Sturmfels, and
  Barlow]{MiraCGMSB15}
P.~M. Mira, K.~Crona, D.~Greene, J.~C. Meza, B.~Sturmfels, and M.~Barlow.
\newblock Rational design of antibiotic treatment plans: a treatment strategy
  for managing evolution and reversing resistance.
\newblock \emph{{PLoS One}}, 10\penalty0 (5):\penalty0 e0122283, 2015.

\bibitem[Meini et~al.(2015)Meini, Tomatis, Weinreich, and Vila]{MeiniTWV15}
M.~R. Meini, P.~E. Tomatis, D.~M. Weinreich, and A.~J. Vila.
\newblock Quantitative description of a protein fitness landscape based on
  molecular features.
\newblock \emph{Mol. Biol. Evol.}, 32\penalty0 (7):\penalty0 1774--1787, 2015.

\bibitem[Lozovsky et~al.(2009)Lozovsky, Chookajorn, Brown, Imwong, Shaw,
  Kamchonwongpaisan, Neafsey, Weinreich, and Hartl]{LozovskyCBISKNWH09}
E~R Lozovsky, T~Chookajorn, K~M Brown, M~Imwong, P~J Shaw, S~Kamchonwongpaisan,
  D~E Neafsey, D~M Weinreich, and D~L Hartl.
\newblock Stepwise acquisition of pyrimethamine resistance in the malaria
  parasite.
\newblock \emph{Proc. Natl. Acad. Sci. U.S.A.}, 106\penalty0 (29):\penalty0
  12025--12030, 2009.

\bibitem[Jiang et~al.(2018)Jiang, Keren, O'Loughlin, Sater, More, Loll, and
  Weinreich]{JiangKOSMLW18}
Peijie Jiang, Istvan Keren, Katherine~G. O'Loughlin, Susan~P. Sater, Shweta~S.
  More, Jayson~S. Loll, and Daniel~M. Weinreich.
\newblock Accessible mutational trajectories for the evolution of
  {Pyrimethamine} resistance in the malaria parasite {Plasmodium vivax}.
\newblock \emph{J. Mol. Evol.}, 86\penalty0 (7):\penalty0 495--508, 2018.

\bibitem[Ogbunugafor(2022)]{Ogbunugafor22}
C.~B. Ogbunugafor.
\newblock The mutation effect reaction norm (mu-rn) highlights environmentally
  dependent mutation effects and epistatic interactions.
\newblock \emph{Evolution}, 76:\penalty0 37--48, 2022.

\bibitem[Flynn et~al.(2013)Flynn, Cooper, Moore, and Cooper]{FlynnCMC13}
K.~M. Flynn, T.~F. Cooper, F.~B. Moore, and V.~S. Cooper.
\newblock {T}he environment affects epistatic interactions to alter the
  topology of an empirical fitness landscape.
\newblock \emph{{PLoS Genet.}}, 9\penalty0 (4):\penalty0 e1003426, 2013.

\bibitem[Malcolm et~al.(1990)Malcolm, Wilson, Matthews, Kirsch, and
  Wilson]{MalcolmWMKW90}
B.~A. Malcolm, K.~P. Wilson, B.~W. Matthews, J.~F. Kirsch, and A.~C. Wilson.
\newblock Ancestral lysozymes reconstructed, neutrality tested, and
  thermostability linked to hydrocarbon packing.
\newblock \emph{Nature}, 345\penalty0 (6270):\penalty0 86--89, 1990.

\bibitem[Guerrero et~al.(2019)Guerrero, Scarpino, Rodrigues, Hartl, and
  Ogbunugafor]{GuerreroSRHO19}
R.~F. Guerrero, S.~V. Scarpino, J.~V. Rodrigues, D.~L. Hartl, and C.~B.
  Ogbunugafor.
\newblock Proteostasis environment shapes higher-order epistasis operating on
  antibiotic resistance.
\newblock \emph{Genetics}, 212\penalty0 (2):\penalty0 565--575, 2019.

\end{thebibliography}

\newpage
\appendix

\renewcommand{\thefigure}{A\arabic{figure}}
\renewcommand{\thetable}{A\arabic{table}}
\renewcommand{\theequation}{A\arabic{equation}}
\setcounter{figure}{0}
\setcounter{table}{0}
\setcounter{equation}{0}

\newrefformat{appfig}{Fig.~\ref{#1}}
\newrefformat{apptab}{Table~\ref{#1}}

\clearpage
\newpage
\section*{NeurIPS Paper Checklist}
\begin{enumerate}

\item {\bf Claims}
    \item[] Question: Do the main claims made in the abstract and introduction accurately reflect the paper's contributions and scope?
    \item[] Answer: \answerYes{} 
    \item[] Justification: As stated in the abstract and introduction, the main scope of this paper is biological fitness prediction, in particular the benchmark of methods for this topic. The main contribution of this paper is the development of the \texttt{GraphFLA} framework to make performance benchmarks more interpretable and insightful by considering fitness landscape features. This paper also contributes to the collection of a new set of 155 combinatorially complete fitness landscapes across DNA, RNA, and protein.

\item {\bf Limitations}
    \item[] Question: Does the paper discuss the limitations of the work performed by the authors?
    \item[] Answer: \answerYes{} 
    \item[] Justification: We discussed the limitations of \texttt{GraphFLA} in~\pref{app:limitations}.

\item {\bf Theory assumptions and proofs}
    \item[] Question: For each theoretical result, does the paper provide the full set of assumptions and a complete (and correct) proof?
    \item[] Answer: \answerNA{} 
    \item[] Justification: The paper does not include any theoretical results.

    \item {\bf Experimental result reproducibility}
    \item[] Question: Does the paper fully disclose all the information needed to reproduce the main experimental results of the paper to the extent that it affects the main claims and/or conclusions of the paper (regardless of whether the code and data are provided or not)?
    \item[] Answer: \answerYes{} 
    \item[] Justification: All experimental setups are clearly described in either~\pref{sec:results} or in corresponding Appendix sections. 

\item {\bf Open access to data and code}
    \item[] Question: Does the paper provide open access to the data and code, with sufficient instructions to faithfully reproduce the main experimental results, as described in supplemental material?
    \item[] Answer: \answerYes{} 
    \item[] Justification: All the artifacts needed to reproduce the experiments, including code for \texttt{GraphFLA} and the 155 combinatorially complete fitness landscapes data, are available in \url{https://github.com/COLA-Laboratory/GraphFLA}. We also used previously published benchmark results from ProteinGym, which is available at \url{https://proteingym.org/}. We are currently unable to share the RNAGym data and benchmark scores without permission from the original authors.

\item {\bf Experimental setting/details}
    \item[] Question: Does the paper specify all the training and test details (e.g., data splits, hyperparameters, how they were chosen, type of optimizer, etc.) necessary to understand the results?
    \item[] Answer: \answerYes{} 
    \item[] Justification: Relevant information are available in~\pref{sec:results} and~\pref{app:de}.

\item {\bf Experiment statistical significance}
    \item[] Question: Does the paper report error bars suitably and correctly defined or other appropriate information about the statistical significance of the experiments?
    \item[] Answer: \answerYes{} 
    \item[] Justification: The experimental results are accompanied by confidence intervals and statistical significance tests.

\item {\bf Experiments compute resources}
    \item[] Question: For each experiment, does the paper provide sufficient information on the computer resources (type of compute workers, memory, time of execution) needed to reproduce the experiments?
    \item[] Answer: \answerYes{} 
    \item[] Justification: \pref{sec:versatility} provides sufficient details regarding computation hardware used as well as the runtime \& memory scalability of \texttt{GraphFLA}.

\item {\bf Code of ethics}
    \item[] Question: Does the research conducted in the paper conform, in every respect, with the NeurIPS Code of Ethics \url{https://neurips.cc/public/EthicsGuidelines}?
    \item[] Answer: \answerYes{} 
    \item[] Justification: The authors have ensured that all aspects of the research adhere to the NeurIPS Code of Ethics.

\item {\bf Broader impacts}
    \item[] Question: Does the paper discuss both potential positive societal impacts and negative societal impacts of the work performed?
    \item[] Answer: \answerNA{} 
    \item[] Justification: No negative social impact is related to this work.
    
\item {\bf Safeguards}
    \item[] Question: Does the paper describe safeguards that have been put in place for responsible release of data or models that have a high risk for misuse (e.g., pretrained language models, image generators, or scraped datasets)?
    \item[] Answer: \answerNA{} 
    \item[] Justification: The paper poses no such risks.

\item {\bf Licenses for existing assets}
    \item[] Question: Are the creators or original owners of assets (e.g., code, data, models), used in the paper, properly credited and are the license and terms of use explicitly mentioned and properly respected?
    \item[] Answer: \answerYes{} 
    \item[] Justification: The paper properly credits the creators of the assets and mentions the license and terms of use.

\item {\bf New assets}
    \item[] Question: Are new assets introduced in the paper well documented and is the documentation provided alongside the assets?
    \item[] Answer: \answerYes{} 
    \item[] Justification: All assets are available in~\url{https://github.com/COLA-Laboratory/GraphFLA}.

\item {\bf Crowdsourcing and research with human subjects}
    \item[] Question: For crowdsourcing experiments and research with human subjects, does the paper include the full text of instructions given to participants and screenshots, if applicable, as well as details about compensation (if any)? 
    \item[] Answer: \answerNA{} 
    \item[] Justification: The paper does not involve crowdsourcing nor research with human subjects

\item {\bf Institutional review board (IRB) approvals or equivalent for research with human subjects}
    \item[] Question: Does the paper describe potential risks incurred by study participants, whether such risks were disclosed to the subjects, and whether Institutional Review Board (IRB) approvals (or an equivalent approval/review based on the requirements of your country or institution) were obtained?
    \item[] Answer: \answerNA{} 
    \item[] Justification: The paper does not involve crowdsourcing nor research with human subjects

\item {\bf Declaration of LLM usage}
    \item[] Question: Does the paper describe the usage of LLMs if it is an important, original, or non-standard component of the core methods in this research? Note that if the LLM is used only for writing, editing, or formatting purposes and does not impact the core methodology, scientific rigorousness, or originality of the research, declaration is not required.
    \item[] Answer: \answerYes{} 
    \item[] Justification: The authors declare the usage of LLMs in~\pref{sec:graphfla} and \pref{app:datasets}.

\end{enumerate}

\clearpage

\section{Limitations}
\label{app:limitations}

While \texttt{GraphFLA} provides extensive quantitative features for characterizing fitness landscapes, effectively visualizing their topography remains challenging due to the inherent high dimensionality and associated curse of dimensionality~\cite{AltmanK18}. Dimensionality reduction methods, such as PCA~\cite{JolliffeC16}, t-SNE~\cite{vandermaatenH08}, and UMAP~\cite{McInnesH18}, partially address this issue and have been effectively applied to diverse biological data~\cite{ZhangZAR24,ZhouT21,HeumosETUZBRMNHSZZKLVSCHEST24,KellerGMMKGN25}. However, these methods risk generating misleading visualizations of fitness landscapes. Specifically, compressing data into fewer dimensions inevitably leads to loss of information, potentially distorting spatial relationships among variants. Although this might be acceptable for general visualization purposes—where overall data trends remain intact—such distortions can result in the incorrect identification of local optima that do not exist in the original high-dimensional space. Consequently, despite the intuitive appeal of visualizing fitness landscape topography, \texttt{GraphFLA}, along with much of the landscape analysis literature, emphasizes quantitative metrics that inherently capture patterns within high-dimensional spaces.

Additionally, despite extensive optimization efforts, the scale of landscapes analyzable by \texttt{GraphFLA} within practical computational times remains small compared to the entire genotype space. For instance, the number of potential RNA sequences of length $n=100$ is $4^{100}$, vastly exceeding the number of atoms in the observable universe. Nonetheless, by efficiently handling landscapes containing millions of variants, \texttt{GraphFLA} aligns well with current experimental capabilities, and can comfortably accommodate even the largest empirically measured fitness landscapes.

\section{Data-driven Literature Survey on Landscape Analysis}
\label{app:datasets}

Building on our prior success in leveraging data-driven methods and large language models (LLMs) to enhance literature comprehension~\cite{HuangZCL25,HuangL24,LiL24}, we adopted a similar strategy in the development of \texttt{GraphFLA}. This involved conducting an extensive literature survey designed to: (1) identify key landscape features that characterize their topography, and (2) gather empirical data for combinatorially complete fitness landscapes. The methodology encompassed several distinct stages:

\subsection{Initial Literature Collection and Filtering}

\textbf{Search Query Formulation.}
The initial step involved crafting a targeted search query to retrieve literature pertinent to fitness landscapes. This topic is central to research where landscape features and combinatorially complete datasets are developed or utilized. The formulated query was:
\begin{quote}
\textit{((``fitness landscape*'' OR ``adaptive landscape*'' OR ``genotype network'' OR ``genotype-phenotype map*'') OR (``epistasis'' OR ``diminishing return*'' OR ``increasing cost*'' OR ``NK landscape*''))}
\end{quote}
This query is structured with two primary components:
\begin{enumerate}
     \item The first component utilizes established terminology for fitness landscapes (e.g., ``fitness landscape'', ``adaptive landscape'') and associated concepts (e.g., ``genotype network'', ``genotype-phenotype mapping'') to ensure a broad capture of relevant studies.
     \item The second component augments the search by incorporating specific terms frequently used in biological landscape analysis, such as ``epistasis'' and ``NK landscape''.
\end{enumerate}
While other relevant concepts exist (e.g., local optima, $r/s$ ratio, see~\pref{app:graphfla}), terms like ``local optima'' are prevalent across diverse optimization fields, making them less specific. Similarly, terms like ``$r/s$ ratio'' can be challenging for effective textual matching. Consequently, these were excluded from the initial query. This search strategy deliberately prioritized high recall, acknowledging that it might retrieve studies from related domains (e.g., energy landscapes in physics/chemistry~\cite{Wales18}, optimization landscapes in evolutionary computation~\cite{HuangL23}). Distinguishing these fields solely through keywords is often infeasible due to their broad scopes; therefore, subsequent filtering steps were planned to refine the selection for biological relevance.

\textbf{Database Search.}
The formulated search query was executed on the Web of Science database\footnote{https://www.webofscience.com}. This platform was chosen for its comprehensive coverage of peer-reviewed literature and high-quality metadata. To enhance the precision of the initial retrieval and minimize noise from full-text searches (such as incidental mentions of keywords), the search scope was restricted to the title, abstract, and author keywords fields. These sections typically encapsulate the core subject matter of a publication.

\textbf{LLM-based Filtering.}
This initial search yielded a substantial corpus of $31,784$ potentially relevant publications. To manage this volume and efficiently identify studies most pertinent to our research scope, we employed an LLM—specifically GPT-4o-mini—for automated initial screening. The title and abstract of each publication were processed by the LLM, which was prompted to classify the study based on two primary criteria:
\begin{enumerate}
    \item Does the publication investigate fitness landscapes or closely related concepts specifically within biological systems?
    \item If the answer to the first criterion is affirmative, does the publication report on empirical data, as opposed to being a purely theoretical analysis, \textit{in-silico} simulation study, or review article?
\end{enumerate}
This LLM-driven filtering process significantly narrowed the candidate pool. After applying the first criterion, the number of papers was reduced to $11,098$. The second criterion further refined this set to $1,673$ publications. This curated collection of papers, focusing on analysis of empirically measured fitness landscapes in biological systems, formed the basis for subsequent landscape feature set construction and the collection of combinatorially complete landscape data.

\subsection{Landscape Feature Set Construction}
\label{subsec:feature_construction}

Following the identification of $1,673$ core publications relevant to empirical fitness landscape analysis, we proceeded to construct a comprehensive and representative set of landscape features. The objective was to distill a manageable yet informative collection of quantitative indicators that capture the fundamental topographical aspects of fitness landscapes.

\textbf{Initial Feature Candidate Identification.}
The full texts of the $1,673$ curated papers were systematically reviewed to identify all quantitative measures used to describe landscape topography. This extensive survey, augmented by an LLM (GPT-4o) to scan for mentions and definitions of landscape metrics, initially yielded a broad list of over $100$ candidate features. These candidates encompassed a wide range of mathematical formulations, statistical measures, and network-based properties that researchers have employed to characterize landscapes. 

\textbf{LLM-assisted Feature Filtering and Selection.}
To refine this extensive list into a practical and impactful feature set, we devised a set of carefully crafted criteria to guide selection:

\begin{enumerate}
    \item \textbf{Empirical prevalence in literature:} How frequently is the feature used or discussed in the surveyed $1,673$ papers? Features with high prevalence were prioritized as they represent established and widely accepted indicators. The number of local optima are employed in $45\%$ of our analyzed literature. 
    \item \textbf{Biological significance:} Does the feature provide meaningful insights into evolutionary processes or other biological phenomena? Features with clear connections to biological interpretations were favored. For instance, features quantifying aspects like diminishing returns epistasis can shed light on the rate of adaptation~\cite{ChouCDSM11}, while measures related to neutrality can have great biological implications for understanding mutational robustness~\cite{LauringFA13}.
    \item \textbf{Coverage across different topographical aspects:} Does the feature contribute to characterizing one of the 4 fundamental topographical aspects: ruggedness, navigability, epistasis, and neutrality? We aimed for a balanced set that provides a holistic view of the landscape.
    \item \textbf{Computational feasibility:} Can the feature be computed efficiently for landscapes of varying sizes and complexities, such as those included in \texttt{GraphFLA}? Features requiring prohibitive computational resources for typical dataset sizes were deprioritized. For example, while the Walsh-Hadamard transform~\cite{ChitraAR25} can be used to calculate a full spectrum of epistatic interactions, its computational demand can be prohibitive for large landscapes.
    \item \textbf{Compatibility with data modalities, Sizes, and Structures:} Is the feature applicable to the types of data commonly found in empirical fitness landscapes? Is it robust to missing data or variations in landscape size? Features with broad applicability and robustness were preferred.
\end{enumerate}

\textbf{Final feature set.} By manual reviewing the initial set of features along with expert consultations, we arrived at the final set of $20$ essential landscape features presented in \pref{tab:landscape_features}, which cover all 4 fundamental aspects of landscape topography and are extensively used in landscape analysis literature to offer different biological insights. They can also be efficiently computed for empirical landscapes with different modalities and sizes, and are applicable to non-complete landscapes. A full introduction to each of these $20$ features, including their definitions, interpretations, and computational considerations, is available in \pref{app:graphfla}.

\subsection{Combinatorially Complete Landscape Collection}

In \texttt{GraphFLA}, we focused on collecting combinatorially complete datasets derived from extensive mutagenesis studies. Unlike datasets generated by randomly sampling mutants around a wild-type sequence, combinatorially complete landscapes encompass measurements for \textit{all} possible genotypes within a defined genotype space. We identified and collected such datasets from our focused set of $1,673$ papers via manual scrutiny of the full-texts with the aid of GPT-4o. Specifically, for each paper, we asked:

\begin{itemize}
     \item Does the publication publishes new empirical landscape data?
     \begin{itemize}
          \item [$\to$] If yes, was the published data combinatorially complete? This criterion specifically excludes datasets focusing only on evolutionary trajectories (i.e., monitoring changes in population mean fitness and genotypic composition) or deep mutational scanning data (i.e., sampling only mutants closely related to a wild-type sequence).
          \item [$\to$] If no, did it mention or use combinatorially complete datasets from previous works?
     \end{itemize}
\end{itemize}

After this final review, we arrived at a total of $155$ datasets as listed in~\pref{tab:datasets}, sourced from more than $67$ studies. This number is much smaller compared to the initial corpus of over $30,000$ papers. The main reason for this is such combinatorially complete landscapes are extremely costly to construct. Consequently, they are regarded as extremely valuable resources and are extensively utilized in subsequent works for both deriving biological insights~\cite{BankMHJ16,Wagner23} and testing ML systems~\cite{YangLBHAKHYA25,FahlbergFHR24,DallagoMJWBGMY21}. 

\subsection{Datasets Processing.}

Here we describe a few standards we applied when preparing these data.

\textbf{Naming convention.} We established a systematic naming convention to uniquely and informatively identify each dataset within the collection, following these guidelines:
\begin{itemize}
    \item \textbf{Base identifier:} The core of the name typically consists of the first author's last name concatenated with the four-digit publication year (e.g., ``Papkou2024'').

    \item \textbf{Common name suffix:} If a dataset is widely recognized by a common identifier, often related to the specific biological system or molecule studied, this identifier may be appended as a suffix for easier recognition. For instance, the study by \citeauthor{WuDOLS16}~\cite{WuDOLS16} investigated the fitness landscape across $20^4 = 160,000$ variants at four sites (V39, D40, G41, V54) within protein G domain B1 (GB1). This landscape is commonly referred to as ``GB1,'' and thus the dataset might be named incorporating this suffix (e.g., ``Wu2016\_GB1'').

    \item \textbf{Disambiguation suffixes:} Additional suffixes are employed when a single publication or study system yields multiple distinct datasets. These suffixes serve to differentiate datasets based on key experimental variables, such as:
    \begin{itemize}
        \item Different subjects and fitness measures (e.g., \citeauthor{PhillipsLMDCJCMWD21}~\cite{PhillipsLMDCJCMWD21} studied binding affinities of different variants of antibodies CR9114 and CR6261 against various influenza HA antigens like H1, H3, etc., which results in separate datasets per antibody-antigen pair).
        \item Variations in experimental conditions or environments (e.g., \citeauthor{SooSFW21}~\cite{SooSFW21} measured the self-splicing activity of $4^8 = 65,536$ Tetrahymena intron variants at two different temperatures, 30°C and 37°C, leading to two distinct datasets).
        \item Exploration of different mutation sites or regions (e.g., \citeauthor{JohnstonAWLPYA24}~\cite{JohnstonAWLPYA24} generated the ``TrpB3D'' landscape from $20^3=8,000$ variants at sites \{T117, A118, A119\} of the thermostable tryptophan synthase $\beta$-subunit (TrpB), and the distinct ``TrpB3E'' landscape based on sites \{F184, G185, S186\}).
        \item A combination of the above factors.
    \end{itemize}
    These distinguishing characteristics are systematically incorporated into the dataset name as further suffixes to ensure clarity and uniqueness.
\end{itemize}

\textbf{Search space representation.} Genotypes within each dataset are presented using two formats: a sequence-based representation (e.g., ``ATTA'') and a vector explicitly listing the allele at each locus (e.g., [``A'', ``T'', ``T'', ``A'']). The combinatorial nature of these representations leads to vast theoretical search spaces. For a sequence of length $L$, the total number of possible genotypes is $4^L$ for DNA or RNA, $20^L$ for proteins, or $2^L$ for binary representations. Binary representations typically indicate the presence or absence of specific mutations and can be applied to DNA, RNA, proteins, or other biological systems, such as microbial communities~\cite{SkwaraGYDRRTKS23,ClarkCS21}.
Although the theoretical search spaces for our collected landscapes are combinatorially complete by design, the experimentally generated data often exhibit incomplete coverage due to experimental constraints or subsequent filtering steps. For example, the DHFR landscape from \citeauthor{PapkouRM23}~\cite{PapkouRM23} measured fitness for $261,382$ variants, which constitutes $99.7\%$ of the total $4^9 = 262,144$ possible genotypes. Similarly, the GB1 protein landscape reported by \citeauthor{WuDOLS16}~\cite{WuDOLS16} includes $149,361$ variants, covering $93.4\%$ of the theoretical space of $20^4 = 160,000$ sequences.

\textbf{Fitness measure.} Following our naming convention, each of the $155$ dataset listed in~\pref{tab:datasets} focuses on a single fitness measure under a given condition (e.g., environment). To ensure the potential for accurate landscape reconstruction and the replication of published analysis results, we have retained the original fitness values as reported in the source publications. No transformations were applied to these values unless such transformations were already part of the published dataset. For datasets where fitness variance across replications is available, this is amended as an additional column.

\section{Fitness Landscape Features}
\label{app:graphfla}

\subsection*{Core Definitions}

We begin by defining fundamental concepts used throughout the fitness landscape analysis.

\begin{definition}[Alleles and Loci]
\label{def:alleles_loci}
Let $n$ be the number of polymorphic loci under consideration. For each locus $i \in \{1, 2, \dots, n\}$, let $\mathcal{A}_i$ denote the set of distinct alleles present. The number of alleles at locus $i$ is $m_i = |\mathcal{A}_i| \ge 2$.
\end{definition}

\begin{example}
    For example, in complete DNA or RNA landscapes, $\mathcal{A}_i$ typically contains four nucleotide bases ($\{A, C, G, T\}$ for DNA, $\{A, C, G, U\}$ for RNA), thus $m_i = |\mathcal{A}_i| = 4$. For proteins, where loci usually represent amino acid positions, $\mathcal{A}_i$ comprises the 20 standard amino acids, resulting in $m_i = |\mathcal{A}_i| = 20$.
\end{example}

\begin{definition}[Genotype Space]
\label{def:genotype_space}
A genotype $g$ is a specific combination of alleles across all $n$ loci, represented as a sequence $(a_1, a_2, \dots, a_n)$ where $a_i \in \mathcal{A}_i$ for each $i$. The set of all possible genotypes constitutes the genotype space $\mathcal{G}$. The total number of genotypes is $|\mathcal{G}| = \prod_{i=1}^n m_i$.
\end{definition}

\begin{definition}[Hamming Distance]
\label{def:hamming}
The Hamming (genetic) distance $d_H(g, g')$ between two genotypes $g = (a_1, \dots, a_n)$ and $g' = (a'_1, \dots, a'_n)$ is the number of loci at which their alleles differ:
$$d_H(g, g') = \sum_{i=1}^n \mathbb{I}(a_i \neq a'_i),$$
where $\mathbb{I}(\cdot)$ is the indicator function ($\mathbb{I}(\text{condition}) = 1$ if the condition is true, $0$ otherwise).
\end{definition}

\begin{definition}[Mutational Neighborhood]
\label{def:neighborhood}
The neighborhood $\mathcal{N}(g)$ of a genotype $g \in \mathcal{G}$ is the set of all genotypes reachable from $g$ by a single point mutation, i.e., genotypes $g'$ differing from $g$ at exactly one locus:
$$\mathcal{N}(g) = \{ g' \in \mathcal{G} \mid d_H(g, g') = 1 \}.$$
The size of the neighborhood is $|\mathcal{N}(g)| = \sum_{i=1}^n (m_i - 1)$.
\end{definition}

\begin{definition}[Fitness Function]
\label{def:fitness_func}
The fitness function $f: \mathcal{G} \to \mathbb{R}$ assigns a scalar fitness value $f(g)$ to each genotype $g \in \mathcal{G}$.
\end{definition}

\begin{definition}[Mutant Genotype Notation]
\label{def:mutant_notation}
Let $g = (a_1, \dots, a_j, \dots, a_n)$. We denote the specific single-mutant neighbor resulting from changing the allele $a_j$ at locus $j$ to a different allele $a'_j \in \mathcal{A}_j \setminus \{a_j\}$ as $g_{[j \leftarrow a'_j]}$. When the specific mutated allele $a'_j$ is not critical or is clear from context (e.g., in biallelic systems where the alternative allele is unique), we may use the shorthand $g_{[j]}$ to denote \textit{any} single mutant differing from $g$ only at locus $j$.
\end{definition}

\begin{definition}[Selection Coefficient]
\label{def:selection_coeff}
Let $g$ be a genotype with allele $a_j$ at locus $j$. The selection coefficient $s_{j, a_j \to a'_j}(g)$ measures the fitness effect of mutating allele $a_j$ to $a'_j$ (where $a'_j \in \mathcal{A}_j \setminus \{a_j\}$) at locus $j$ within the genetic background of $g$:
$$s_{j, a_j \to a'_j}(g) = f(g_{[j \leftarrow a'_j]}) - f(g).$$
When the specific mutation $a_j \to a'_j$ is unambiguous or when referring generally to the effect of a mutation at locus $j$, we may use the simpler notation $s_j(g) = f(g_{[j]}) - f(g)$.
\end{definition}

\subsection{Ruggedness}

\subsubsection{Number of Local Optima}
\label{app:num_optima}

\begin{definition}[Local and Global Optima]
     \label{def:optima}
     A genotype $g^\ell \in \mathcal{G}$ is a \textbf{local optimum} if its fitness is greater than or equal to that of all its neighbors: $f(g^\ell) \ge f(g')$ for all $g' \in \mathcal{N}(g^\ell)$. A \textbf{global optimum} $g^*$ is a genotype with the maximum fitness across the entire landscape: $g^* \in \arg\max_{g \in \mathcal{G}} f(g)$. We generally assume a unique global optimum for simplicity, although the concepts readily extend to scenarios with multiple global optima.
\end{definition}

The \textit{number of local optima}, $n_\text{lo}$, serves as a primary indicator of landscape ruggedness~\cite{PapkouRM23}. A smooth landscape possesses only a single local optimum, which coincides with the global optimum. In contrast, highly rugged landscapes, such as Kauffman's $NK$ model~\cite{KauffmanW89} with $k=n-1$ interactions, can feature a large number of local optima, potentially scaling exponentially with system size (e.g., approximately $\frac{2^n}{n+1}$ local optima for certain $NK$ parameters~\cite{Weinberger91}).

In the graph-based representation employed by \texttt{GraphFLA}, where directed edges implicitly encode fitness comparisons between neighboring genotypes, local optima correspond directly to \textit{sink} nodes (i.e., nodes with no outgoing edges). These sinks can be efficiently identified using standard graph algorithms (e.g., via the \texttt{igraph} library). To facilitate comparisons between landscapes of varying sizes, the absolute number of local optima is often normalized by the total number of genotypes, $|\mathcal{G}| = \prod_{i=1}^n m_i$. This yields the dimensionless quantity \textit{fraction of local optima}.

\subsubsection{Autocorrelation}
\label{app:autocorrelation}

Autocorrelation assesses the smoothness or ruggedness of a fitness landscape by measuring the correlation between the fitness values of genotypes encountered along random walks~\cite{Weinberger90,VisserK14}. Specifically, consider a random walk $g_0, g_1, \dots, g_L$ of length $L$ through the genotype space $\mathcal{G}$, where each step $g_{t+1}$ is typically chosen uniformly at random from the neighbors $\mathcal{N}(g_t)$. The lag-$k$ autocorrelation $\rho_a(k)$ is defined as:

\begin{equation}
     \rho_a(k) = \frac{ \mathbb{E}[ (f(g_t) - \langle f \rangle)(f(g_{t+k}) - \langle f \rangle) ] }{ \text{Var}[f(g_t)] }
     \label{eq:autocorrelation}
\end{equation}

where the expectation $\mathbb{E}[\cdot]$ and variance $\text{Var}[\cdot]$ are taken over all valid time steps $t$ (and potentially multiple walks), and $\langle f \rangle$ represents the average fitness value across all considered steps $g_t$.

The primary focus is often on the lag-1 autocorrelation, $\rho_a(1)$ (simplified as $\rho_a$), which measures the fitness correlation between genotypes separated by a single mutation (Hamming distance 1).

In practice, \texttt{GraphFLA} estimates $\rho_a$ by simulating a large number ($N_{walks}$, e.g., $1,000$) of independent random walks, each of length $L$. A common choice for the walk length is $L=n$, where $n$ is the number of loci. The covariance term (numerator) and variance term (denominator) in Eq.~\ref{eq:autocorrelation} are estimated by averaging over all adjacent pairs $(g_t, g_{t+1})$ across all simulated walks. Averaging over multiple walks provides a robust estimate.

A value of $\rho_a$ close to 1 indicates a smooth landscape where fitness changes gradually between neighboring genotypes. Conversely, $\rho_a$ values close to 0 suggest a rugged landscape where the fitness of neighbors is largely uncorrelated, implying rapid and unpredictable fitness changes.

\subsubsection{Neighbor Fitness Correlation (NFC)}
\label{app:neighbor_fitness_corr}

Another intuitive measure of landscape ruggedness assesses the relationship between a genotype's fitness and the average fitness of its mutational neighbors. This evaluates the tendency for high-fitness genotypes to be surrounded by neighbors that also have high fitness, and conversely for low-fitness genotypes.

Specifically, we compute the Pearson correlation coefficient, denoted $\rho_{f, \langle f \rangle_\mathcal{N}}$, between the fitness $f(g)$ of each genotype $g \in \mathcal{G}$ and the average fitness of its neighborhood $\langle f \rangle_{\mathcal{N}(g)}$:

\begin{equation}
     \text{NFC} = \text{Cor} \left[ f(g), \langle f \rangle_{\mathcal{N}(g)} \right],
     \label{eq:neighbor_fitness_corr}
\end{equation}

where $\langle f \rangle_{\mathcal{N}(g)}$ is the mean fitness over all single-mutant neighbors of $g$:

\begin{equation}
     \langle f \rangle_{\mathcal{N}(g)} = \frac{1}{|\mathcal{N}(g)|} \sum_{g' \in \mathcal{N}(g)} f(g').
     \label{eq:avg_neighbor_fitness}
\end{equation}

The correlation in Eq.~\eqref{eq:neighbor_fitness_corr} is calculated across all genotypes $g \in \mathcal{G}$ in the landscape.

This neighbor fitness correlation provides insight into the local smoothness or ruggedness of the landscape structure:
\begin{itemize}
    \item A value of $\text{NFC} \approx 1$ indicates a relatively smooth landscape where fitness changes tend to be gradual; high-fitness genotypes are typically surrounded by other high-fitness genotypes.
    \item A value of $\text{NFC} \approx 0$ suggests a rugged landscape where a genotype's fitness provides little predictive power for its neighbors' fitness, indicating abrupt changes.
    \item A value of $\text{NFC} \approx -1$, though less common, would imply an anticorrelated or oscillatory landscape structure where high-fitness genotypes are predomiNaNtly surrounded by low-fitness neighbors, and vice versa.
\end{itemize}
This measure focuses specifically on the fitness relationship between immediate neighbors, and complements measures like autocorrelation (\ref{app:autocorrelation}) which consider correlations along walks.

\subsubsection{Roughness-Slope Ratio (\texorpdfstring{$r/s$}{r/s})}
\label{app:rs_ratio}

The roughness-slope ratio ($r/s$) quantifies the deviation of a fitness landscape from a purely additive model, thereby measuring the relative contribution of epistasis to the landscape structure \cite{SzendroSFEV13, CarneiroH10}. It is defined as the ratio of the landscape's 'roughness' ($r$), representing the magnitude of non-additive effects (residuals from an additive fit), to its 'slope' ($s$), representing the average magnitude of additive effects. A higher $r/s$ value indicates greater ruggedness and stronger relative epistasis, while $r/s = 0$ corresponds to a perfectly additive (non-epistatic) landscape.

To compute $r/s$, an additive fitness model $f^{\mathrm{add}}(g)$ is fitted to the observed fitness data $f(g)$ using ordinary least squares (OLS) regression. For multi-allelic landscapes, genotypes are typically represented using one-hot encoding. The additive model is generally specified as:

\begin{equation}
     f^{\mathrm{add}}(g) = \beta_0 + \sum_{i=1}^{n} \sum_{a \in \mathcal{A}_i'} \beta_{i,a} X_{i,a}(g),
     \label{eq:additive_model}
\end{equation}

where $X_{i,a}(g)$ is a binary indicator variable (1 if genotype $g$ has allele $a$ at locus $i$, 0 otherwise), $\beta_0$ is the intercept, and $\beta_{i,a}$ are the fitted coefficients representing the additive effect of allele $a$ at locus $i$ relative to a reference allele. The inner sum $\sum_{a \in \mathcal{A}_i'}$ typically runs over $m_i-1$ alleles for each locus $i$ (excluding one reference allele per locus, denoted implicitly by $\mathcal{A}_i'$) to ensure model identifiability.

The roughness $r$ is defined as the root-mean-square error (RMSE) between the true fitness values $f(g)$ and the fitness predicted by the additive model $f^{\mathrm{add}}(g)$:

\begin{equation}
    r = \sqrt{ \frac{1}{|\mathcal{G}|} \sum_{g \in \mathcal{G}} (f(g) - f^{\mathrm{add}}(g))^2 }.
    \label{eq:roughness}
\end{equation}

The slope $s$ is calculated as the average absolute value of the estimated additive coefficients (excluding the intercept and reference allele coefficients, which are implicitly zero or absorbed):

\begin{equation}
    s = \frac{1}{\sum_{k=1}^{n} (m_k-1)} \sum_{i=1}^{n} \sum_{a \in \mathcal{A}_i'} |\beta_{i,a}|.
    \label{eq:slope}
\end{equation}

Here, the denominator $\sum_{k=1}^{n} (m_k-1)$ represents the total number of independent additive coefficients fitted in the model (one for each non-reference allele across all loci), and the summation $\sum_{a \in \mathcal{A}_i'}$ covers these specific fitted coefficients for locus $i$.

The $r/s$ ratio thus provides a scale-independent measure comparing the magnitude of epistatic deviations to the average strength of individual additive allelic effects.

\subsubsection{Gamma statistic.} 

This measure is initially introduced in~\cite{FerrettiSWYKTA16} for di-allelic data (i.e., $m_i = 2$, e.g., genotypes encoded with ``0''s and ``1''s as is common in mutation data) and is extended to multi-allelic data (i.e., $m_i \geq 3$, DNA sequences) in~\cite{BankMHJ16}. It is defined as the single-step correlation of fitness effects for mutations  between neighboring genotypes. It quantifies how the fitness effect of a focal mutation if altered when it occurs in a different genetic background, averaged over all genotypes of the fitness landscape. Geometrically, $\gamma$ measures the correlation between ``slopes'' (i.e., ) of the same mutation put into different genetic backgrounds. Thus, if the fitness effect of a mutation is independent of its genetic background (i.e., if there is no epistasis), the correlation in slopes will be perfect ($\gamma = 1$), whereas it will be zero if the fitness slopes of each genotype are independent of the fitnesses of other genotypes. Depending on the scale $\gamma$ can either be used to quantify the strength of gene$\times$gene interactions between specific mutations or as an overall measure for the entire fitness landscape.

Then the matrix of epistatic effects between loci $i$ and $j$ carrying alleles $a_i, b_i \in \mathcal{A}_i$ and $A_j, B_j \in \mathcal{A}_j$ is given by

\begin{equation}
\gamma_{(a_i, b_i) \to (A_j, B_j)} = \text{Cor}\left[ s_{(A_j, B_j)}(g), s_{(A_j, B_j)}\left(g_{[(a_i, b_i)]}\right) \right] 
= \frac{\sum_{g} s_j(g) s_j(g_{[i]})}{\sum_{g} (s_j(g))^2}.
\end{equation}

where $g := \{x \in \mathcal{G} | x_i = a_i \text{ or } x_i = b_i \text{ and } x_j = A_j \text{ or } x_j = B_j\} \subseteq \mathcal{G}$ such that the sum is only calculated over the subset of genotypes carrying one of the two focal alleles at each focal locus. Thus, $\gamma_{(a_i, b_i) \to (A_j, B_j)}$ is a quadratic matrix of dimension $(\sum_{i=1}^n \frac{|\mathcal{A}_i|(|\mathcal{A}_i|-1)}{2})$.

Likewise, the epistatic effect of a mutation in locus $i$ with alleles $(a_i, b_i)$ on other loci (and pairs of alleles) can be calculated as

\begin{equation}
\gamma_{(a_i, b_i) \to} = \text{Cor}\left[ s(g), s\left(g_{[(a_i, b_i)]}\right) \right] 
= \frac{\sum_{j \neq i} \sum_{a_j} \sum_{g} s_j(g) s_j(g_{[i]})}
       {\sum_{j \neq i} \sum_{a_j} \sum_{g} (s_j(g))^2},
\end{equation}

where the summation index $\mathfrak{a}_j = \{(A_j, B_j) \mid A_j, B_j \in \mathcal{A}_j \text{ and } A_j \neq B_j\}$ is over the set of subsets of size two that can be constructed from all alleles found at locus $j$. Note that the third summation index $g$ changes depending on $\mathfrak{a}_j$.

An additional summation allows calculation of the epistatic effect of a mutation in locus $i$ carrying allele $(a_i)$ on other loci (and pairs of alleles) can be calculated as:

\begin{equation}
\gamma_{a_i \to} = \text{Cor}[s(g), s(g_{[a_i]})] = 
\frac{ \sum_{j \neq i} \sum_{\mathfrak{f}_i} \sum_{\mathfrak{a}_j} \sum_{g} s_j(g) s_j(g_{[i]}) }
     { \sum_{j \neq i} \sum_{\mathfrak{f}_i} \sum_{\mathfrak{a}_j} \sum_{g} (s_j(g))^2 },
\end{equation}

where \( \mathfrak{f}_i = \{ (a_i, b_i) \mid b_i \in \mathcal{A}_i \text{ and } a_i \neq b_i \} \) 
such that the sum is only calculated over the elements of the set of subsets of size two 
that can be constructed from all alleles found at locus \( i \) that contain allele \( a_i \).

Then, summing over \( l_i = \{ (a_i, b_i) \mid a_i, b_i \in \mathcal{A}_i \text{ and } a_i \neq b_i \} \), 
i.e., the elements of the set of subsets of size two that can be constructed from all alleles found at locus \( i \),
gives the epistatic effect of a mutation in locus \( i \):

\begin{equation}
\gamma_{i \to} = \text{Cor}[s(g), s(g_{[i]})] = 
\frac{ \sum_{j \neq i} \sum_{\mathfrak{l}_i} \sum_{\mathfrak{a}_j} \sum_{g} s_j(g) s_j(g_{[i]}) }
     { \sum_{j \neq i} \sum_{\mathfrak{l}_i} \sum_{\mathfrak{a}_j} \sum_{g} (s_j(g))^2 }.
\end{equation}

Similarly, the epistatic effect of other mutations (again considering pairs of alleles first) on locus \( j \) with alleles \( (a_i, b_i) \) can be calculated as

\begin{equation}
\gamma_{\to (A_j, B_j)} = \text{Cor}\left[ s_{(A_j, B_j)}(g), s_{(A_j, B_j)}(g_1) \right] = 
\frac{ \sum_{i \neq j} \sum_{\mathfrak{a}_i} \sum_{g} s_j(g) s_j(g_{[i]}) }
     { \sum_{i \neq j} \sum_{\mathfrak{a}_i} \sum_{g} (s_j(g))^2 },
\end{equation}

the epistatic effect of other mutations on locus \( j \) carrying allele \( A_j \) is given by

\begin{equation}
\gamma_{\to A_j} = \text{Cor}\left[ s_{(A_j)}(g), s_{(A_j)}(g_1) \right] = 
\frac{ \sum_{i \neq j} \sum_{\mathfrak{f}_j} \sum_{\mathfrak{a}_i} \sum_{g} s_j(g) s_j(g_{[i]}) }
     { \sum_{i \neq j} \sum_{\mathfrak{f}_j} \sum_{\mathfrak{a}_i} \sum_{g} (s_j(g))^2 },
\end{equation}

and the epistatic effect of other mutations on locus \( j \) becomes

\begin{equation}
\gamma_{\to j} = \text{Cor}[s_j(g), s_j(g_1)] = 
\frac{ \sum_{i \neq j} \sum_{\mathfrak{l}_j} \sum_{\mathfrak{a}_i} \sum_{g} s_j(g) s_j(g_{[i]}) }
     { \sum_{i \neq j} \sum_{\mathfrak{l}_j} \sum_{\mathfrak{a}_i} \sum_{g} (s_j(g))^2 },
\end{equation}

Finally, \(\gamma_d\), that is the decay of correlation of fitness effects with Hamming distance \(d\) (i.e., the cumulative epistatic effect of \(d\) mutations averaged over the entire fitness landscape) is calculated as

\begin{equation}
  \gamma_d = \text{Cor}[s(g), s_j(g_d)] = 
  \frac{ \sum_{g} \sum_{g_d} \sum_{j \neq i_1,i_2,\dots,i_d} \sum_{\mathcal{A}_j \setminus \{A_j\}} s_j(g) s_j(g_{[i_1 i_2 \dots i_d]}) }
      { \sum_{g} \sum_{\mathcal{A}_j \setminus \{A_j\}} (s_j(g))^2 },
  \label{eq:gamma}
\end{equation}

where the last summation is over all different alleles present at locus \( j \) except the one carried by genotype \( g \) at locus \( j \). We provide implementation of $\gamma_1$ in \texttt{GraphFLA} since increasing $d$ would significantly increase the amount of calculations required (e.g., $\gamma_d$ would require $\binom{n}{d}$ times more calculations compared to $\gamma_1$). In doing this, we utilize vector operation from \texttt{pandas} and \texttt{numpy} whenever possible to avoid nested loops as would appear with a brute-force implementation of \pref{eq:gamma}. 

\subsection{Navigability}

Navigability refers to the ease with which an evolving population can traverse the fitness landscape, typically towards genotypes of higher fitness, under the influence of mutation and natural selection. A highly navigable landscape allows populations to readily find high-fitness peaks, potentially the global optimum, through series of fitness-increasing mutations. Conversely, low navigability implies that evolutionary trajectories might be hindered, often getting trapped on suboptimal peaks due to fitness valleys or complex landscape structures~\cite{AguilarRodriguezPW17, PapkouRM23, WestmannGW24}. 

\subsubsection{Global Optima Accessibility}
\label{app:go_accessibility}

\begin{definition}[Adaptive Walk]
     \label{def:adaptive_walk}
     An adaptive walk is a sequence of genotypes $g_0, g_1, \dots, g_k$ such that each genotype $g_{t+1}$ is a neighbor of $g_t$ ($g_{t+1} \in \mathcal{N}(g_t)$) and has strictly higher fitness ($f(g_{t+1}) > f(g_t)$) for all steps $t \in \{0, \dots, k-1\}$. The walk terminates at step $k$ when $g_k$ is a local optimum (Definition~\ref{def:optima}). Under the strong selection weak mutation (SSWM) regime~\cite{Gillespie84}, evolution often proceeds along such paths, as natural selection favors higher-fitness genotypes and prevents populations from crossing fitness valleys.
\end{definition}

\begin{definition}[Evolutionary Accessibility]
\label{def:accessibility}
A target genotype $g_T \in \mathcal{G}$ is considered \textbf{accessible} from a starting genotype $g_S \in \mathcal{G}$ if there exists at least one adaptive walk (Definition~\ref{def:adaptive_walk}) connecting $g_S$ to $g_T$. Specifically, a sequence of single mutations $g_S = g_0, g_1, \dots, g_k = g_T$ exists such that $g_{t+1} \in \mathcal{N}(g_t)$ and $f(g_{t+1}) > f(g_t)$ for all $t \in \{0, \dots, k-1\}$.
\end{definition}

Global optimum accessibility quantifies the extent to which the global optimum ($g^*$) is accessible (Definition~\ref{def:accessibility}) from other genotypes in the landscape via adaptive walks (Definition~\ref{def:adaptive_walk}). Fitness landscape theory posits that rugged landscapes can hinder adaptation by trapping evolving populations on suboptimal local optima, thereby limiting access to the global optimum~\cite{VisserK14, KauffmanL87}.

Following~\cite{PapkouRM23}, we measure global optimum accessibility as the fraction of genotypes residing within the basin of attraction of $g^*$, denoted $\mathcal{B}(g^*)$ (Definition~\ref{def:basin}). This fraction represents the probability that an adaptive walk initiated from a random genotype will eventually reach $g^*$:

\begin{equation}
    \alpha_\text{go} = \frac{|\mathcal{B}(g^*)|}{|\mathcal{G}|} = \frac{|\mathcal{B}(g^*)|}{\prod_{i=1}^n m_i}.
    \label{eq:go_accessibility}
\end{equation}

An accessibility value $\alpha_\text{go} \approx 1$ suggests a highly navigable landscape where the global optimum is readily reachable via simple hill-climbing dynamics from most starting points. Conversely, a low $\alpha_\text{go}$ indicates that reaching the global optimum through adaptive walks alone is improbable for populations starting from random initial genotypes, suggesting they are likely to become trapped on local optima.

It is noteworthy that recent findings suggest some rugged landscapes can still be highly navigable, with the global optimum accessible from a large fraction of genotypes~\cite{PapkouRM23}. Furthermore, even if $g^*$ is inaccessible via adaptive walks from certain regions, this does not necessarily prevent its discovery through methods used in directed evolution. Many computational optimizers (e.g., simulated annealing~\cite{BlumR03}, machine learning-guided directed evolution) employ mechanisms to traverse fitness valleys and escape local optima. Nevertheless, highly rugged landscape topographies generally pose greater challenges for locating global fitness peaks~\cite{AguilarRodriguezPW17, WestmannGW24}.

\subsubsection{Basin of Attraction and BFC}
\label{sec:app_basin}

\begin{definition}[Basin of Attraction]
     \label{def:basin}
     The basin of attraction $\mathcal{B}(g^\ell)$ of a local optimum $g^\ell$ is the set of all genotypes $g \in \mathcal{G}$ from which at least one adaptive walk starting at $g$ can reach $g^\ell$:
     $$\mathcal{B}(g^\ell) = \{ g \in \mathcal{G} \mid \exists \text{ an adaptive walk } g_0=g, \dots, g_k=g^\ell \}.$$
     The size of the basin, $|\mathcal{B}(g^\ell)|$, reflects the accessibility of the local optimum $g^\ell$ within the genotype space. Larger basins are often associated with higher-fitness local optima~\cite{PapkouRM23}. We use the \textit{basin size-fitness correlation} to quantify this relationship.
\end{definition}

\texttt{GraphFLA} estimates basin sizes by analyzing adaptive walks. It supports two distinct methods, yielding different definitions and computational properties for basin size:

\begin{itemize}
     \item \textbf{Stochastic adaptive walks:} Also known as \textit{first-improvement hill climbing}~\cite{WhitleyHH13}. At each step, one neighbor $g' \in \mathcal{N}(g)$ with $f(g') > f(g)$ is chosen uniformly at random from all such improving neighbors. Because of this stochasticity, walks from the same starting genotype can terminate at different local optima, leading to overlapping basins~\cite{PapkouRM23}. \texttt{GraphFLA} calculates the basin size deterministically by identifying the set of \textit{all} genotypes from which \textit{at least one} such adaptive walk can reach the local optimum $g^\ell$. In the graph representation, this corresponds to finding all ancestors of the node $g^\ell$ reachable via directed paths representing fitness increases, typically using functions like those available in \texttt{igraph}. This computation can be expensive for large landscapes (e.g., $>300,000$ genotypes) with numerous local optima.

     \item \textbf{Greedy adaptive walks:} Also known as \textit{best-improvement hill climbing}. At each step, this walk deterministically selects a neighbor $g' \in \mathcal{N}(g)$ that maximizes the fitness increase, $\Delta f = f(g') - f(g)$. If multiple neighbors offer the same maximal increase, a consistent tie-breaking rule (e.g., random choice implemented once or lexicographical order) ensures determinism. The path from any starting genotype $g$ is unique, meaning each genotype belongs to the basin of exactly one local optimum. Consequently, these basins partition the genotype space $\mathcal{G}$, and their sizes sum to $|\mathcal{G}| = \prod_{i=1}^n m_i$. \texttt{GraphFLA} calculates these basin sizes by simulating a greedy walk starting from every genotype $g \in \mathcal{G}$ and recording the local optimum reached.
\end{itemize}

Using these two measures of basin size, \texttt{GraphFLA} provides two features, $\text{BFC}_\text{acc}$ and $\text{BFC}_\text{greedy}$ that assess whether local optima with higher fitness tend to have larger basin of attraction:

\begin{equation}
    \text{BFC}_\text{acc} = \text{Cor} \left[ f(g^\ell), |\mathcal{B}_\text{acc}(g^\ell)| \right]
    \label{eq:bfc_acc}
\end{equation}

\begin{equation}
    \text{BFC}_\text{greedy} = \text{Cor} \left[ f(g^\ell), |\mathcal{B}_\text{greedy}(g^\ell)| \right]
    \label{eq:bfc_greedy}
\end{equation}

A higher value of BFC indicates that fitter local optima would have larger basin of attraction compared to those have lower fitness, which results in their higher evolutionary accessibility and increased navigability of the whole landscape~\cite{PapkouRM23}.

\subsubsection{Fitness Distance Correlation (FDC)}
\label{app:fdc}

Fitness Distance Correlation (FDC) is a measure used to assess the global structure of a fitness landscape and its potential navigability by an evolutionary process~\cite{JonesF95}. Specifically, it quantifies the relationship between the fitness of genotypes and their distance to a known global optimum $g^*$. In biological fitness landscapes, this distance is typically the Hamming distance $d_H(g, g^*)$ (Definition~\ref{def:hamming}) between a genotype $g \in \mathcal{G}$ and the global optimum $g^* \in \mathcal{G}$ (Definition~\ref{def:optima}).

The FDC is calculated as the Pearson correlation coefficient between the fitness values $f(g)$ of all genotypes in the landscape (or a representative sample) and their respective Hamming distances to the global optimum $g^*$:

\begin{equation}
    \text{FDC} = \text{Cor} \left[ f(g), d_H(g, g^*) \right],
    \label{eq:fdc}
\end{equation}

The interpretation of the FDC value provides insights into the navigability of the fitness landscape:
\begin{itemize}
    \item \textbf{FDC $\approx -1$}: A strong negative correlation indicates that genotypes with higher fitness tend to be closer (i.e., have a smaller Hamming distance) to the global optimum $g^*$. This signifies a relatively smooth, ``funnel-like'' landscape structure where fitness gradients consistently guide an evolutionary search towards $g^*$. Such landscapes are considered highly navigable by processes like adaptive walks (Definition~\ref{def:adaptive_walk}), as selection for increased fitness generally directs the population towards the global peak.
    \item \textbf{FDC $\approx 0$}: A correlation close to zero suggests that there is no clear relationship between a genotype's fitness and its distance to the global optimum. This is characteristic of rugged or random landscapes, where fitness values can change erratically and provide little information about the direction towards $g^*$. In such landscapes, navigability is low, and evolutionary processes are more likely to become trapped on local optima (Definition~\ref{def:optima}) far from $g^*$.
    \item \textbf{FDC $\approx +1$}: A strong positive correlation implies that genotypes with higher fitness tend to be further away from the global optimum $g^*$. This indicates a ``deceptive'' landscape, where selection for immediate fitness gains would systematically lead an evolving population away from the global optimum. Such landscapes are exceptionally difficult to navigate towards $g^*$ using simple hill-climbing strategies.
\end{itemize}

\subsubsection{Evolvability-Enhancing Mutations}
\label{app:ee_mutation}

\textit{Evolvability}~\cite{PayneW19} refers to the capacity of a biological system to generate adaptive heritable variation. Unlike measures focusing solely on the direct fitness impact of a mutation (first-order selection), evolvability emphasizes the potential for future adaptive change (second-order selection)~\cite{PayneW19}. On top of this, \citeauthor{Wagner23} introduced the concept of an evolvability-enhancing (EE) mutation, defined as one that modifies the genetic background such that subsequent mutations at other loci tend to be, on average, more beneficial or less deleterious, irrespective of the initial mutation's own fitness effect.

\begin{definition}[Evolvability-Enhancing (EE) Mutation]
     \label{def:ee_mutation}
     Consider a mutation at locus $j$ that converts genotype $g$ to $g_{[j]}$. Let $\mathcal{N}_{-j}(g)$ denote the set of single-mutant neighbors of $g$ resulting from mutations at any locus $k \neq j$. The size of this set is $|\mathcal{N}_{-j}(g)| = \sum_{k \neq j, k=1}^n (m_k - 1)$. Let $\langle f \rangle_{\mathcal{N}_{-j}(g)}$ be the average fitness over the genotypes in $\mathcal{N}_{-j}(g)$. Similarly, let $\mathcal{N}_{-j}(g_{[j]})$ be the set of single-mutant neighbors of $g_{[j]}$ resulting from mutations at loci $k \neq j$, and let $\langle f \rangle_{\mathcal{N}_{-j}(g_{[j]})}$ be their average fitness.

     The mutation $g \to g_{[j]}$ is defined as evolvability-enhancing (EE) if:
     \begin{itemize}
         \item It is $\sigma$-neutral ($|s_j(g)| < \sigma$, see Def.~\ref{def:neutral_mutation}) and increases the average fitness of subsequent mutants:
         \begin{equation}
         \langle f \rangle_{\mathcal{N}_{-j}(g_{[j]})} - \langle f \rangle_{\mathcal{N}_{-j}(g)} > 0.
         \end{equation}
         \item It is beneficial ($s_j(g) > 0$) and enhances the fitness prospects of subsequent mutations beyond its own additive contribution:
         \begin{equation}
         \langle f \rangle_{\mathcal{N}_{-j}(g_{[j]})} - \langle f \rangle_{\mathcal{N}_{-j}(g)} > s_j(g).
         \end{equation}
         This condition is equivalent to requiring that the average fitness effect of subsequent mutations (relative to the background they arise in) is greater in the $g_{[j]}$ background than in the $g$ background:
         \begin{equation}
         \left( \langle f \rangle_{\mathcal{N}_{-j}(g_{[j]})} - f(g_{[j]}) \right) > \left( \langle f \rangle_{\mathcal{N}_{-j}(g)} - f(g) \right).
         \label{eq:ee_beneficial}
         \end{equation}
     \end{itemize}
\end{definition}

This definition implies that an EE mutation at locus $j$ exhibits, on average, positive epistasis with mutations occurring at other loci $k \neq j$ \cite{Wagner23}. Following \cite{Wagner23}, we primarily consider beneficial EE mutations. Such mutations can spread through populations via direct (first-order) selection due to their immediate fitness advantage, potentially increasing future evolvability as a byproduct without needing selection to act directly on evolvability itself (second-order selection). Beneficial EE mutations are expected to shift the distribution of fitness effects (DFE) of subsequent mutations favourably, for instance, by reducing the impact of deleterious mutations or increasing the frequency and/or magnitude of beneficial ones.

\subsubsection{Mean Accessible Path Length}
\label{app:mean_acc_path}

The minimum number of single mutations required to transition between a genotype $g$ and the global optimum $g^*$ is their Hamming distance, $d_H(g, g^*)$ (Definition~\ref{def:hamming}). However, not all paths realizing this minimum distance are necessarily evolutionarily accessible; that is, they may not consist solely of fitness-increasing steps (Section~\ref{app:go_accessibility}). Consequently, the shortest accessible path, composed entirely of fitness-increasing mutations (an adaptive walk, Definition~\ref{def:adaptive_walk}), can be longer than the Hamming distance, potentially requiring detours to navigate around fitness valleys~\cite{WuDOLS16,Cariani02}.

We measure the typical length of such paths using the mean shortest accessible path length to the global optimum, averaged over all genotypes from which $g^*$ is reachable.

\begin{definition}[Shortest Accessible Path Length]
     \label{def:shortest_acc_path}
     For a genotype $g$ within the basin of attraction of the global optimum $g^*$ (i.e., $g \in \mathcal{B}(g^*)$), the shortest accessible path length, $d_{\text{acc}}(g, g^*)$, is the minimum number of steps $k$ in an adaptive walk $g_0=g, \dots, g_k=g^*$ terminating at $g^*$. If $g$ is not in the basin of $g^*$ ($g \notin \mathcal{B}(g^*)$), then by definition $d_{\text{acc}}(g, g^*) = \infty$.
\end{definition}

The mean accessible path length to the global optimum, $\langle d_{\text{acc}} \rangle_{g^*}$, is calculated as:
$$ \langle d_{\text{acc}} \rangle_{g^*} = \frac{1}{|\mathcal{B}(g^*)|} \sum_{g \in \mathcal{B}(g^*)} d_{\text{acc}}(g, g^*). $$
The average is taken over all genotypes $g$ belonging to the basin of attraction $\mathcal{B}(g^*)$ of the global optimum (Definition~\ref{def:basin}).

In \texttt{GraphFLA}, $d_{\text{acc}}(g, g^*)$ is computed for all $g \in \mathcal{B}(g^*)$ using shortest path algorithms on the subgraph containing only fitness-increasing transitions, implemented via \texttt{igraph}'s \texttt{distances} targeting $g^*$. While the absolute value of $\langle d_{\text{acc}} \rangle_{g^*}$ is context-dependent (influenced by landscape size and dimensionality), comparing it to the mean Hamming distance between genotypes in the basin and the optimum, $\langle d_H(g, g^*) \rangle_{g \in \mathcal{B}(g^*)}$, can provide valuable insights into landscape navigability. In a perfectly smooth landscape with a single peak, $d_{\text{acc}}(g, g^*) = d_H(g, g^*)$ for all $g$. In rugged landscapes, accessible paths often meander, leading to $\langle d_{\text{acc}} \rangle_{g^*} > \langle d_H(g, g^*) \rangle_{g \in \mathcal{B}(g^*)}$~\cite{PapkouRM23, PoelwijkKWT07}. A larger difference signifies greater path indirectness imposed by the landscape's rugged structure.

\subsection{Epistasis}

\subsubsection{Classification of Epistasis}
\label{app:mag_sign_epistasis}

This section defines different types of pairwise epistatic interactions between mutations at two distinct loci, say $i$ and $j$. Epistasis occurs when the fitness effect of a mutation at one locus depends on the allele present at the other locus. We classify epistasis based on how the fitness effects change across genetic backgrounds. Consider a reference genotype $g$, the single mutants $g_{[i]}$ and $g_{[j]}$, and the double mutant $g_{[ij]}$ (assuming specific mutations $a_i \to a'_i$ at locus $i$ and $a_j \to a'_j$ at locus $j$ are implied or defined). The interaction epistasis term $\epsilon_{ij}$ measures the deviation from additivity:

\begin{align}
     \epsilon_{ij} &= f(g_{[ij]}) - f(g) - [f(g_{[i]}) - f(g)] - [f(g_{[j]}) - f(g)] \\
                   &= f(g_{[ij]}) - f(g_{[i]}) - f(g_{[j]}) + f(g)
\end{align}

Using the selection coefficient notation from Definition~\ref{def:selection_coeff}, where $s_i(g) = f(g_{[i]}) - f(g)$ is the effect of the mutation at locus $i$ in the background $g$, and $s_i(g_{[j]}) = f(g_{[ij]}) - f(g_{[j]})$ is the effect of the same mutation at locus $i$ but in the background $g_{[j]}$, the epistasis term can be equivalently written as:

$$ \epsilon_{ij} = s_i(g_{[j]}) - s_i(g) = s_j(g_{[i]}) - s_j(g) $$

Based on the sign and magnitude of the selection coefficients involved, we can classify the interaction:

\begin{itemize}
    \item \textbf{No epistasis ($\epsilon_{ij} = 0$):} The effects of the mutations are additive. The effect of mutation $i$ is the same regardless of the allele at locus $j$, i.e., $s_i(g_{[j]}) = s_i(g)$.

    \item \textbf{Magnitude epistasis ($\epsilon_{ij} \neq 0$, no sign changes):} The fitness effects are non-additive ($\epsilon_{ij} \neq 0$), but the sign of each mutation's effect remains consistent across the backgrounds considered. That is, $s_i(g)$ and $s_i(g_{[j]})$ have the same sign (or zero), and $s_j(g)$ and $s_j(g_{[i]})$ also have the same sign (or zero). This occurs when the combined effect deviates from the sum of individual effects.
    \begin{itemize}
        \item \textbf{Positive epistasis ($\epsilon_{ij} > 0$):} The combined effect is greater than expected from additivity ($f(g_{[ij]}) - f(g) > s_i(g) + s_j(g)$). This includes synergistic interactions where, for example, two beneficial mutations together yield a larger benefit than their sum, or two deleterious mutations are less harmful together than expected (antagonistic interaction between deleterious mutations).
        \item \textbf{Negative epistasis ($\epsilon_{ij} < 0$):} The combined effect is less than expected from additivity ($f(g_{[ij]}) - f(g) < s_i(g) + s_j(g)$). This includes antagonistic interactions like diminishing returns, where two beneficial mutations yield a smaller benefit together than their sum~\cite{ChouCDSM11} (see Section~\ref{app:global_epistasis}), or synergistic interactions where two deleterious mutations are more harmful together than expected. Negative epistasis can decelerate adaptation~\cite{KhanDSLC11,ChouCDSM11,KryazhimskiyRJD14}, create concave fitness peaks~\cite{PayneW14}, and increase mutational robustness near peaks~\cite{PayneW14, SarkisyanEtAl16}.
    \end{itemize}

    \item \textbf{Sign epistasis ($\epsilon_{ij} \neq 0$, one sign change):} The sign of the fitness effect of one mutation (e.g., beneficial vs. deleterious) flips depending on the background provided by the other mutation, while the second mutation's sign remains consistent. For instance, mutation $i$ might be beneficial in background $g$ ($s_i(g) > 0$) but deleterious in background $g_{[j]}$ ($s_i(g_{[j]}) < 0$), while mutation $j$'s sign remains the same ($s_j(g)$ and $s_j(g_{[i]})$ have the same sign). Sign epistasis restricts accessible mutational trajectories~\cite{WeinreichDDH06,PoelwijkKWT07,AguilarRodriguezPW17} and contributes to landscape ruggedness~\cite{Bank22,RomeroA09}.

    \item \textbf{Reciprocal sign epistasis ($\epsilon_{ij} \neq 0$, two sign changes):} A symmetric form where the sign of the effect of both mutations changes depending on the background provided by the other. For example, both single mutations might be deleterious ($s_i(g) < 0, s_j(g) < 0$), but each becomes beneficial in the background containing the other mutation ($s_i(g_{[j]}) > 0, s_j(g_{[i]}) > 0$), often leading to a beneficial double mutant ($f(g_{[ij]}) > f(g)$).
\end{itemize}

Formally, the type of epistatic interaction $\mathfrak{e}(g, i, j)$ between specific mutations at loci $i$ and $j$ relative to a reference genotype $g$ can be classified based on the selection coefficients:

\begin{equation}
\mathfrak{e}(g, i, j) =
\begin{cases}
\text{None} & \text{if } \epsilon_{ij} = 0 \\
\text{Magnitude} & \text{if } \epsilon_{ij} \neq 0 \text{ and } [s_i(g) \cdot s_i(g_{[j]}) \ge 0 \text{ and } s_j(g) \cdot s_j(g_{[i]}) \ge 0] \\
\text{Reciprocal Sign} & \text{if } s_i(g) \cdot s_i(g_{[j]}) < 0 \text{ and } s_j(g) \cdot s_j(g_{[i]}) < 0 \\
\text{Sign} & \text{otherwise (i.e., if } \epsilon_{ij} \neq 0 \text{ and exactly one sign product is negative)}
\end{cases}
\label{eq:epistasis_classification}
\end{equation}

where the condition $\epsilon_{ij} = s_i(g_{[j]}) - s_i(g) = 0$ defines the non-epistatic case. The product conditions check for sign changes (a negative product indicates a sign change, assuming neither term is zero).

The prevalence of each epistasis type across the entire landscape can be estimated by enumerating all pairs of single mutations originating from all possible reference genotypes $g$. Computationally, this involves analyzing local structures corresponding to double mutants relative to a reference genotype. For graph-based representations, these structures correspond to specific 4-node motifs. As noted by~\cite{PapkouRM23}, specific non-isomorphic directed motifs identifiable using graph libraries like \texttt{igraph} correspond to these epistasis types (e.g., motifs 66, 52, and 19 were identified as potentially representing magnitude, sign, and reciprocal sign epistasis, respectively).

\subsubsection{Global epistasis.}
\label{app:global_epistasis}

Global epistasis refers to systematic trends where the fitness effect of a mutation exhibits a predictable relationship with the overall fitness of the genetic background it occurs in \cite{BakerleeNSRD22, KryazhimskiyRJD14, KhanDSLC11}. Global epistasis often manifests in two particular forms:

\begin{itemize}
     \item \textbf{Diminishing returns epistasis:} This pattern describes scenarios where the fitness benefit ($f(g') - f(g) > 0$) conferred by a beneficial mutation decreases as the fitness of the genetic background, $f(g)$, increases. In other words, the positive impact of a beneficial mutation diminishes in already fit genotypes~\cite{ChouCDSM11}. To quantify this, we identify all beneficial single-step mutations (where the selection coefficient $f(g') - f(g)$ is positive) across the landscape. Diminishing returns is then measured by calculating the Pearson correlation coefficient between the fitness of the background genotype, $f(g)$, and the corresponding positive selection coefficient, $s(g \to g')$, across all such beneficial mutations. A negative correlation value indicates the presence of diminishing returns epistasis.
     \item \textbf{Increasing costs epistasis:} This describes a pattern where the fitness cost (negative effect) of a deleterious mutation ($f(g') - f(g) < 0$) becomes larger (more negative) as the fitness of the genetic background, $f(g)$, increases. This implies that fitter genotypes are less tolerant to deleterious mutations~\cite{JohnsonMKD19}. To quantify this, we identify all deleterious single-step mutations (where the selection coefficient $s(g \to g')$ is negative). Increasing costs epistasis is measured by calculating the Pearson correlation coefficient between the fitness of the background genotype, $f(g)$, and the magnitude (absolute value) of the corresponding negative selection coefficient, $|s(g \to g')|$, across all such deleterious mutations. A positive correlation value indicates the presence of increasing costs epistasis, meaning the fitness cost tends to be larger in higher-fitness backgrounds.
\end{itemize}

These global epistatic trends suggest a general "coupling" of mutations through overall fitness, potentially leading to predictable macro-evolutionary dynamics like decelerating rates of adaptation~\cite{KhanDSLC11,KryazhimskiyRJD14}. This contrasts with idiosyncratic epistasis (see the next section), where interactions depend more specifically on the identities and combination of the mutations involved~\cite{BakerleeNSRD22}.

\subsubsection{Idiosyncratic Epistasis}
\label{sec:idiosyncratic_epistasis}

Idiosyncratic epistasis describes genetic interactions where the fitness effect of a mutation depends strongly and often unpredictably on the specific genetic background~\cite{LyonsZXZ20,BakerleeNSRD22,AguirreHHML23}. The term "idiosyncratic" highlights that these interactions are specific to the identities of the involved loci and alleles, resulting in context-dependent effects that can vary significantly even across similar genetic backgrounds. This contrasts with global epistasis models in the previous section (e.g., diminishing returns, increasing costs), where a mutation's effect is predicted to vary systematically based mainly on the background genotype's fitness.

To quantify the extent of idiosyncratic epistasis in a landscape, we measure the variability of individual mutation effects across different genetic backgrounds. Following~\citeauthor{LyonsZXZ20}, an idiosyncrasy index can be calculated for each specific mutational transition.

Consider a specific mutation at locus $i$ changing allele $a_i \in \mathcal{A}_i$ to allele $b_i \in \mathcal{A}_i$ ($a_i \neq b_i$). Let $\mathcal{G}_{i, a_i} = \{g \in \mathcal{G} \mid \text{allele at locus } i \text{ in } g \text{ is } a_i\}$ denote the set of all genotypes (backgrounds) carrying allele $a_i$ at locus $i$. The selection coefficient for this specific mutation $a_i \to b_i$ occurring in a background $g \in \mathcal{G}_{i, a_i}$ is defined according to Definition~\ref{def:selection_coeff}:

\begin{equation}
s_{i, a_i \to b_i}(g) = f(g_{[i \leftarrow b_i]}) - f(g),
\end{equation}

where $g_{[i \leftarrow b_i]}$ is the genotype identical to $g$ but with allele $b_i$ at locus $i$.

The variability of this specific mutation's effect across all possible backgrounds is captured by its variance:

\begin{equation}
V_{i, a_i \to b_i} = \text{Var}_{g \in \mathcal{G}_{i, a_i}} \left[ s_{i, a_i \to b_i}(g) \right] = \frac{1}{|\mathcal{G}_{i, a_i}|} \sum_{g \in \mathcal{G}_{i, a_i}} \left( s_{i, a_i \to b_i}(g) - \bar{s}_{i, a_i \to b_i} \right)^2,
\end{equation}

where $\bar{s}_{i, a_i \to b_i}$ is the mean selection coefficient of the mutation $a_i \to b_i$ averaged over all backgrounds $g \in \mathcal{G}_{i, a_i}$.

To normalize this measure, the variability is compared to the overall variability of selection coefficients across the entire landscape. Let $\mathcal{S}$ represent the set of all possible single-mutation selection coefficients:

\begin{equation}
\mathcal{S} = \{ s_{k, C_k \to D_k}(h) \mid k \in \{1,\dots,n\}, C_k, D_k \in \mathcal{A}_k, C_k \neq D_k, h \in \mathcal{G}_{k, C_k} \}.
\end{equation}

The total variance of selection coefficients across the landscape is:

\begin{equation}
V_s = \text{Var}[s \in \mathcal{S}] = \frac{1}{|\mathcal{S}|} \sum_{s \in \mathcal{S}} (s - \bar{s})^2,
\end{equation}

where $\bar{s}$ is the global mean selection coefficient averaged over all single mutations in all backgrounds.

The idiosyncrasy index for the specific mutation $a_i \to b_i$ is the ratio of the standard deviation of its effect across backgrounds to the standard deviation of all selection coefficients in the landscape:

\begin{equation}
I_{\text{id}}(i, a_i \to b_i) = \frac{\sqrt{V_{i, a_i \to b_i}}}{\sqrt{V_s}}.
\end{equation}

An index value $I_{\text{id}}(i, a_i \to b_i) \approx 1$ signifies high idiosyncrasy, implying that the effect of this mutation is highly context-dependent, varying across backgrounds almost as much as selection coefficients vary globally. Conversely, $I_{\text{id}}(i, a_i \to b_i) \approx 0$ indicates low idiosyncrasy, suggesting the mutation has a relatively consistent effect regardless of the genetic background.

To derive a single measure representing the overall level of idiosyncrasy for the entire landscape, we average the index across all possible single mutations. Let $\mathcal{M}$ be the set of all possible directed single mutations, $\mathcal{M} = \{(i, a_i \to b_i) \mid i \in \{1,\dots,n\}, a_i, b_i \in \mathcal{A}_i, a_i \neq b_i\}$. The total number of such mutations is $|\mathcal{M}| = \sum_{k=1}^n m_k(m_k-1)$. The average idiosyncrasy index for the landscape is:

\begin{equation}
I_{\text{id}} = \frac{1}{|\mathcal{M}|} \sum_{(i, a_i \to b_i) \in \mathcal{M}} I_{\text{id}}(i, a_i \to b_i).
\label{eq:avg_idiosyncrasy}
\end{equation}

A high average idiosyncrasy $I_{\text{id}}$ suggests that, overall, predicting a mutation's fitness effect requires detailed knowledge of the specific genetic background, and highlights the prevalence of complex, context-specific interactions within the landscape.

\subsubsection{Pairwise and Higher-order Epistasis}
\label{app:pairwise}

Epistatic interactions can occur between pairs of mutations (pairwise epistasis) or among multiple ones (high-order epistasis). Ideally, to exactly determine the fraction of each pair of epistasis would require decompising the landscape into products of the single-locus variables according to the following expansion~\cite{HansenW01,WeinreichLWH13}:

\begin{equation}
f(g) = a^{(0)} + \sum_i a_i^{(1)} a_i + \sum_{ij} a_{ij}^{(2)} a_i a_j + \sum_{ijk} a_{ijk}^{(3)} a_i a_j a_k + \ldots + a_{12\dots n}^{(n)} a_1 a_2 \cdots a_n
\end{equation}

There are $\binom{n}{k}$ coefficients of type $a^{(k)}$ in this expansion, one for each subset of $k$ of $n$ loci. According to the binomial theorem, the total number of coefficients equals $2^n$, which makes it evident that the mapping between fitness values and expansion coefficients is one-to-one. The first-order coefficient $a^{(1)}$ describes the linear, non-epistatic effects, the second-order coefficient $a^{(2)}$ denotes pairwise epistatic interactions and so on.

However, such expansion is often computationally prohibitive because of the exponential growth in epistasis terms. As an alternative, researchers often specifically distinguish between pairwise and higher-order epistasis because the latter describes complex dependencies that are not reducible to combinations of pairwise interactions and can have profound consequences for the fitness landscape~\cite{WeinreichLWH13}. For example, higher-order epistasis can cause the effects of mutations, and even the nature of pairwise interactions, to change depending on the broader genetic background, sometimes leading to mutations switching between being beneficial and detrimental—outcomes that cannot be predicted by considering only pairwise effects~\cite{DomingoDL18}. 

To measure the prevalence of higher-order epistasis in the landscape, \texttt{GraphFLA} provides a measure, $\epsilon_\text{(2)}$, which assesses how much variance in fitness distributions can be explained by pairwise epistasis alone. This is performed by fitting a polynomial linear regression model with interaction terms up to the second order. $\epsilon_\text{(2)}$ is then derived as the $R^2$ score of this model in fitting the data. 

\subsection{Neutrality}

\subsubsection{Neutrality}
\label{app:neutrality} 

\textit{Mutational robustness}~\cite{LauringFA13,DraghiPWP10,PayneW15} measures the extent to which a genotype's fitness remains unchanged by mutations, reflecting its ability to buffer genetic perturbations. Related is the concept of \textit{\textit{neutral mutations}}, which cause little to no change in fitness. Neutral mutations allow populations to explore the genotype space without incurring significant fitness costs. Sets of genotypes connected by neutral mutations and sharing approximately the same fitness level form \textit{neutral networks}~\cite{PayneW14,PittF10,LauringFA13,GreenburyLA22}.

\begin{definition}[Neutral Mutation and Neighbors]
     \label{def:neutral_mutation}
     A mutation converting genotype $g$ into a neighbor $g' \in \mathcal{N}(g)$ is considered $\sigma$-neutral if the absolute fitness change is below a predefined tolerance $\sigma \ge 0$: $|f(g') - f(g)| < \sigma$. The set of $\sigma$-neutral neighbors of $g$ is:
     $$\mathcal{N}_{\sigma}(g) = \{ g' \in \mathcal{N}(g) \mid |f(g') - f(g)| < \sigma \}.$$
\end{definition}

Empirically, $\sigma$ is often set based on the variance in the measured fitness across replications and act as a noise threshold. With the above definition, we can now define the mutational robustness for a specific genotype:

\begin{definition}[Mutational Robustness]
     \label{def:robustness}
     The mutational robustness $R(g)$ of a genotype $g$ is the fraction of its neighbors that are $\sigma$-neutral~\cite{PayneW15,PayneW19}:
     $$R(g) = \frac{|\mathcal{N}_{\sigma}(g)|}{|\mathcal{N}(g)|}.$$
\end{definition}

To characterize the neutrality of the entire fitness landscape, we average the mutational robustness across all genotypes:

\begin{equation}
     \eta = \frac{1}{|\mathcal{G}|} \sum_{g \in \mathcal{G}} R(g).
     \label{eq:landscape_neutrality}
\end{equation}

This quantity, referred to as \textit{landscape neutrality}, measures the overall prevalence of neutrality. A high landscape neutrality ($\langle R \rangle_{\mathcal{G}}$) indicates the presence of extensive neutral networks~\cite{PayneW14,PittF10,LauringFA13,GreenburyLA22}. These networks can facilitate exploration of the genotype space via neutral drift without much fitness costs, which could potentially enhance \textit{evolvability}~\cite{PayneW19}, as discussed in the following section (\ref{app:ee_mutation}). 

\section{Landscape Models}
\label{app:landscape_models}

In the following, we briefly introduce the various fitness landscape models implemented in \texttt{GraphFLA}. For an excellent general overview, we refer the reader to a review by~\citeauthor{SzendroSFEV13}.

\textbf{The additive model.} In the additive model, the fitness of each genotype is given by the sum/product over the individual per locus fitness effects. Thus, the fitness effect of a specific allele, drawn from a Normal distribution with mean $\mu_a$ and variance $\sigma^2_a$~\cite{FerrettiSWYKTA16} (see also~\cite{NeidhartSK14}), is independent of its genetic background (i.e., it is constant across all genetic backgrounds), such that all mutations are non-interacting (i.e., there is no epistasis) and the resulting unimodal fitness landscape is (maximally) smooth. In particular, the roughness-to-slope ratio is 0 and $\mathbb{E}[\gamma_d] = 1$ for the entire range of mutational distances $d$. Note that when fitnesses are given by the product over the individual fitness effects, this model is also referred to as the multiplicative model.

\textbf{The House-of-cards model.} On the other extreme, in the House-of-cards (HoC) model~\cite{Kingman78} the fitness of each genotype is an i.i.d. normally distributed random variable with zero mean and variance $\sigma^2_{\mathrm{HoC}}$ resulting in an uncorrelated, maximally rugged fitness landscape that is characterized by multiple local optima~\cite{KauffmanL87,Kauffman93}. In particular, the fitness effect of an allele entirely depends on its genetic background such that there is complete interaction between all loci (i.e., full epistasis) which is also reflected in the roughness-to-slope ratio and $\mathbb{E}[\gamma_d]$ that become infinity and zero (for $d>0$), respectively.

\textbf{The Rough Mount Fuji model.} Introduced by~\cite{AitaUINKKHY00}, the Rough-Mount-Fuji (RMF) model, named after the eponymous mountain in Japan, which was initially formulated in the context of protein evolution [see also~\cite{NeidhartSK14}, for a simplified version], interpolates between the former two extremes. The fitness of a genotype is composed by an additive component (parametrized by $\mu_a$ and $\sigma_a^2$; see above) and a HoC component (parametrized by $\sigma_{\mathrm{HoC}}^2$) such that the extent of epistatic interactions ranges between none (additive model) to complete (HoC model) depending on the relative sizes of these three parameters. In particular, when $\sigma_{\mathrm{HoC}}^2 \ll \mu_a^2 + \sigma_a^2$ the RMF model becomes an additive model whereas for $\sigma_{\mathrm{HoC}}^2 \gg \mu_a^2, \sigma_a^2$ it essentially behaves like a HoC model~\cite{FerrettiSWYKTA16}. Accordingly, $0 \leq \mathbb{E}[\gamma_d] = \text{const} \leq 1$ for $d > 0$ and the roughness-to-slope ratio ranges from zero to infinity.

\textbf{The Kauffman NK model.} Another frequently used fitness landscape model that, as the RMF model, also interpolates between the additive and the HoC model~\cite{KauffmanL87,Kauffman93} is the Kauffman NK model, where $N$ di-allelic loci interact with $K \in \{0,1,\ldots,L-1\}$ randomly assigned other loci. In particular, for $K = 0$ the NK model collapses to an additive model whereas for $K = L - 1$ it approaches the HoC model. Although there are different ways how groups of interacting loci can be chosen~\cite{SchmiegeltK14}, properties such as the mean number and height of local optima tend only to be weakly dependent on the exact choice being made. For the NK model, $\mathbb{E}[\gamma_d]$ is a non-negative monotonically decreasing function in $d$, and the roughness-to-slope ratio can again range from zero to infinity.

\textbf{The eggbox model.} Introduced by~\cite{FerrettiSWYKTA16}, the eggbox model is a maximally epistatic, anticorrelated fitness landscape model (i.e., all loci interact with each other up to interactions of order $L$), in which the fitness effect of an allele switches from the highest to lowest value (or vice versa) between genetic backgrounds one step apart. Accordingly, depending on whether two genotypes are separated by an odd or even Hamming distance, their absolute fitness difference is either twice the mean allelic fitness effect or zero. Thus, this model generates an extreme case of reciprocal sign epistasis in which each mutation is either deleterious or compensatory, multiple local optima exist, and $\gamma_d$ accordingly oscillates between -1 and 1.

\section{Analysis for Phenotype Landscapes}
\label{app:phenotype}

\subsection{Phenotype Landscape Models}

To demonstrate the applicability of \texttt{GraphFLA} to analyzing phenotype landscapes, we consider several well-known systems, including $
\blacktriangleright$ the RNA secondary structure phenotype landscape for lengths $n=12$ and $n=15$ (RNA12, RNA15) representing the RNA sequence's minimum free energy folded secondary structure~\cite{SchusterFSH94,HofackerEFSTBS94,AguirreBSM11}, $
\blacktriangleright$ the Polyomino lattice self-assembly maps ($S_{2,8}$ and $S_{3,8}$) modelling the topology of protein quaternary structure assembled from interacting constituent tiles~\cite{GreenburySAL16,GreenburyJLA14,JohnstonDGCDAL22}, and $
\blacktriangleright$ several hydrophobic-polar (HP) lattice protein models for folding of a sequence into a tertiary structure (two compact models, HP\textsubscript{5x5} and HP\textsubscript{3x3x3}, and two non-compact ones, HP\textsubscript{20} and HP\textsubscript{25})~\cite{Dill85,IrbackT02,FerradaW12}. These phenotype landscapes have been thoroughly studied and compared in~\cite{GreenburySAL16,GreenburyLA22}

\textbf{RNA secondary structure:} The search space $\mathcal{G}$ is made of RNA sequences $g$ where each position can take 4 RNA nucleotide bases ($\mathcal{A}_i=\{A,C,G,U\}$). Phenotypes $\mathcal{P}$ are the secondary structure bonding pattern of the minimum free energy fold of the genotype, represented with the dot-bracket notation~\cite{HofackerEFSTBS94}. We use the Vienna package~\cite{HofackerEFSTBS94} with default parameters to convert RNA sequences $g \in \mathcal{G}$ to dot-bracket secondary structures $p \in \mathcal{P}$. Phenotype landscapes is the mapping from the sequence space $\mathcal{G}$ to their phenotypes $\mathcal{P}$, and are represented as RNA-$n$ with sequences of length $n$. As illustration, we consider $n = 12, 15$, resulting in the RNA12 and RNA15 phenotype landscapes. 

\textbf{HP lattice model:} In this model, genotypes $\mathcal{G}$ comprise sequences of hydrophobic (H) or polar (P) amino acids (alphabet $\mathcal{A}_i=\{H,P\}$)~\cite{Dill85,LauD89}. Phenotypes $\mathcal{P}$ correspond to the unique minimum energy conformation of a genotype when folded onto a 2D (square) or 3D (cubic) lattice. Folds are represented as strings of directional moves (2D: ``Up'', ``Down'', ``Left'', ``Right''; 3D additionally ``Forward'', ``Back''). Following~\cite{IrbackT02,FerradaW12}, only non-adjacent H-H pairs contribute to energy ($E_{HH} = -1$), while $E_{HP} = E_{PP} = 0$. Sequences lacking a unique minimum energy structure are considered undefined. We investigate both non-compact (HP\textsubscript{L}) and compact phenotype landscapes. In HP\textsubscript{L}, the phenotype is the minimum energy fold among all possible folds of a specific length. In compact maps (e.g., 2D HP\textsubscript{lxw} like HP\textsubscript{5x5}; 3D HP\textsubscript{lxwxh} like HP\textsubscript{3x3x3}), folds are confined to a prescribed grid. These compact models aim to better emulate the globular nature of native proteins~\cite{LiHTW96}, significantly reducing the conformational space while enhancing fidelity to observed protein topologies. We analyzed compact (HP\textsubscript{3x3x3} and HP\textsubscript{5x5}) and non-compact (HP\textsubscript{20} and HP\textsubscript{25}) landscapes for illustration.

\textbf{Polyomino model:} This model represents protein quaternary structure on a 2D square lattice using an assembly kit of tiles. Genotypes $g \in \mathcal{G}$ define this kit of $n_t$ tiles, where each tile edge has one of $n_c$ colors (interface types) denoted by integers. We follow~\cite{GreenburySAL16,GreenburyJLA14}, focusing on phenotype landscapes $S_{n_t,n_c}$, specifically $S_{2,8}$. We use $n_c = 8$ colors; tile edges are assigned bases from the alphabet $\mathcal{A}_i=\{0,1,2,3,4,6,7\}$. Interactions are restricted to $1 \leftrightarrow 2$, $3 \leftrightarrow 4$, and $5 \leftrightarrow 6$, while colors 0 and 7 are neutral. The genotype sequence, consisting of bases from $A$, is encoded clockwise onto the four edges of each tile in the kit. Phenotype construction begins by ``seeding'' the lattice with the first tile. Subsequent tiles from the kit are stochastically placed at complementary interaction sites on the lattice. Assembly halts if no further placements are possible or if the structure grows unboundedly. This assembly process is repeated $k=200$ times. The phenotype is the unique, rotationally invariant, bounded polyomino shape observed across the ensemble of assemblies.

\subsection{Phenotype Landscape Features}

Following~\cite{GreenburySAL16}, we consider the following 3 landscape features that are dedicated to characterizing phenotype landscape topography. The analysis results for the previously described landscapes are summarized in~\pref{tab:phenotype_landscape} below.

\begin{table*}[htp]
    \centering
    \caption{Features of different phenotype landscapes analyzed}
    \small
    \renewcommand{\arraystretch}{1.1}
    \begin{tabular}{lccllccccc}
    \hline
    Phenotype landscape & $|\mathcal{A}|$ & $n$ & $|\mathcal{G}|$ & $|\mathcal{P}|$ & $\phi_\text{del}$ & $\log_{10}R$ & $\eta_p$ \\
    \hline
    RNA12      & 4 & 12 & $4^{12}$ & 58       & 0.854 & 4.6 & 0.465 \\
    RNA15      & 4 & 15 & $4^{15}$& 432      & 0.650 & 5.9 & 0.482 \\
    $S_{2,8}$  & 8 &  8 & $8^{8}$& 14       & 0.537 & 5.8 & 0.487 \\
    HP5x5      & 2 & 25 & $2^{25}$& 550      & 0.816 & 4.1 & 0.285 \\
    HP3x3x3    & 2 & 27 & $2^{27}$& 49,808   & 0.939 & 2.2 & 0.115 \\
    HP20       & 2 & 20 & $2^{20}$& 5,311    & 0.976 & 0.7 & 0.102 \\
    HP25       & 2 & 25 & $2^{25}$& 107,337  & 0.977 & 0.9 & 0.099 \\
    \hline
    \end{tabular}
\label{tab:phenotype_landscape}
\end{table*}

\textbf{Redundancy.} Denoted by $R$, it is defined as the average number of distinct genotypes $g \in \mathcal{G}$ that map to each non-deleterious phenotype $p \in \mathcal{P}$. Redundancy is intrinsically linked to the average size of phenotypically neutral networks, which consist of sets of genotypes that share the same phenotype and are often connected by single mutations.

\textbf{Deleterious frequency.} Denoted as $\phi_{\text{del}}$, this metric represents the fraction of the entire genotype space $\mathcal{G}$ that is occupied by genotypes failing to map to a well-defined, functional phenotype. The nature of a deleterious phenotype is model-specific:
\begin{itemize}
    \item In RNA secondary structure landscapes, such as RNA12 and RNA15, a deleterious phenotype corresponds to an unfolded RNA sequence that lacks any defined secondary structure.
    \item For Hydrophobic-Polar (HP) lattice protein models, like HP\textsubscript{5x5} (a compact 2D model) or HP\textsubscript{20} (a non-compact model), a deleterious outcome signifies an amino acid sequence that does not fold into a unique minimum energy conformation.
    \item In the context of Polyomino lattice self-assembly models, for example $S_{2,8}$ or $S_{3,8}$ which model protein quaternary structure, a deleterious genotype is one that results in an unbounded or non-deterministic assembly process.
\end{itemize}

\textbf{Phenotypic neutrality.} Denoted as $\eta_p$, it is the average proportion of mutational neighbors of a genotype $g$ (i.e., genotypes $g' \in \mathcal{N}(g)$) that exhibit the same phenotype as $g$. This average is computed over all genotypes in $\mathcal{G}$ that correspond to non-deleterious phenotypes. The value of $\eta_p$ provides a measure of local neutral connectivity within the phenotypic landscape, indicating the extent to which mutations can occur without altering the observable phenotype. This concept of phenotypic neutrality is distinct from fitness-based neutrality ($\eta$) defined in~\ref{def:robustness}, as it specifically pertains to the preservation of phenotype rather than fitness.

\section{Directed Evolution}
\label{app:de}

This section provides details regarding how each directed evolution (DE) approach in~\pref{sec:broader_app} is implemented. Specifically, we considered $5$ DE variants. For each approach, the results are evaluated by the highest fitness variant they identified. To enable comparison across tasks, this is reported as percentiles from $0$ to $1$, where $1$ represents the fitness of the global optimum. Each approach is run with random initialization for $100$ repeatations, and we report the average in~\pref{sec:broader_app}.

\textbf{Basic DE.} In this simpliest form, DE is implemented via a greedy adaptive walk algorithm starting from a random variant $g \in \mathcal{G}$. For each step, it exhaustively searches within its neighborhood for the single-point mutation $g \to g^\prime, (g^\prime \in \mathcal{N}(g))$ that yields the highest fitness increase (i.e, $\Delta f$), until a local optimum is reached. 

\textbf{MLDE.} In this paradigm, a supervised ML model (TabPFN~\cite{HollmannMPKKHSH25})\footnote{We also experimented with other common models including XGBoost~\cite{ChenG16} and a convolutionary neural network (CNN)~\cite{FahlbergFHR24}. We report results for TabPFN in~\pref{sec:broader_app} as it yielded the highest performance.} is trained on a set of $N$ randomly sampled protein variants from $\mathcal{G}$ along with their fitness. Following~\cite{BiswasKAEEC21,WangTHPLYMY23,YangLBHAKHYA25}, we set $N=384$ as performance typically plateaus at this value. Increasing the sample size beyond this point yields only marginal gains while incurring higher costs. During training, protein sequences were represented using one-hot encoding flattened over the mutated sites. The trained model was then employed to predict fitness values for the entire library, with the top $96$ predicted variants selected for evaluation.

\textbf{MLDE with zero-shot warm start.} Instead of randomly selecting the training set from the entire search space $\mathcal{G}$, this approach first uses a zero-shot predictor (ESM~\cite{MeierRVLSR21} here) to identify prominent regions composed of the top ranked $10\%$ variants. The initial training samples are then sampled from these regions to bias the learning towards them. The subsequent steps are the same as in MLDE. 

\textbf{ALDE.} This active learning-assisted DE implements an iterative learning strategy with $3$ or $5$ rounds. In each round, the ML model (TabPFN) is trained on all data acquired up to that point. The initial round involves the same random sampling as in MLDE with $N=96$. For subsequent rounds, the ML model serves as an acquisition function to rank all variants in the library, thus guiding the selection of the next batch of variants for fitness evaluation. After the final round, the trained model predicts fitness values, and the top $96$ variants are selected for analysis for evaluation.

\textbf{ALDE with zero-shot warm start.} This approach is almost identical to ALDE, except that insteading of randomly sampling the initial $96$ variants, it employs the same zero-shot ranking as in MLDE with zero-shot warm start. 

\section{Additional Results}
\label{app:results}

\begin{figure*}[t!]
    \centering
    \includegraphics[width=\linewidth]{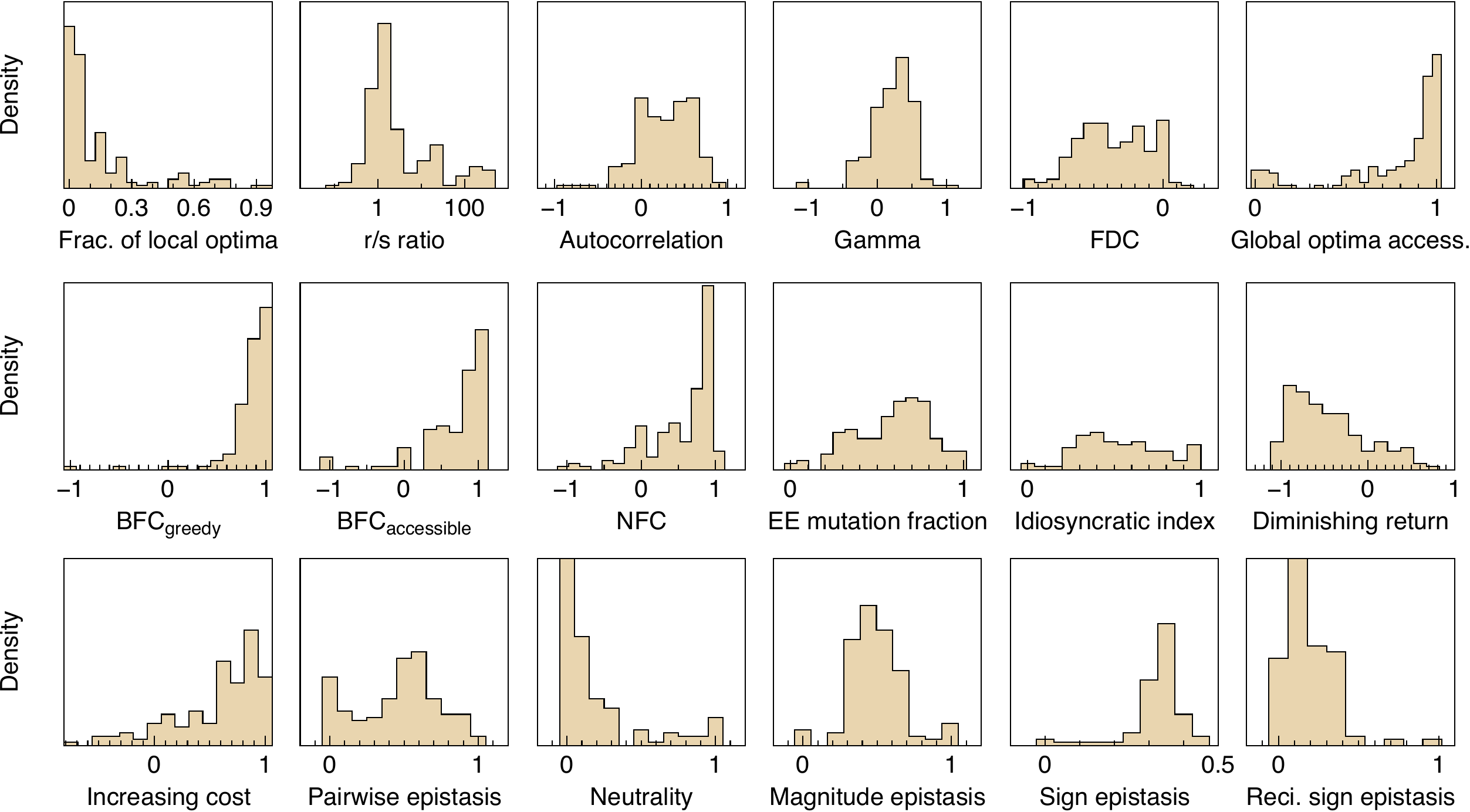}
    \caption{\small\textbf{Distribution of landscape features across our combinatorially complete datasets.}}
    \label{fig:dist_feature}
\end{figure*}

\begin{figure*}[t!]
    \centering
    \includegraphics[width=\linewidth]{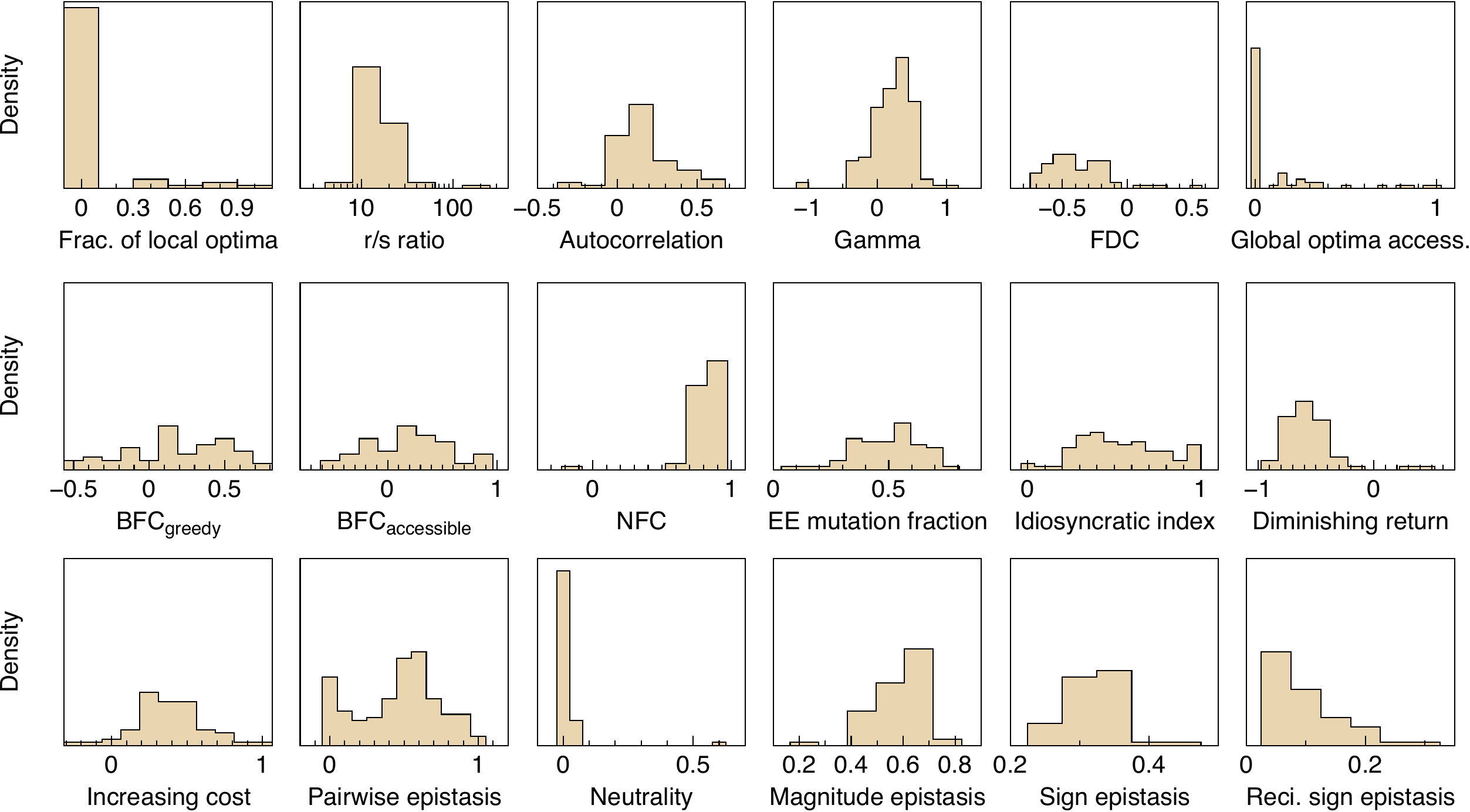}
    \caption{\small\textbf{Distribution of landscape features across ProteinGym tasks.}}
    \label{fig:dist_feature_proteingym}
\end{figure*}

\begin{figure*}[t!]
    \centering
    \includegraphics[width=\linewidth]{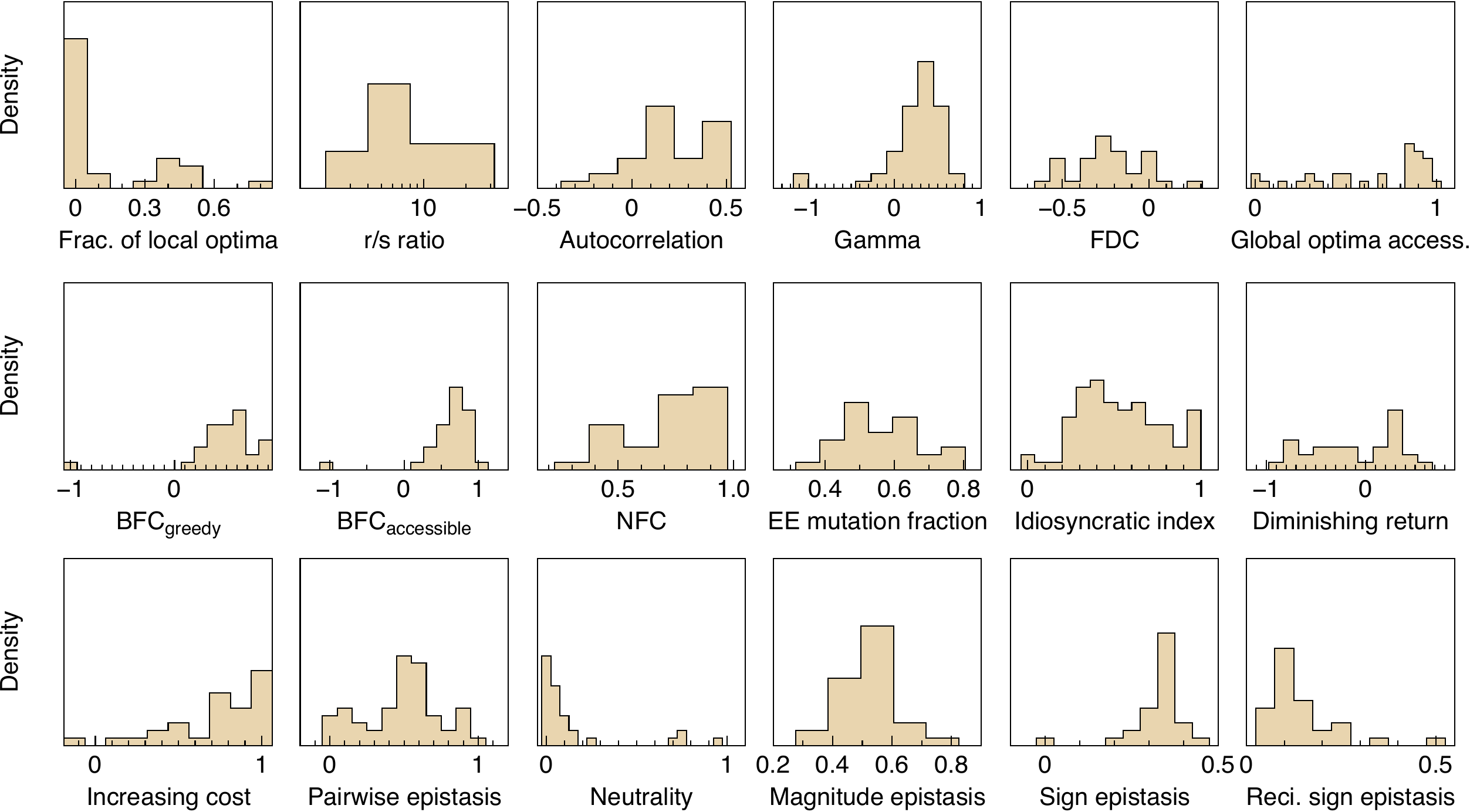}
    \caption{\small\textbf{Distribution of landscape features across ProteinGym tasks.}}
    \label{fig:dist_feature_rnagym}
\end{figure*}

\begin{figure*}[t!]
    \centering
    \includegraphics[width=\linewidth]{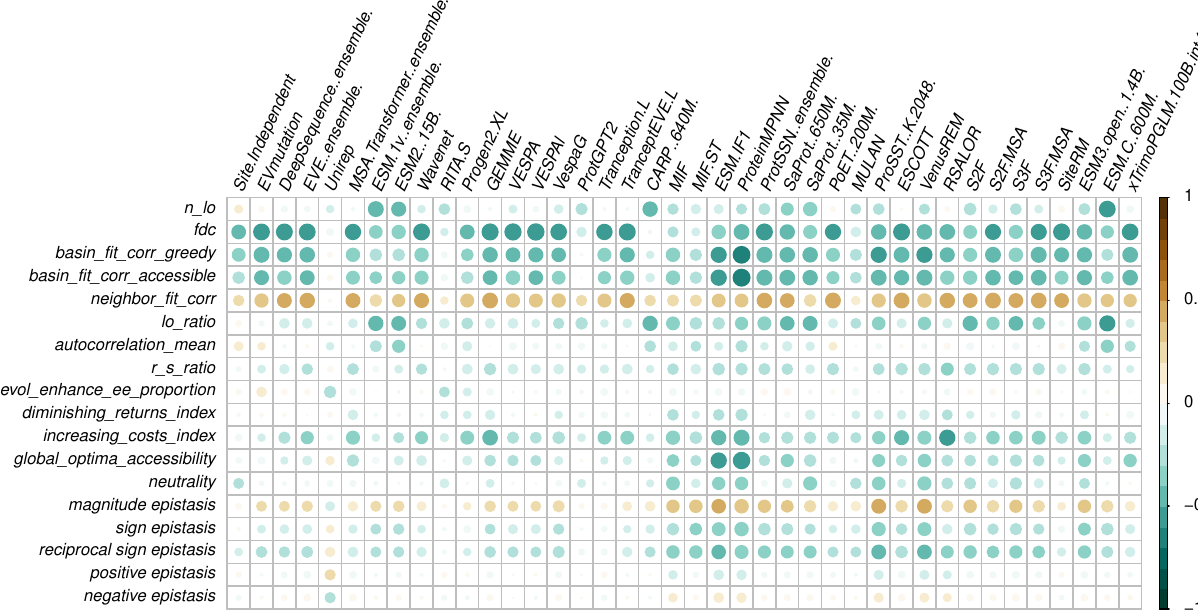}
    \caption{\small\textbf{Spearman correlation between fitness landscape features and the performance of zero-shot protein fitness models on ProteinGym tasks.} Due to the vast number of baslines on ProteinGym leaderboard, we only display representative ones here for each model series. Model performances are reported as Spearman's $\rho$ between predicted and true fitness.}
    \label{fig:heatmap_proteingym}
\end{figure*}

\begin{figure*}[t!]
    \centering
    \includegraphics[width=.5\linewidth]{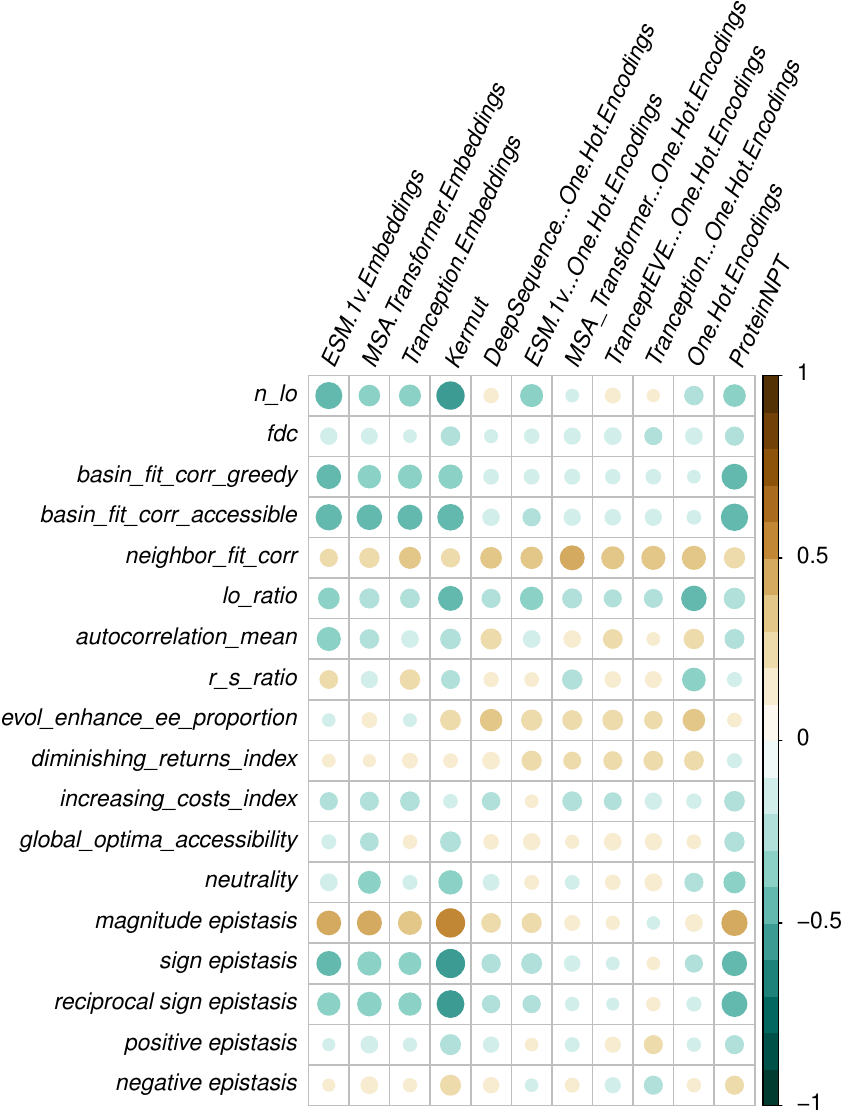}
    \caption{\small\textbf{Spearman correlation between fitness landscape features and the performance of supervised protein fitness models on ProteinGym tasks.} Model performances are reported as Spearman's $\rho$ between predicted and true fitness.}
    \label{fig:heatmap_proteingym_supervised}
\end{figure*}

\begin{figure*}[t!]
    \centering
    \includegraphics[width=.5\linewidth]{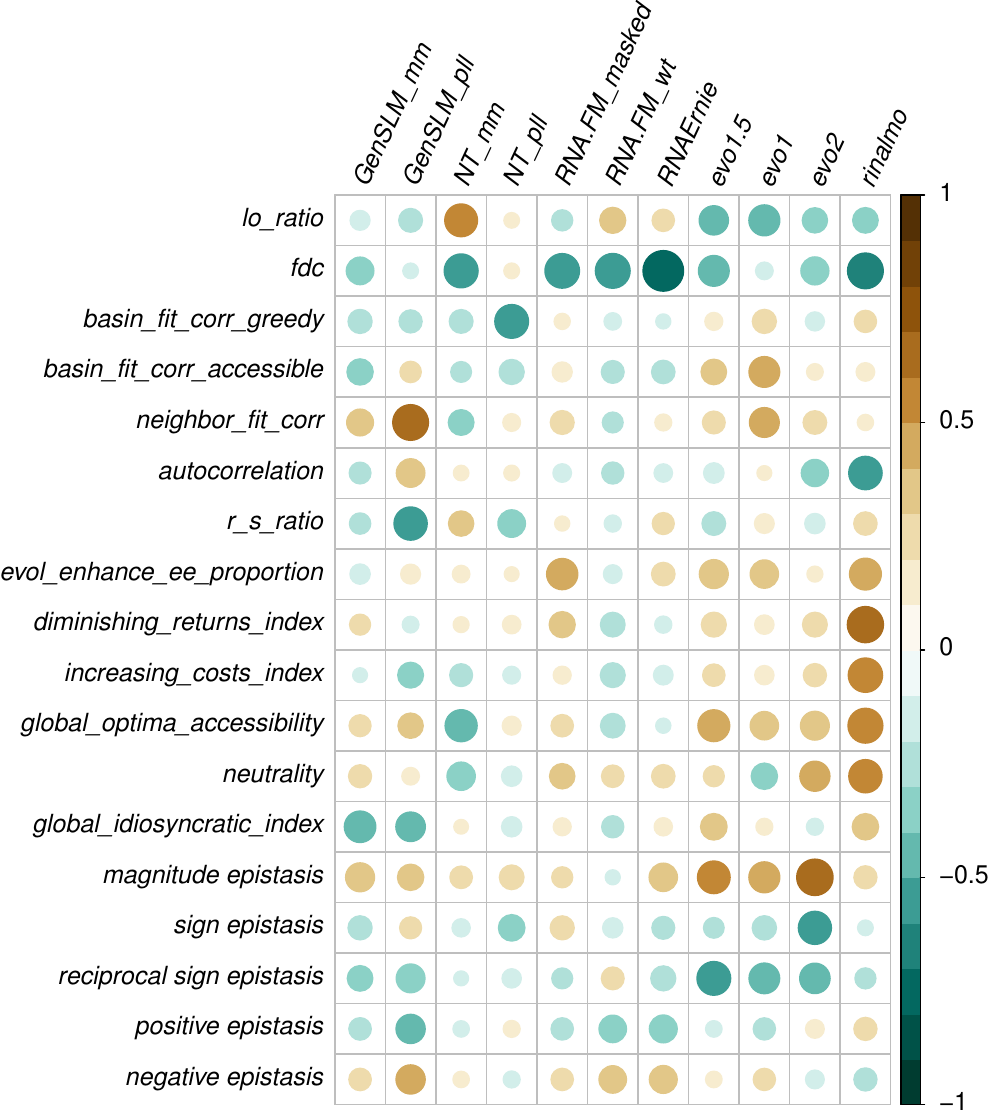}
    \caption{\small\textbf{Spearman correlation between fitness landscape features and the performance of supervised RNA fitness models on RNAGym tasks.} Model performances are reported as Spearman's $\rho$ between predicted and true fitness.}
    \label{fig:heatmap_rnagym}
\end{figure*}

\begin{figure*}[t!]
    \centering
    \includegraphics[width=.6\linewidth]{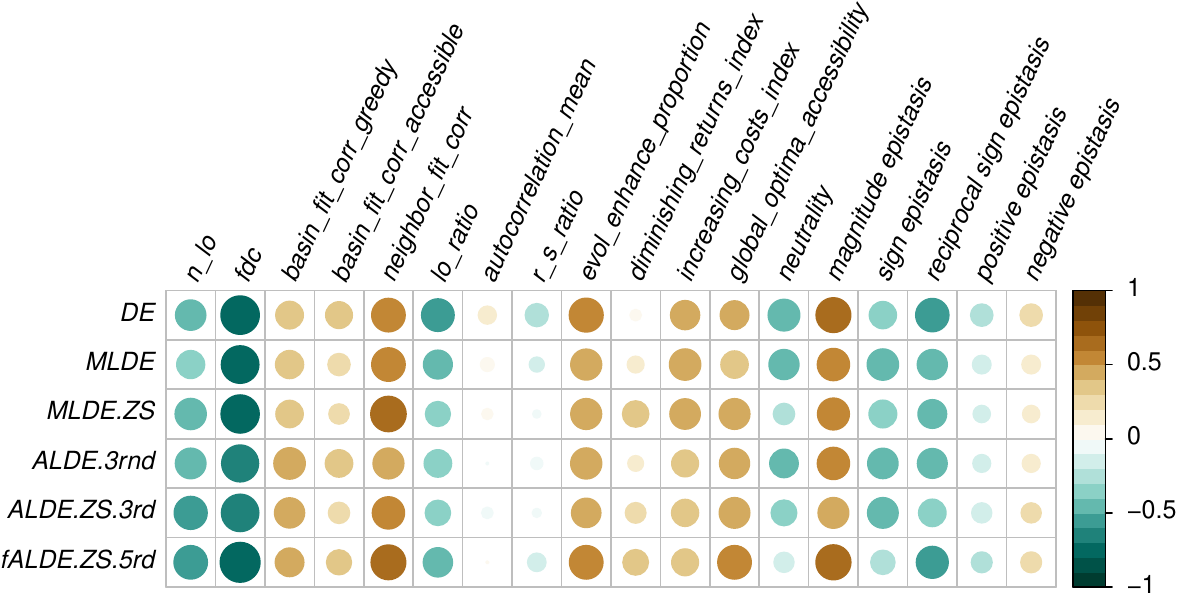}
    \caption{\small\textbf{Spearman correlation between fitness landscape features and the performance of 5 directed evolution (DE) approaches on 20 combinatorially complete protein fitness landscapes.} Model performances are reported as the maximum fitness (normalized to [0,1] by taking percentiles) reached, averaged across 100 randomly initialized repeatations.}
    \label{fig:de_corr}
\end{figure*}

\begin{figure*}[t!]
    \centering
    \includegraphics[width=\linewidth]{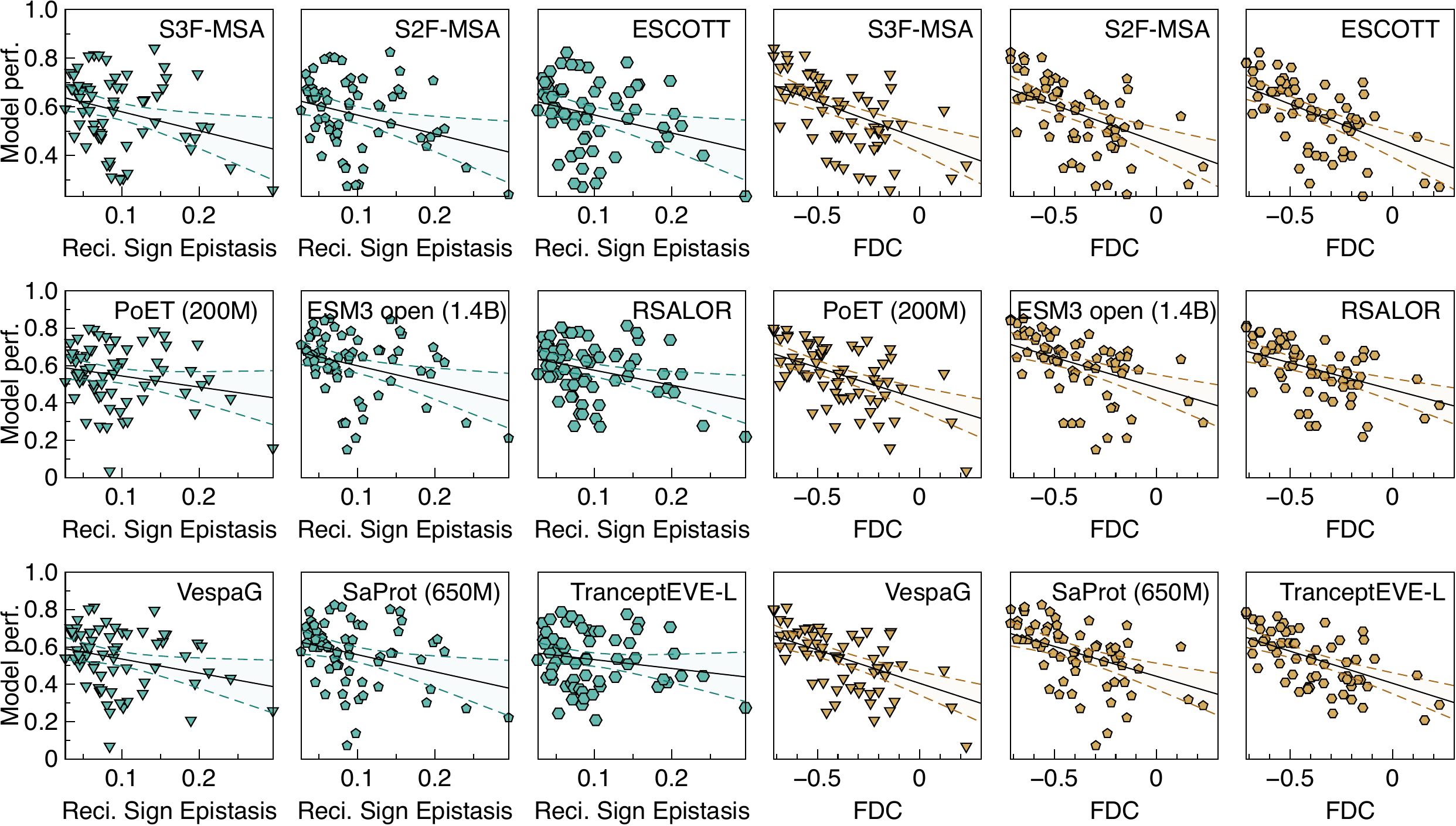}
    \caption{\small\textbf{Influence of landscape features on protein fitness model performance on ProteinGym.} We plot the distribution of model (name specified in each plot) performance ($y$-axis; measured as Spearman's $\rho$) against landscape features ($x$-axis). Straight lines show a fit of the linear regression model, and shaded regions depict the $95\%$ confidence intervals. References: S3F-MSA~\cite{ZhangNHLCMD024}, ESCOTT~\cite{Tekpinar24}, PoET~\cite{TruongB23}, ESM3~\cite{HayesRA+25}, RSALOR~\cite{TsishynHP+25}, VespaG~\cite{MarquetHO+22}, SaProt~\cite{SuHZSZY24}, and TranceptEVE-L~\cite{NotinVL22}.}
    \label{fig:app_proteingym_q1}
\end{figure*}

\begin{figure*}[t!]
    \centering
    \includegraphics[width=\linewidth]{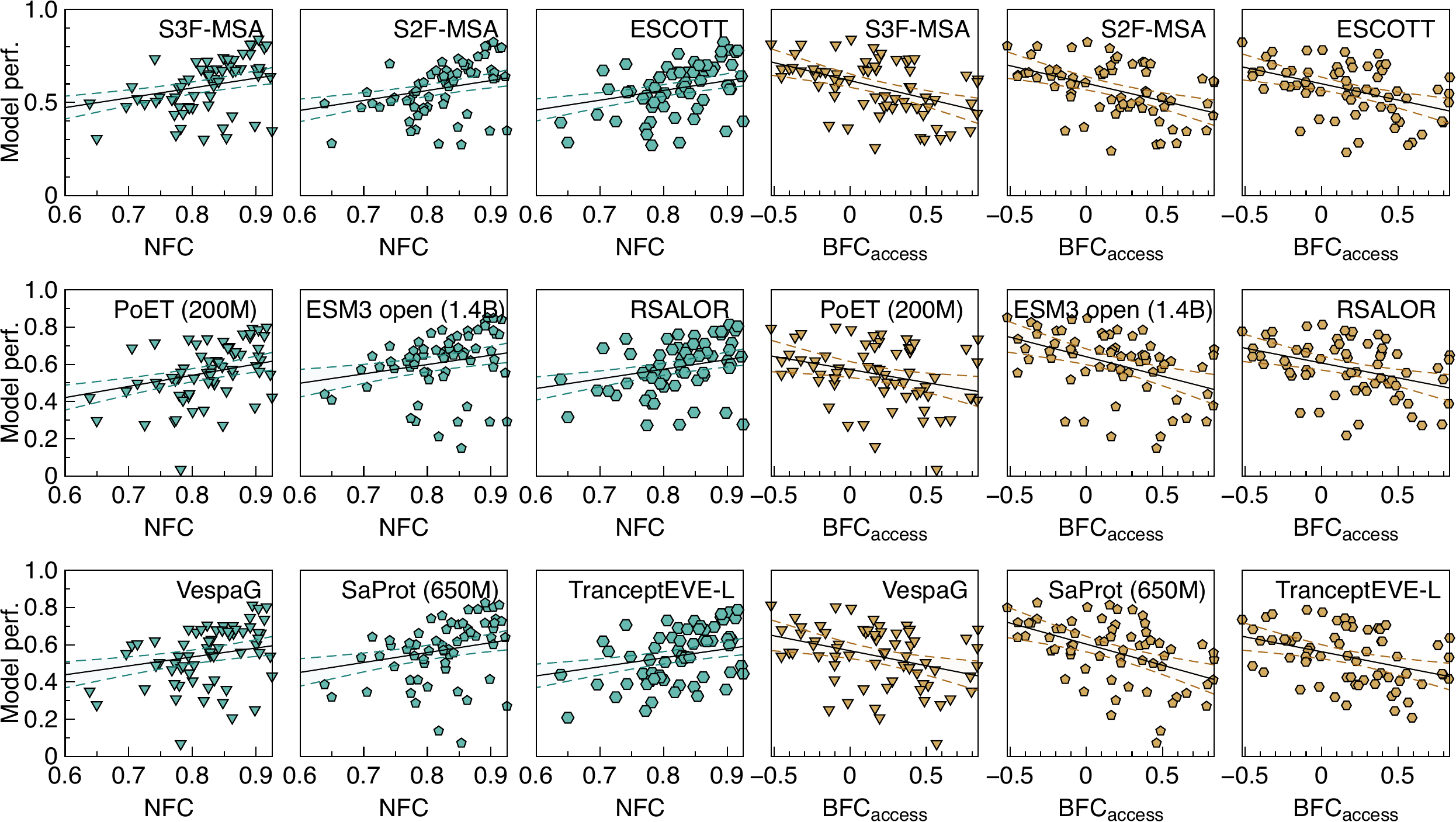}
    \caption{\small\textbf{Influence of landscape features on protein fitness model performance on ProteinGym.} We plot the distribution of model (name specified in each plot) performance ($y$-axis; measured as Spearman's $\rho$) against landscape features ($x$-axis). Straight lines show a fit of the linear regression model, and shaded regions depict the $95\%$ confidence intervals.}
    \label{fig:app_proteingym_q1_2}
\end{figure*}

\begin{figure*}[t!]
    \centering
    \includegraphics[width=\linewidth]{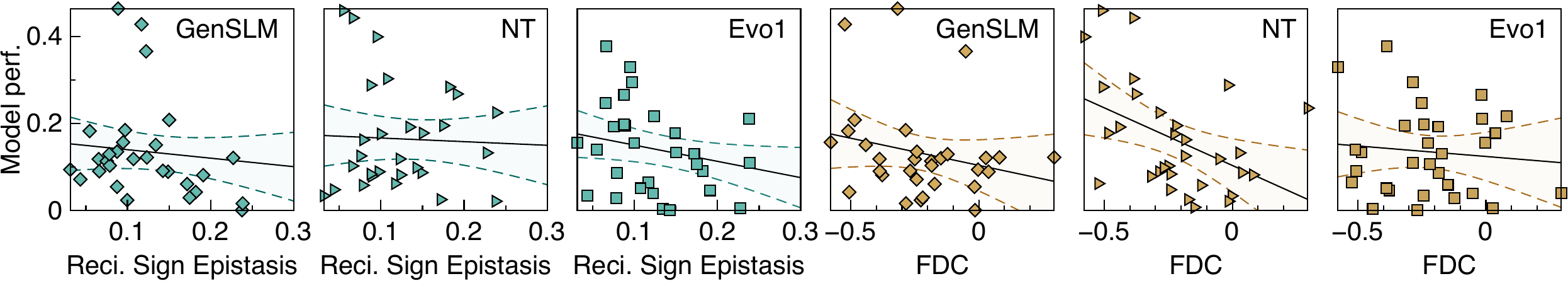}
    \caption{\small\textbf{Influence of landscape features on RNA fitness model performance on RNAGym.} We plot the distribution of model (name specified in each plot) performance ($y$-axis; measured as Spearman's $\rho$) against landscape features ($x$-axis). Straight lines show a fit of the linear regression model, and shaded regions depict the $95\%$ confidence intervals. References: GenSLM~\cite{ZvyaginBH23}, NT~\cite{DallaGM23}, and Evo1~\cite{NguyenPDKLLTTKBSNLALERBHH24}.}
    \label{fig:app_rnagym_q1}
\end{figure*}

\begin{figure*}[t!]
    \centering
    \includegraphics[width=\linewidth]{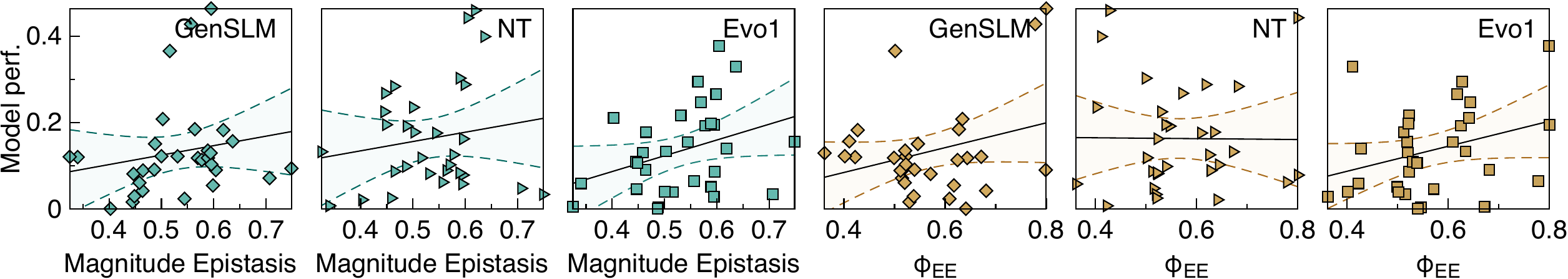}
    \caption{\small\textbf{Influence of landscape features on RNA fitness model performance on RNAGym.} We plot the distribution of model (name specified in each plot) performance ($y$-axis; measured as Spearman's $\rho$) against landscape features ($x$-axis). Straight lines show a fit of the linear regression model, and shaded regions depict the $95\%$ confidence intervals.}
    \label{fig:app_rnagym_q1_2}
\end{figure*}

\begin{figure*}[t!]
    \centering
    \includegraphics[width=\linewidth]{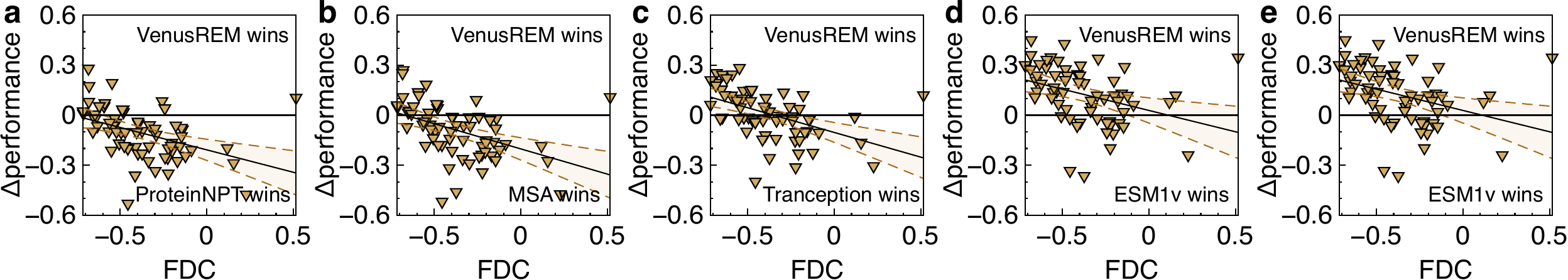}
    \caption{\small Difference in performance ($y$-axis) between $5$ supervised baselines and VenusREM in ProteinGym is plotted against landscape features on the $x$-axis. Straight lines show a fit of the linear regression model, and shaded regions depict the $95\%$ confidence intervals. References: VenusREM~\cite{TanWWZH24}, ProteinNP~\cite{NotinWMG23}, MSA~\cite{RaoLVMCASR21}, and Tranception~\cite{NotinDFMGMG22}.}
    \label{fig:app_comparison}
\end{figure*}

\begin{table*}[htbp]
    \centering
    \caption{Summary of landscape features for different microbial community-function landscapes.}
    \small 
    \renewcommand{\arraystretch}{1.1} 
    \begin{tabular}{lrrrrrr}
        \hline
        Feature & \cite{LangenhederBSP10} & \cite{SkwaraGYDRRTKS23} & \cite{Sanchez-GorostiagaBOPS19} & \cite{Diaz-ColungaSVBS24} & \cite{KeheKOATSGKGFB19} & \cite{ClarkCS21} \\
        \hline
        $n$ & 63 & 54 & 279 & 614 & 21198 & 1561 \\
        $\phi_{\text{lo}}$ & 0.032 & 0.074 & 0.075 & 0.040 & 0.158 & 0.584 \\
        $r/s$ & 0.944 & 0.863 & 1.417 & 1.778 & 3.087 & 2.294 \\
        $\rho_a$ & 0.528 & 0.491 & 0.556 & 0.490 & 0.559 & 0.196 \\
        $\gamma$ & 0.384 & 0.345 & 0.449 & 0.330 & 0.146 & 0.161 \\
        $\text{FDC}$ & -0.559 & -0.780 & -0.587 & -0.161 & -0.378 & -0.487 \\
        $\alpha_{\text{go}}$ & 0.984 & 0.944 & 0.888 & 0.853 & 0.003 & 0.003 \\
        $\text{BFC}_{\text{greedy}}$ & 1.000 & 0.800 & 0.861 & 0.724 & 0.502 & 0.156 \\
        $\text{BFC}_{\text{acc}}$ & 1.000 & 1.000 & 0.861 & 0.950 & 0.409 & 0.025 \\
        $\text{NFC}$ & 0.919 & 0.902 & 0.873 & 0.820 & 0.790 & 0.635 \\
        $\phi_{\text{EE}}$ & 0.683 & 0.709 & 0.644 & 0.625 & 0.535 & 0.402 \\
        $I_{\text{id}}$ & $0.398$ & $0.395$ & $0.539$ & $0.639$ & $0.552$ & $0.605$ \\
        $\epsilon_{\text{DR}}$ & -0.741 & -0.314 & -0.603 & -0.194 & 0.043 & -0.319 \\
        $\epsilon_{\text{IC}}$ & 0.648 & 0.871 & 0.574 & 0.728 & 0.816 & 0.583 \\
        $\epsilon_{\text{(2)}}$ & 0.861 & 0.869 & 0.633 & 0.361 & 0.547 & 0.675 \\
        $\eta$ & 0.038 & 0.000 & 0.527 & 0.183 & 0.000 & 0.000 \\
        $\epsilon_{\text{mag}}$ & 0.489 & 0.543 & 0.496 & 0.447 & 0.404 & 0.457 \\
        $\epsilon_{\text{sign}}$ & 0.431 & 0.413 & 0.396 & 0.375 & 0.395 & 0.354 \\
        $\epsilon_{\text{reci}}$ & 0.080 & 0.043 & 0.108 & 0.178 & 0.201 & 0.189 \\
        $\epsilon_{\text{pos}}$ & 0.764 & 0.768 & 0.742 & 0.840 & 0.841 & 0.774 \\
        $\epsilon_{\text{neg}}$ & 0.236 & 0.232 & 0.258 & 0.160 & 0.159 & 0.226 \\
        \bottomrule
    \end{tabular}
    \label{tab:microbiem}
\end{table*}

\begin{landscape}
     \centering
     \begingroup
     \small
     \renewcommand{\arraystretch}{1.1}
     \begin{longtable}{lllllrrrrrrrrrrc}
 
         \caption{Selected landscape features for combinatorially complete fitness landscapes} 
         \label{tab:datasets} \\

         \hline
         Author & Ref. & SubID & Space & Size & $\phi_{\text{lo}} (\%)$ & $\rho_a$ (\%) & NFC & $\gamma$ & $\epsilon_\text{reci}$ & $\epsilon_\text{DR}$ & FDC & $\phi_{\text{EE}}$ & $\eta$\\
         \hline
         \endfirsthead
 
         \multicolumn{14}{c}%
         {{\bfseries \tablename\ \thetable{} -- continued from previous page}} \\ 
         \hline
         Author & Ref. & SubID & Space & Size & $\phi_{\text{lo}}$ & $\rho_a$ & NFC & $\gamma$ & $\epsilon_\text{reci}$ & $\epsilon_\text{DR}$ & FDC & $\phi_{\text{EE}}$ & $\eta$  \\
         \hline
         \endhead
 
         \hline
         \multicolumn{14}{r}{{\small\textit{Continued on next page}}} \\ 
         \endfoot
 
         \hline
         \endlastfoot
 
         Kuo2020 & \cite{KuoJCCLHWC20} & - & Nucleotide & $4^9=262,144$ & 0.012 & 0.573 & 0.917 & 0.454 & 0.206 & 0.391 & -0.235 & 0.669 & 0.074\\
         Papkou2023 & \cite{PapkouRM23} & DHFR & Nucleotide & $4^9=262,144$ & 0.004 & 0.641 & 0.954 & 0.399 & 0.255 & 0.065 & -0.140 & 0.553 & 0.076\\
         PodgornaiaL15 & \cite{PodgornaiaL15} & PhoQ & Protein & $20^4=160,000$ & 0.013 & 0.241 & 0.540 & NaN & 0.234 & 0.469 & -0.196 & 0.609 & 0.282\\
         Wu2016 & \cite{WuDOLS16} & GB1 & Protein & $20^4=160,000$ & 0.001 & 0.406 & 0.872 & 0.440 & 0.221 & 0.499 & -0.123 & 0.621 & 0.717\\
         Jalal2020 & \cite{JalalTSCLTNLL20} & ParB & Protein & $20^4=160,000$ & 0.005 & 0.450 & 0.594 & 0.120 & 0.284 & -0.107 & -0.310 & 0.519 & 0.002\\
         Jalal2020 & \cite{JalalTSCLTNLL20} & Noc & Protein & $20^4=160,000$ & 0.004 & 0.346 & 0.388 & 0.131 & 0.331 & -0.164 & -0.313 & 0.426 & 0.003\\
         Tu2022 & \cite{TuSE22} & TEV & Protein & $20^4=160,000$ & 0.007 & 0.053 & 0.242 & 0.031 & 0.309 & -0.655 & 0.008 & 0.510 & 0.141\\
         Tu2022 & \cite{TuSE22} & T7 & Protein & $20^3=8,000$ & 0.018 & -0.025 & -0.172 & NaN & 0.341 & -0.855 & 0.036 & 0.380 & 0.018\\
         Johnston2024 & \cite{JohnstonAWLPYA24} & TrpB4 & Protein & $20^4=160,000$ & 0.005 & 0.399 & 0.831 & 0.195 & 0.294 & 0.393 & -0.112 & 0.562 & 0.343\\
         Johnston2024 &  \cite{JohnstonAWLPYA24} & TrpB3A & Protein & $20^3=8,000$ & 0.008 & 0.161 & 0.485 & 0.036 & 0.256 & -0.239 & -0.178 & 0.621 & 0.167\\
         Johnston2024 &  \cite{JohnstonAWLPYA24} & TrpB3B & Protein & $20^3=8,000$ & 0.009 & 0.074 & 0.306 & 0.025 & 0.295 & -0.326 & -0.021 & 0.555 & 0.205\\
         Johnston2024 &  \cite{JohnstonAWLPYA24} & TrpB3C & Protein & $20^3=8,000$ & 0.008 & 0.115 & 0.395 & 0.020 & 0.284 & -0.585 & -0.065 & 0.572 & 0.089\\
         Johnston2024 &  \cite{JohnstonAWLPYA24} & TrpB3D & Protein & $20^3=8,000$ & 0.006 & 0.251 & 0.676 & 0.145 & 0.264 & 0.325 & -0.125 & 0.611 & 0.175\\
         Johnston2024 &  \cite{JohnstonAWLPYA24} & TrpB3E & Protein & $20^3=8,000$ & 0.012 & 0.113 & 0.498 & 0.030 & 0.316 & 0.153 & -0.036 & 0.500 & 0.339\\
         Johnston2024 &  \cite{JohnstonAWLPYA24} & TrpB3F & Protein & $20^3=8,000$ & 0.011 & 0.105 & 0.390 & 0.026 & 0.309 & -0.236 & -0.197 & 0.500 & 0.179\\
         Johnston2024 &  \cite{JohnstonAWLPYA24} & TrpB3G & Protein & $20^3=8,000$ & 0.010 & 0.162 & 0.543 & 0.043 & 0.305 & 0.184 & -0.176 & 0.514 & 0.313\\
         Johnston2024 &  \cite{JohnstonAWLPYA24} & TrpB3H & Protein & $20^3=8,000$ & 0.015 & 0.046 & 0.253 & 0.011 & 0.322 & 0.110 & -0.012 & 0.475 & 0.346\\
         Johnston2024 &  \cite{JohnstonAWLPYA24} & TrpB3I & Protein & $20^3=8,000$ & 0.002 & 0.433 & 0.856 & 0.275 & 0.171 & 0.373 & -0.240 & 0.726 & 0.145\\
         Domingo2018 & \cite{DomingoDL18} & - & Nucleotide & $2^6 \times 3^4 = 5,184$ & 0.021 & 0.517 & 0.777 & 0.114 & 0.182 & -0.757 & -0.506 & 0.681 & 0.076\\
         Phillips2021 & \cite{PhillipsLMDCJCMWD21} & CR6261-h1 & Mutation & $2^{11}=2,048$ & 0.011 & 0.561 & 0.820 & 0.538 & 0.044 & -0.956 & -0.379 & 0.794 & 0.075\\
         Phillips2021 & \cite{PhillipsLMDCJCMWD21} & CR6261-h9 & Mutation & $2^{11}=2,048$ & 0.019 & 0.547 & 0.794 & 0.420 & 0.026 & -0.901 & -0.575 & 0.794 & 0.037\\
         Phillips2021 & \cite{PhillipsLMDCJCMWD21} & CR9114-h1 & Mutation & $2^{16}=65,536$ & 0.013 & 0.765 & 0.956 & 0.391 & 0.118 & -0.945 & -0.219 & 0.673 & 0.093\\
         Phillips2021 & \cite{PhillipsLMDCJCMWD21} & CR9114-h3 & Mutation & $2^{16}=65,536$ & 0.563 & 0.411 & 0.198 & NaN & 0.071 & -0.400 & -0.354 & 0.586 & 0.019\\
         Phillips2021 & \cite{PhillipsLMDCJCMWD21} & CR9114-flueB & Mutation & $2^{16}=65,536$ & 0.972 & -0.062 & -0.162 & NaN & 0.118 & -0.304 & -0.089 & 0.278 & 0.018\\
         Phillips2023 & \cite{PhillipsMBDSD23} & CH65-SI06 & Mutation & $2^{16}=65,536$ & 0.481 & 0.655 & 0.741 & 0.442 & 0.082 & -0.474 & -0.498 & 0.672 & 0.026\\
         Phillips2023 & \cite{PhillipsMBDSD23} & CH65-MA90 & Mutation & $2^{16}=65,536$ & 0.000 & 0.831 & 0.991 & 0.481 & 0.078 & -0.320 & -0.534 & 0.798 & 0.070\\
         Phillips2023 & \cite{PhillipsMBDSD23} & CH65-G189E & Mutation & $2^{16}=65,536$ & 0.144 & 0.664 & 0.860 & 0.408 & 0.072 & 0.569 & -0.554 & 0.763 & 0.034\\
         Westmann24 & \cite{WestmannGW24} & - & Nucleotide & $4^{8}=65,536$ & 0.118 & 0.202 & 0.362 & 0.072 & 0.305 & -0.170 & -0.111 & 0.457 & 0.053\\
         Soo2021 & \cite{SooSFW21} & 30C & Nucleotide & $4^8=65,536$ & 0.015 & 0.445 & 0.773 & 0.219 & 0.231 & -0.743 & -0.260 & 0.648 & 0.007\\
         Soo2021 & \cite{SooSFW21} & 37C & Nucleotide & $4^8=65,536$ & 0.021 & 0.419 & 0.748 & 0.196 & 0.240 & -0.785 & -0.004 & 0.639 & 0.008\\
         Wong2018 & \cite{WongKK18} & BRCA2 & Nucleotide & $32,768$ & 0.022 & 0.535 & 0.897 & 0.444 & 0.290 & 0.247 & 0.019 & 0.529 & 0.143\\
         Wong2018 & \cite{WongKK18} & SMN1 & Nucleotide & $32,768$ & 0.009 & 0.502 & 0.861 & 0.421 & 0.183 & 0.290 & 0.011 & 0.668 & 0.096\\
         Wong2018 & \cite{WongKK18} & IKBKAP & Nucleotide & $32,768$ & 0.036 & 0.392 & 0.808 & 0.380 & 0.375 & 0.216 & 0.004 & 0.372 & 0.102\\
         Moulana2022 &  \cite{MoulanaDP22} & ACE & Mutation & $2^{15}=32,768$ & 0.001 & 0.833 & 0.993 & 0.536 & 0.058 & -0.875 & -0.485 & 0.848 & 0.050\\
         Moulana2023 & \cite{MoulanaDPCRGSBD23} & CB6 & Mutation & $2^{15}=32,768$ & 0.006 & 0.811 & 0.965 & 0.368 & 0.107 & -0.593 & -0.394 & 0.774 & 0.040\\
         Moulana2023 & \cite{MoulanaDPCRGSBD23} & CoV555 & Mutation & $2^{15}=32,768$ & 0.020 & 0.770 & 0.957 & 0.180 & 0.201 & -0.707 & -0.150 & 0.598 & 0.040\\
         Moulana2023 & \cite{MoulanaDPCRGSBD23} & REGN10987 & Mutation & $2^{15}=32,768$ & 0.007 & 0.716 & 0.924 & 0.222 & 0.160 & -0.817 & -0.449 & 0.680 & 0.055\\
         Moulana2023 & \cite{MoulanaDPCRGSBD23} & S309 & Mutation & $2^{15}=32,768$ & 0.006 & 0.775 & 0.962 & 0.348 & 0.115 & -0.535 & -0.339 & 0.696 & 0.071\\
         Bendixsen2019 & \cite{BendixsenCOH19} & HDV & Mutation & $2^{14}=16,384$ & 0.071 & 0.239 & 0.240 & 0.218 & 0.410 & 0.192 & 0.140 & 0.271 & 0.232\\
         Bendixsen2019 & \cite{BendixsenCOH19} & Ligase & Mutation & $2^{14}=16,384$ & 0.004 & 0.347 & 0.803 & 0.510 & 0.137 & 0.418 & -0.437 & 0.705 & 0.810\\
         Poelwijk2019 & \cite{PoelwijkSR19} & eqFP611 & Mutation & $2^{13}=8,192$ & 0.009 & 0.675 & 0.979 & 0.452 & 0.187 & -0.281 & 0.077 & 0.623 & 0.146\\
         Lite2020 & \cite{LiteGNLGL20} & ParD2 & Protein & $20^3=8,000$ & 0.001 & 0.577 & 0.960 & 0.319 & 0.113 & -0.348 & -0.302 & 0.796 & 0.058\\
         Lite2020 & \cite{LiteGNLGL20} & ParD3 & Protein & $20^3=8,000$ & 0.001 & 0.579 & 0.957 & 0.287 & 0.068 & -0.560 & -0.245 & 0.851 & 0.052\\
         Centurion2019 & \cite{BaezaCenturionMSVL19} & - & Mutation & $2^{10} \times 3=3,072$ & 0.032 & 0.657 & 0.919 & 0.366 & 0.155 & -0.850 & -0.050 & 0.751 & 0.053\\
         Schulz2025 & \cite{SchulzTWW25} & - & Mutation & $2^{10}=1,024$ & 0.036 & 0.653 & 0.911 & 0.215 & 0.153 & -0.820 & -0.223 & 0.689 & 0.016\\
         Bakerlee2022 & \cite{BakerleeNSRD22} & hap-4NQ0 & Mutation & $2^{10}=1,024$ & 0.479 & 0.029 & 0.025 & 0.175 & 0.273 & -0.536 & -0.037 & 0.351 & 0.953\\
         Bakerlee2022 & \cite{BakerleeNSRD22} & hap-37C & Mutation & $2^{10}=1,024$ & 0.630 & 0.013 & 0.039 & 0.081 & 0.422 & 0.044 & 0.043 & 0.278 & 0.977\\
         Bakerlee2022 & \cite{BakerleeNSRD22} & hap-gu & Mutation & $2^{10}=1,024$ & 0.681 & -0.009 & -0.004 & -0.039 & 0.377 & -0.377 & 0.034 & 0.205 & 0.964\\
         Bakerlee2022 & \cite{BakerleeNSRD22} & hap-salt & Mutation & $2^{10}=1,024$ & 0.774 & -0.101 & 0.045 & 0.299 & 0.282 & 0.408 & 0.015 & 0.245 & 0.959\\
         Bakerlee2022 & \cite{BakerleeNSRD22} & hap-suloc & Mutation & $2^{10}=1,024$ & 0.510 & -0.024 & -0.013 & -0.036 & 0.434 & 0.114 & -0.046 & 0.224 & 0.992\\
         Bakerlee2022 & \cite{BakerleeNSRD22} & hap-YPDA & Mutation & $2^{10}=1,024$ & 0.633 & 0.007 & 0.012 & 0.083 & 0.378 & -0.943 & 0.039 & 0.271 & 0.972\\
         Bakerlee2022 & \cite{BakerleeNSRD22} & hom-4NQO & Mutation & $2^{10}=1,024$ & 0.570 & 0.065 & 0.048 & 0.185 & 0.305 & -0.417 & -0.033 & 0.326 & 0.936\\
         Bakerlee2022 & \cite{BakerleeNSRD22} & hom-37C & Mutation & $2^{10}=1,024$ & 0.535 & 0.022 & 0.024 & 0.052 & 0.412 & -0.163 & -0.015 & 0.235 & 0.992\\
         Bakerlee2022 & \cite{BakerleeNSRD22} & hom-gu & Mutation & $2^{10}=1,024$ & 0.740 & -0.043 & 0.012 & 0.086 & 0.357 & -0.452 & 0.000 & 0.209 & 0.925\\
         Bakerlee2022 & \cite{BakerleeNSRD22} & hom-salt & Mutation & $2^{10}=1,024$ & 0.770 & -0.129 & 0.027 & 0.190 & 0.250 & -0.005 & -0.032 & 0.277 & 0.881\\
         Bakerlee2022 & \cite{BakerleeNSRD22} & hom-suloc & Mutation & $2^{10}=1,024$ & 0.549 & 0.005 & -0.004 & 0.001 & 0.320 & -1.000 & 0.026 & 0.325 & 0.982\\
         Bakerlee2022 & \cite{BakerleeNSRD22} & hom-YPDA & Mutation & $2^{10}=1,024$ & 0.569 & 0.031 & 0.020 & 0.103 & 0.355 & -0.707 & -0.051 & 0.268 & 0.958\\
         Bank2016 & \cite{BankMHJ16} & - & Mutation & $640$ & 0.027 & 0.476 & 0.807 & 0.210 & 0.117 & 0.190 & -0.411 & 0.780 & 0.000\\
         Bank2016 & \cite{BankMHJ16} & - & Mutation & $2^{6}=64$ & 0.062 & 0.403 & 0.886 & 0.606 & 0.125 & -0.965 & -0.126 & 0.719 & 0.250\\
         Wu2020 & \cite{WuOTNNW20} & Bei89 & Mutation & $576$ & 0.005 & 0.651 & 0.980 & 0.475 & 0.024 & -0.556 & -0.428 & 0.916 & 0.008\\
         Wu2020 & \cite{WuOTNNW20} & Bk79 & Mutation & $576$ & 0.002 & 0.627 & 0.967 & 0.475 & 0.055 & -0.354 & -0.519 & 0.856 & 0.006\\
         Wu2020 & \cite{WuOTNNW20} & Bris07L194 & Mutation & $576$ & 0.007 & 0.680 & 0.951 & 0.466 & 0.141 & -0.087 & -0.463 & 0.726 & 0.009\\
         Wu2020 & \cite{WuOTNNW20} & Bris07P194 & Mutation & $576$ & 0.052 & 0.359 & 0.719 & 0.191 & 0.272 & 0.288 & -0.405 & 0.540 & 0.006\\
         Wu2020 & \cite{WuOTNNW20} & HK68 & Mutation & $576$ & 0.016 & 0.595 & 0.946 & 0.522 & 0.070 & -0.711 & -0.260 & 0.846 & 0.007\\
         Wu2020 & \cite{WuOTNNW20} & Mos99 & Mutation & $576$ & 0.012 & 0.632 & 0.941 & 0.399 & 0.089 & -0.621 & -0.498 & 0.809 & 0.006\\
         Wu2020 & \cite{WuOTNNW20} & NDako16 & Mutation & $576$ & 0.003 & 0.700 & 0.971 & 0.523 & 0.040 & -0.251 & -0.644 & 0.875 & 0.011\\
         Lunzer2005 & \cite{LunzerMFD05} & fitness & Protein & $512$ & 0.002 & 0.496 & 0.918 & 0.639 & 0.019 & -0.978 & -0.586 & 0.939 & 0.003\\
         Lunzer2005 & \cite{LunzerMFD05} & NAD & Protein & $512$ & 0.002 & 0.561 & 0.953 & -0.024 & 0.000 & -0.762 & -0.503 & 1.000 & 0.036\\
         Lunzer2005 & \cite{LunzerMFD05} & NADP & Protein & $512$ & 0.002 & 0.685 & 0.972 & -0.004 & 0.000 & -0.671 & -0.665 & 1.000 & 0.000\\
         Doud2024 & \cite{DoudGLMDFM24} & base & Mutation & $2^9=512$ & 0.016 & 0.568 & 0.921 & 0.549 & 0.105 & -0.462 & -0.346 & 0.740 & 0.005\\
         Doud2024 & \cite{DoudGLMDFM24} & LamB & Mutation & $2^9=512$ & 0.012 & 0.625 & 0.939 & 0.511 & 0.133 & -0.057 & -0.548 & 0.729 & 0.011\\
         Doud2024 & \cite{DoudGLMDFM24} & Lspec & Mutation & $2^9=512$ & 0.026 & 0.586 & 0.932 & 0.628 & 0.107 & -0.594 & -0.270 & 0.722 & 0.008\\
         Doud2024 & \cite{DoudGLMDFM24} & OmpF & Mutation & $2^9=512$ & 0.040 & 0.558 & 0.901 & 0.538 & 0.124 & -0.460 & -0.265 & 0.681 & 0.009\\
         Doud2024 & \cite{DoudGLMDFM24} & Ospec & Mutation & $2^9=512$ & 0.020 & 0.541 & 0.931 & 0.619 & 0.097 & -0.859 & -0.108 & 0.753 & 0.011\\
         Colunga2024 & \cite{DiazColungaCSRAS24} & colorants & Mutation & $2^8=256$ & 0.004 & 0.730 & 0.996 & 0.162 & 0.001 & -0.507 & -0.854 & 0.981 & 0.015\\
         Colunga2024 & \cite{DiazColungaCSRAS24} & pseudo & Mutation & $2^8=256$ & 0.035 & 0.396 & 0.782 & 0.290 & 0.198 & -0.838 & -0.306 & 0.615 & 0.055\\
         Hall2020 & \cite{HallRWRCCLSPEA20} & NfsA-2039 & Mutation & $2^{7}=128$ & 0.016 & 0.586 & 0.945 & 0.404 & 0.074 & -0.461 & -0.363 & 0.738 & 0.518\\
         Hall2020 & \cite{HallRWRCCLSPEA20} & NfsA-3637 & Mutation & $2^{7}=128$ & 0.031 & 0.601 & 0.946 & 0.392 & 0.071 & -0.496 & -0.509 & 0.770 & 0.107\\
         Frohlich2021 & \cite{Frohlich21} & CAZtraj1 & Mutation & $2^{4}=16$ & 0.062 & 0.228 & 0.867 & 0.322 & 0.083 & -0.333 & -0.714 & 0.719 & 0.156\\
         Frohlich2021 & \cite{Frohlich21} & CAZtraj2 & Mutation & $2^{6}=64$ & 0.031 & 0.523 & 0.942 & 0.600 & 0.026 & -0.149 & -0.446 & 0.880 & 0.047\\
         Frohlich2021 & \cite{Frohlich21} & CAZtraj3 & Mutation & $2^{6}=64$ & 0.016 & 0.513 & 0.962 & 0.320 & 0.100 & -0.367 & -0.433 & 0.647 & 0.258\\
         Frohlich2021 & \cite{Frohlich21} & PIPtraj1 & Mutation & $2^{4}=16$ & 0.125 & 0.182 & 0.745 & 0.044 & 0.125 & -0.621 & 0.000 & 0.625 & 0.688\\
         Frohlich2021 & \cite{Frohlich21} & PIPtraj2 & Mutation & $2^{6}=64$ & 0.016 & 0.418 & 0.862 & 0.423 & 0.106 & -0.171 & -0.373 & 0.649 & 0.455\\
         Frohlich2021 & \cite{Frohlich21} & PIPtraj3 & Mutation & $2^{6}=64$ & 0.062 & 0.581 & 0.958 & 0.030 & 0.151 & -0.719 & -0.441 & 0.587 & 0.032\\
         Hall2010 & \cite{HallAP10} & Haploid & Mutation & $2^6=64$ & 0.141 & 0.239 & 0.483 & -0.150 & 0.292 & -0.814 & -0.263 & 0.469 & 0.245\\
         Hall2010 & \cite{HallAP10} & Diploid & Mutation & $2^6=64$ & 0.125 & 0.214 & 0.451 & -0.235 & 0.342 & -0.746 & -0.316 & 0.443 & 0.130\\
         Tamer2019 & \cite{TamerGARARAT19} & kcat-trajr & Mutation & $2^{5}=32$ & 0.125 & 0.222 & 0.706 & 0.253 & 0.163 & -0.941 & -0.214 & 0.575 & 0.163\\
         Tamer2019 & \cite{TamerGARARAT19} & kcat-trajg & Mutation & $2^{5}=32$ & 0.125 & 0.256 & 0.800 & 0.480 & 0.150 & -0.961 & -0.357 & 0.662 & 0.075\\
         Tamer2019 & \cite{TamerGARARAT19} & ki-trajr & Mutation & $2^{5}=32$ & 0.031 & 0.471 & 0.966 & 0.192 & 0.000 & -0.066 & -0.952 & 1.000 & 0.000\\
         Tamer2019 & \cite{TamerGARARAT19} & ki-trajg & Mutation & $2^{5}=32$ & 0.031 & 0.472 & 0.971 & 0.510 & 0.000 & 0.333 & -0.959 & 1.000 & 0.000\\
         Lozovsky2021 & \cite{LozovskyDHJH21} & ic50-c57 & Mutation & $2^{4}=16$ & 0.125 & 0.006 & 0.320 & -0.059 & 0.133 & -0.271 & -0.594 & 0.414 & 0.000\\
         Lozovsky2021 & \cite{LozovskyDHJH21} & ic50-c58 & Mutation & $2^{4}=16$ & 0.188 & 0.034 & 0.342 & -0.330 & 0.125 & -0.233 & -0.539 & 0.483 & 0.000\\
         Lozovsky2021 & \cite{LozovskyDHJH21} & ic50-c59 & Mutation & $2^{4}=16$ & 0.125 & -0.039 & 0.132 & -0.169 & 0.167 & 0.594 & -0.606 & 0.407 & 0.000\\
         Lozovsky2021 & \cite{LozovskyDHJH21} & ic50-c60 & Mutation & $2^{4}=16$ & 0.250 & -0.167 & -0.064 & -0.232 & 0.222 & -0.477 & -0.184 & 0.400 & 0.240\\
         Lozovsky2021 & \cite{LozovskyDHJH21} & ic50-c61 & Mutation & $2^{4}=16$ & 0.312 & -0.139 & -0.010 & 0.015 & 0.167 & -0.394 & -0.536 & 0.273 & 0.000\\
         Hall2019 & \cite{HallKCBCM19} & Acetate & Mutation & $2^{5}=32$ & 0.094 & 0.171 & 0.494 & -0.060 & 0.200 & -0.757 & -0.250 & 0.500 & 0.100\\
         Hall2019 & \cite{HallKCBCM19} & Beef & Mutation & $2^{5}=32$ & 0.062 & 0.365 & 0.878 & 0.558 & 0.125 & -0.477 & -0.561 & 0.738 & 0.075\\
         Hall2019 & \cite{HallKCBCM19} & Casamino & Mutation & $2^{5}=32$ & 0.062 & 0.303 & 0.774 & 0.309 & 0.125 & -0.735 & -0.493 & 0.625 & 0.050\\
         Hall2019 & \cite{HallKCBCM19} & Glucose & Mutation & $2^{5}=32$ & 0.031 & 0.421 & 0.904 & 0.162 & 0.050 & -0.775 & -0.843 & 0.850 & 0.050\\
         Hall2019 & \cite{HallKCBCM19} & Milk & Mutation & $2^{5}=32$ & 0.062 & 0.392 & 0.876 & 0.276 & 0.050 & 0.089 & -0.625 & 0.775 & 0.062\\
         Hall2019 & \cite{HallKCBCM19} & NAG & Mutation & $2^{5}=32$ & 0.062 & 0.411 & 0.911 & 0.420 & 0.075 & -0.742 & -0.601 & 0.713 & 0.100\\
         Hall2019 & \cite{HallKCBCM19} & Rhamnose & Mutation & $2^{5}=32$ & 0.062 & 0.329 & 0.858 & 0.461 & 0.087 & -0.303 & -0.779 & 0.800 & 0.037\\
         Hall2019 & \cite{HallKCBCM19} & Trypsin & Mutation & $2^{5}=32$ & 0.094 & 0.261 & 0.753 & 0.432 & 0.163 & -0.785 & -0.640 & 0.650 & 0.113\\
         Whitlock2000 & \cite{WhitlockB00} & - & Mutation & $2^5=32$ & 0.094 & 0.267 & 0.734 & 0.391 & 0.138 & -0.897 & -0.739 & 0.750 & 0.000\\
         deVisser2009 & \cite{deVisserPK09} & - & Mutation & $2^5=32$ & 0.156 & 0.156 & 0.416 & -0.401 & 0.300 & -0.617 & -0.568 & 0.388 & 0.200\\
         daSilva2010 & \cite{daSilvaCNPM10} & CCR5 & Mutation & $2^5=32$ & 0.094 & 0.240 & 0.680 & 0.336 & 0.200 & -0.615 & -0.569 & 0.636 & 0.013\\
         daSilva2010 & \cite{daSilvaCNPM10} & CXCR5 & Mutation & $2^5=32$ & 0.094 & 0.137 & 0.469 & 0.215 & 0.197 & -0.476 & 0.112 & 0.544 & 0.000\\
         Sunden2015 & \cite{SundenPSRH15} & AP & Mutation & $2^5=32$ & 0.031 & 0.456 & 0.960 & 0.529 & 0.037 & -0.242 & -0.714 & 0.863 & 0.000\\
         Anderson2021 & \cite{AndersonBYT21} & MPH-CaPTM & Mutation & $2^{5}=32$ & 0.036 & 0.269 & 0.844 & 0.710 & 0.038 & -0.511 & -0.507 & 0.810 & 0.016\\
         Anderson2021 & \cite{AndersonBYT21} & MPH-CdPTM & Mutation & $2^{5}=32$ & 0.031 & 0.407 & 0.905 & 0.538 & 0.075 & -0.464 & -0.470 & 0.738 & 0.025\\
         Anderson2021 & \cite{AndersonBYT21} & MPH-CoPTM & Mutation & $2^{5}=32$ & 0.031 & 0.373 & 0.887 & 0.494 & 0.025 & -0.273 & -0.445 & 0.887 & 0.000\\
         Anderson2021 & \cite{AndersonBYT21} & MPH-CuPTM & Mutation & $2^{5}=32$ & 0.031 & 0.326 & 0.834 & 0.441 & 0.087 & -0.204 & -0.476 & 0.750 & 0.013\\
         Anderson2021 & \cite{AndersonBYT21} & MPH-MgPTM & Mutation & $2^{5}=32$ & 0.062 & 0.362 & 0.893 & 0.770 & 0.050 & -0.376 & -0.465 & 0.863 & 0.000\\
         Anderson2021 & \cite{AndersonBYT21} & MPH-MnPTM & Mutation & $2^{5}=32$ & 0.062 & 0.411 & 0.931 & 0.831 & 0.037 & -0.548 & -0.440 & 0.863 & 0.025\\
         Mira2015 & \cite{MiraCGMSB15} & TEM-AMP & Mutation & $2^4=16$ & 0.688 & -0.672 & -0.723 & -0.222 & 1.000 & -1.000 & -0.368 & 0.067 & 0.200\\
         Mira2015 & \cite{MiraCGMSB15} & TEM-AM & Mutation & $2^4=16$ & 0.125 & -0.008 & 0.382 & 0.351 & 0.083 & -0.989 & -0.169 & 0.594 & 0.719\\
         Mira2015 & \cite{MiraCGMSB15} & TEM-CEC & Mutation & $2^4=16$ & 0.188 & 0.065 & 0.407 & 0.259 & 0.333 & -0.848 & 0.038 & 0.406 & 0.000\\
         Mira2015 & \cite{MiraCGMSB15} & TEM-CTX & Mutation & $2^4=16$ & 0.250 & 0.002 & 0.271 & -0.260 & 0.375 & -0.544 & -0.184 & 0.312 & 0.000\\
         Mira2015 & \cite{MiraCGMSB15} & TEM-ZOX & Mutation & $2^4=16$ & 0.125 & 0.015 & 0.313 & -0.401 & 0.333 & -0.715 & -0.614 & 0.344 & 0.344\\
         Mira2015 & \cite{MiraCGMSB15} & TEM-CXM & Mutation & $2^4=16$ & 0.125 & 0.090 & 0.537 & 0.304 & 0.167 & -0.721 & -0.683 & 0.594 & 0.000\\
         Mira2015 & \cite{MiraCGMSB15} & TEM-CRO & Mutation & $2^4=16$ & 0.250 & -0.015 & 0.192 & -0.306 & 0.250 & -0.599 & -0.161 & 0.406 & 0.031\\
         Mira2015 & \cite{MiraCGMSB15} & TEM-AMC & Mutation & $2^4=16$ & 0.875 & -0.938 & -1.000 & NaN & 0.000 & NaN & -0.214 & 0.000 & 0.000\\
         Mira2015 & \cite{MiraCGMSB15} & TEM-CAZ & Mutation & $2^4=16$ & 0.688 & -0.743 & -0.953 & NaN & 1.000 & 0.408 & -0.160 & 0.000 & 0.143\\
         Mira2015 & \cite{MiraCGMSB15} & TEM-CTT & Mutation & $2^4=16$ & 0.312 & -0.142 & -0.212 & 0.228 & 0.375 & -0.949 & -0.107 & 0.250 & 0.000\\
         Mira2015 & \cite{MiraCGMSB15} & TEM-SAM & Mutation & $2^4=16$ & 0.062 & 0.101 & 0.701 & 0.292 & 0.048 & -0.953 & -0.430 & 0.677 & 0.645\\
         Mira2015 & \cite{MiraCGMSB15} & TEM-CPR & Mutation & $2^4=16$ & 0.188 & 0.018 & 0.259 & -0.242 & 0.292 & -0.875 & -0.268 & 0.375 & 0.000\\
         Mira2015 & \cite{MiraCGMSB15} & TEM-CPD & Mutation & $2^4=16$ & 0.125 & 0.116 & 0.515 & -0.245 & 0.125 & -0.617 & -0.445 & 0.562 & 0.000\\
         Mira2015 & \cite{MiraCGMSB15} & TEM-TZP & Mutation & $2^4=16$ & 0.125 & 0.180 & 0.745 & 0.468 & 0.125 & -0.903 & -0.545 & 0.719 & 0.688\\
         Mira2015 & \cite{MiraCGMSB15} & TEM-FSP & Mutation & $2^4=16$ & 0.250 & -0.202 & -0.437 & -0.241 & 0.458 & -0.963 & -0.499 & 0.219 & 0.000\\
         Meini2015 & \cite{MeiniTWV15} & - & Mutation & $2^4=16$ & 0.062 & 0.209 & 0.798 & 0.501 & 0.000 & -0.682 & -0.627 & 0.741 & 0.000\\
         Lozovsky2009 & \cite{LozovskyCBISKNWH09} & P. falciparum & Mutation & $2^4=16$ & 0.125 & -0.009 & 0.111 & -0.103 & 0.167 & -0.875 & -0.736 & 0.562 & 0.031\\
         Jiang2013 & \cite{JiangKOSMLW18} & P. vivax & Mutation & $2^4=16$ & 0.062 & 0.212 & 0.762 & 0.013 & 0.125 & -0.927 & -0.652 & 0.625 & 0.000\\
         Ogbunugafor2022 & \cite{Ogbunugafor22} & Pyrimethamine & Mutation & $(2^4) \times 12$ & 0.200 & -0.027 & 0.235 & 0.251 & 0.333 & -0.884 & -0.187 & 0.321 & 0.071\\
         Ogbunugafor2022 & \cite{Ogbunugafor22} & Cycloguanil & Mutation & $(2^4) \times 12$ & 0.200 & -0.022 & 0.235 & 0.250 & 0.333 & -0.884 & -0.187 & 0.321 & 0.071\\
         Weinreich20016 & \cite{WeinreichDDH06} & Cefotaxime & Mutation & $2^5=32$ & 0.031 & 0.363 & 0.821 & 0.149 & 0.049 & -0.484 & -0.728 & 0.682 & 0.000\\
         Khan2011 & \cite{KhanDSLC11} & DM25 & Mutation & $2^5=32$ & 0.062 & 0.253 & 0.652 & -0.057 & 0.087 & -0.614 & -0.724 & 0.700 & 0.287\\
         Flynn2013 & \cite{FlynnCMC13} & DM25-EGTA & Mutation & $2^5=32$ & 0.094 & 0.214 & 0.597 & 0.202 & 0.087 & -0.156 & -0.608 & 0.662 & 0.175\\
         Flynn2013 & \cite{FlynnCMC13} & DM25-guanazole & Mutation & $2^5=32$ & 0.094 & 0.212 & 0.597 & 0.202 & 0.087 & -0.156 & -0.608 & 0.662 & 0.175\\
         Chou2011 & \cite{ChouCDSM11} & - & Mutation & $2^4=16$ & 0.062 & 0.310 & 0.993 & 0.570 & 0.000 & -0.742 & -0.836 & 1.000 & 0.000\\
         Malcolm1990 & \cite{MalcolmWMKW90} & Diploid & Mutation & $2^3=8$ & 0.125 & 0.013 & 0.961 & 0.333 & 0.000 & 0.085 & -0.951 & 0.917 & 0.000\\
         Guerrero2019 & \cite{GuerreroSRHO19} & C-muri-GroEL & Mutation & $2^{3}=8$ & 0.250 & -0.062 & 0.571 & 0.333 & 0.167 & -0.992 & -0.676 & 0.583 & 0.750\\
         Guerrero2019 & \cite{GuerreroSRHO19} & C-muri-LON & Mutation & $2^{3}=8$ & 0.250 & -0.227 & -0.150 & -0.333 & 0.333 & -0.976 & -0.025 & 0.333 & 0.500\\
         Guerrero2019 & \cite{GuerreroSRHO19} & C-muri-WT & Mutation & $2^{3}=8$ & 0.250 & -0.373 & -0.864 & -0.333 & 0.333 & -0.733 & -0.150 & 0.250 & 0.083\\
         Guerrero2019 & \cite{GuerreroSRHO19} & E.coli-GroEL & Mutation & $2^{3}=8$ & 0.250 & -0.266 & -0.218 & -0.333 & 0.333 & -0.964 & -0.401 & 0.333 & 0.250\\
         Guerrero2019 & \cite{GuerreroSRHO19} & E.coli-LON & Mutation & $2^{3}=8$ & 0.125 & -0.180 & -0.286 & -0.333 & 0.000 & -0.943 & -0.476 & 0.417 & 0.333\\
         Guerrero2019 & \cite{GuerreroSRHO19} & E.coli-WT & Mutation & $2^{3}=8$ & 0.375 & -0.358 & -0.403 & -1.000 & 0.667 & -0.744 & -0.175 & 0.083 & 0.250\\
         Guerrero2019 & \cite{GuerreroSRHO19} & L-grayi-GroEL & Mutation & $2^{3}=8$ & 0.250 & -0.337 & -0.505 & -0.333 & 0.333 & -0.978 & -0.601 & 0.333 & 0.500\\
         Guerrero2019 & \cite{GuerreroSRHO19} & L-grayi-LON & Mutation & $2^{3}=8$ & 0.250 & -0.263 & -0.110 & 1.000 & 0.333 & -0.999 & -0.200 & 0.333 & 0.167\\
         Guerrero2019 & \cite{GuerreroSRHO19} & L-grayi-WT & Mutation & $2^{3}=8$ & 0.375 & -0.281 & -0.173 & -1.000 & 0.667 & -0.981 & -0.300 & 0.083 & 0.833\\
 
     \end{longtable} 
 
     \endgroup
 
 \end{landscape}

\begin{landscape}
    \centering
    \begingroup
    \small
    \renewcommand{\arraystretch}{1.1}
    \begin{longtable}{llrrrrrrrrr} 

        \caption{Selected landscape features for ProteinGym tasks with mean mutation depth > 1.}
        \label{tab:proteingym} \\

        \hline
        Dataset & Size & Avg. Mutation & $\phi_{\text{lo}}$ & $\rho_a$ & NFC & $I_d$ & $\epsilon\_\text{reci}$ & FDC & $\phi_{\text{EE}}$ & $\eta$\\
        \hline
        \endfirsthead

        \multicolumn{11}{c}%
        {{\bfseries \tablename\ \thetable{} -- continued from previous page}} \\
        \hline
        Dataset & Size & Avg. Mutation & $\phi_{\text{lo}}$ & $\rho_a$ & NFC & $I_d$ & $\epsilon\_\text{reci}$ & FDC & $\phi_{\text{EE}}$ & $\eta$  \\
        \hline
        \endhead

        \hline
        \multicolumn{11}{r}{{\small\textit{Continued on next page}}} \\
        \endfoot

        \hline
        \endlastfoot

        PIN1\_HUMAN\_Tsuboyama\_2023\_1I6C & 802 & 1.145 & 0.050 & -0.047 & 0.806 & 0.157 & -0.664 & -0.330 & 0.116 & 0.022 \\
        RAD\_ANTMA\_Tsuboyama\_2023\_2CJJ & 912 & 1.151 & 0.064 & -0.054 & 0.800 & 0.195 & -0.595 & -0.524 & 0.064 & 0.017 \\
        RCD1\_ARATH\_Tsuboyama\_2023\_5OAO & 1261 & 1.216 & 0.048 & -0.026 & 0.696 & 0.182 & -0.777 & -0.411 & 0.204 & 0.016 \\
        RD23A\_HUMAN\_Tsuboyama\_2023\_1IFY & 1019 & 1.217 & 0.044 & 0.026 & 0.748 & 0.064 & -0.806 & -0.089 & 0.261 & 0.015 \\
        SRBS1\_HUMAN\_Tsuboyama\_2023\_2O2W & 1556 & 1.222 & 0.042 & 0.038 & 0.849 & 0.033 & -0.731 & -0.532 & 0.357 & 0.016 \\
        PSAE\_PICP2\_Tsuboyama\_2023\_1PSE & 1579 & 1.228 & 0.042 & 0.028 & 0.830 & 0.153 & -0.514 & 0.515 & 0.303 & 0.022 \\
        RPC1\_BP434\_Tsuboyama\_2023\_1R69 & 1459 & 1.230 & 0.045 & 0.010 & 0.823 & 0.091 & -0.697 & -0.495 & 0.316 & 0.011 \\
        RL20\_AQUAE\_Tsuboyama\_2023\_1GYZ & 1461 & 1.233 & 0.041 & 0.036 & 0.907 & 0.040 & -0.871 & -0.680 & 0.347 & 0.022 \\
        TNKS2\_HUMAN\_Tsuboyama\_2023\_5JRT & 1479 & 1.244 & 0.041 & 0.050 & 0.778 & 0.049 & -0.783 & -0.226 & 0.371 & 0.015 \\
        UBR5\_HUMAN\_Tsuboyama\_2023\_1I2T & 1453 & 1.247 & 0.039 & 0.051 & 0.786 & 0.053 & -0.789 & -0.374 & 0.367 & 0.018 \\
        NUSG\_MYCTU\_Tsuboyama\_2023\_2MI6 & 1380 & 1.262 & 0.040 & 0.063 & 0.773 & 0.045 & -0.768 & -0.157 & 0.409 & 0.015 \\
        RBP1\_HUMAN\_Tsuboyama\_2023\_2KWH & 1332 & 1.268 & 0.041 & 0.052 & 0.725 & 0.071 & -0.697 & -0.200 & 0.395 & 0.015 \\
        RFAH\_ECOLI\_Tsuboyama\_2023\_2LCL & 1326 & 1.269 & 0.041 & 0.045 & 0.650 & 0.102 & -0.351 & 0.155 & 0.380 & 0.017 \\
        SPG2\_STRSG\_Tsuboyama\_2023\_5UBS & 1451 & 1.291 & 0.040 & 0.049 & 0.823 & 0.042 & -0.589 & -0.566 & 0.404 & 0.013 \\
        CATR\_CHLRE\_Tsuboyama\_2023\_2AMI & 1903 & 1.296 & 0.040 & 0.019 & 0.814 & 0.034 & -0.765 & -0.141 & 0.324 & 0.012 \\
        SAV1\_MOUSE\_Tsuboyama\_2023\_2YSB & 965 & 1.296 & 0.048 & -0.016 & 0.768 & 0.212 & -0.659 & -0.223 & 0.326 & 0.023 \\
        CBPA2\_HUMAN\_Tsuboyama\_2023\_1O6X & 2068 & 1.344 & 0.036 & 0.090 & 0.890 & 0.043 & -0.660 & -0.681 & 0.455 & 0.015 \\
        FECA\_ECOLI\_Tsuboyama\_2023\_2D1U & 1886 & 1.354 & 0.038 & 0.092 & 0.743 & 0.063 & -0.476 & -0.202 & 0.478 & 0.012 \\
        NUSA\_ECOLI\_Tsuboyama\_2023\_1WCL & 2028 & 1.356 & 0.035 & 0.104 & 0.863 & 0.041 & -0.474 & -0.541 & 0.497 & 0.018 \\
        EPHB2\_HUMAN\_Tsuboyama\_2023\_1F0M & 1960 & 1.368 & 0.035 & 0.086 & 0.894 & 0.064 & -0.600 & -0.651 & 0.485 & 0.012 \\
        CUE1\_YEAST\_Tsuboyama\_2023\_2MYX & 1580 & 1.396 & 0.035 & 0.088 & 0.782 & 0.089 & -0.641 & 0.118 & 0.478 & 0.013 \\
        ODP2\_GEOSE\_Tsuboyama\_2023\_1W4G & 1134 & 1.410 & 0.043 & 0.096 & 0.782 & 0.084 & -0.808 & 0.227 & 0.461 & 0.022 \\
        TCRG1\_MOUSE\_Tsuboyama\_2023\_1E0L & 1058 & 1.413 & 0.036 & 0.069 & 0.879 & 0.155 & -0.487 & -0.646 & 0.441 & 0.021 \\
        PR40A\_HUMAN\_Tsuboyama\_2023\_1UZC & 2033 & 1.428 & 0.032 & 0.132 & 0.904 & 0.142 & -0.527 & -0.713 & 0.443 & 0.021 \\
        BCHB\_CHLTE\_Tsuboyama\_2023\_2KRU & 1572 & 1.434 & 0.033 & 0.120 & 0.794 & 0.038 & -0.821 & -0.395 & 0.586 & 0.012 \\
        SR43C\_ARATH\_Tsuboyama\_2023\_2N88 & 1583 & 1.438 & 0.031 & 0.127 & 0.865 & 0.037 & -0.600 & -0.522 & 0.572 & 0.011 \\
        MBD11\_ARATH\_Tsuboyama\_2023\_6ACV & 2116 & 1.454 & 0.031 & 0.121 & 0.913 & 0.079 & -0.517 & -0.646 & 0.518 & 0.017 \\
        DNJA1\_HUMAN\_Tsuboyama\_2023\_2LO1 & 2264 & 1.463 & 0.029 & 0.149 & 0.916 & 0.057 & -0.508 & -0.714 & 0.571 & 0.015 \\
        MAFG\_MOUSE\_Tsuboyama\_2023\_1K1V & 1429 & 1.467 & 0.030 & 0.102 & 0.838 & 0.100 & -0.558 & -0.593 & 0.536 & 0.019 \\
        RCRO\_LAMBD\_Tsuboyama\_2023\_1ORC & 2278 & 1.475 & 0.028 & 0.150 & 0.876 & 0.054 & -0.608 & -0.481 & 0.586 & 0.019 \\
        BBC1\_YEAST\_Tsuboyama\_2023\_1TG0 & 2069 & 1.476 & 0.031 & 0.127 & 0.779 & 0.070 & -0.510 & -0.336 & 0.563 & 0.010 \\
        PITX2\_HUMAN\_Tsuboyama\_2023\_2L7M & 1824 & 1.486 & 0.033 & 0.121 & 0.829 & 0.089 & -0.654 & -0.397 & 0.529 & 0.013 \\
        THO1\_YEAST\_Tsuboyama\_2023\_2WQG & 1279 & 1.487 & 0.033 & 0.109 & 0.828 & 0.075 & -0.513 & -0.551 & 0.583 & 0.012 \\
        SPA\_STAAU\_Tsuboyama\_2023\_1LP1 & 2105 & 1.508 & 0.025 & 0.149 & 0.834 & 0.129 & -0.343 & -0.471 & 0.546 & 0.024 \\
        YAIA\_ECOLI\_Tsuboyama\_2023\_2KVT & 1890 & 1.509 & 0.026 & 0.143 & 0.847 & 0.059 & -0.739 & -0.589 & 0.569 & 0.010 \\
        ISDH\_STAAW\_Tsuboyama\_2023\_2LHR & 1944 & 1.516 & 0.030 & 0.153 & 0.772 & 0.053 & -0.791 & -0.292 & 0.614 & 0.014 \\
        VILI\_CHICK\_Tsuboyama\_2023\_1YU5 & 2568 & 1.532 & 0.028 & 0.222 & 0.904 & 0.047 & -0.639 & -0.621 & 0.616 & 0.017 \\
        NKX31\_HUMAN\_Tsuboyama\_2023\_2L9R & 2482 & 1.537 & 0.029 & 0.153 & 0.826 & 0.145 & -0.566 & -0.506 & 0.541 & 0.025 \\
        DOCK1\_MOUSE\_Tsuboyama\_2023\_2M0Y & 2915 & 1.584 & 0.027 & 0.149 & 0.802 & 0.143 & -0.397 & -0.333 & 0.549 & 0.017 \\
        CSN4\_MOUSE\_Tsuboyama\_2023\_1UFM & 3295 & 1.589 & 0.022 & 0.234 & 0.833 & 0.027 & -0.644 & -0.398 & 0.673 & 0.014 \\
        CBX4\_HUMAN\_Tsuboyama\_2023\_2K28 & 2282 & 1.598 & 0.021 & 0.164 & 0.858 & 0.084 & -0.633 & -0.456 & 0.592 & 0.012 \\
        OBSCN\_HUMAN\_Tsuboyama\_2023\_1V1C & 3197 & 1.621 & 0.023 & 0.194 & 0.891 & 0.070 & -0.511 & -0.561 & 0.654 & 0.013 \\
        SPTN1\_CHICK\_Tsuboyama\_2023\_1TUD & 3201 & 1.672 & 0.018 & 0.286 & 0.868 & 0.035 & -0.583 & -0.482 & 0.698 & 0.011 \\
        YNZC\_BACSU\_Tsuboyama\_2023\_2JVD & 2300 & 1.690 & 0.017 & 0.180 & 0.900 & 0.093 & -0.442 & -0.632 & 0.670 & 0.017 \\
        UBE4B\_HUMAN\_Tsuboyama\_2023\_3L1X & 3622 & 1.691 & 0.022 & 0.177 & 0.795 & 0.073 & -0.578 & -0.170 & 0.646 & 0.012 \\
        SDA\_BACSU\_Tsuboyama\_2023\_1PV0 & 2770 & 1.699 & 0.016 & 0.266 & 0.922 & 0.045 & -0.496 & -0.658 & 0.717 & 0.015 \\
        MYO3\_YEAST\_Tsuboyama\_2023\_2BTT & 3297 & 1.713 & 0.019 & 0.212 & 0.825 & 0.074 & -0.374 & -0.405 & 0.709 & 0.011 \\
        AMFR\_HUMAN\_Tsuboyama\_2023\_4G3O & 2972 & 1.724 & 0.021 & 0.264 & 0.848 & 0.081 & -0.665 & -0.414 & 0.682 & 0.010 \\
        HECD1\_HUMAN\_Tsuboyama\_2023\_3DKM & 5586 & 1.777 & 0.015 & 0.288 & 0.898 & 0.084 & -0.427 & -0.455 & 0.680 & 0.018 \\
        POLG\_PESV\_Tsuboyama\_2023\_2MXD & 5130 & 1.806 & 0.011 & 0.352 & 0.905 & 0.044 & -0.491 & -0.488 & 0.775 & 0.016 \\
        DLG4\_HUMAN\_Faure\_2021 & 6976 & 1.817 & 0.040 & 0.441 & 0.794 & 0.127 & -0.637 & -0.258 & 0.431 & 0.027 \\
        RASK\_HUMAN\_Weng\_2022\_binding-DARPin\_K55 & 24873 & 1.876 & 0.023 & 0.646 & 0.903 & 0.061 & -0.633 & -0.299 & 0.460 & 0.054 \\
        RASK\_HUMAN\_Weng\_2022\_abundance & 26012 & 1.882 & 0.027 & 0.487 & 0.774 & 0.107 & -0.558 & -0.123 & 0.441 & 0.028 \\
        A4\_HUMAN\_Seuma\_2022 & 14811 & 1.946 & 0.032 & 0.365 & 0.785 & 0.106 & -0.710 & -0.206 & 0.562 & 0.005 \\
        YAP1\_HUMAN\_Araya\_2012 & 10075 & 1.964 & 0.094 & 0.264 & 0.639 & 0.127 & 0.245 & -0.242 & 0.598 & 0.017 \\
        PABP\_YEAST\_Melamed\_2013 & 37708 & 1.969 & 0.038 & 0.446 & 0.856 & 0.106 & -0.441 & -0.145 & 0.631 & 0.059 \\
        GRB2\_HUMAN\_Faure\_2021 & 63366 & 1.984 & 0.021 & 0.450 & 0.841 & 0.089 & -0.354 & -0.179 & 0.660 & 0.029 \\
        Q6WV12\_9MAXI\_Somermeyer\_2022 & 31401 & 2.685 & 0.459 & 0.579 & 0.853 & 0.086 & -0.727 & -0.298 & 0.444 & 0.000 \\
        D7PM05\_CLYGR\_Somermeyer\_2022 & 24515 & 3.038 & 0.547 & 0.437 & 0.705 & 0.058 & -0.778 & -0.476 & 0.350 & 0.002 \\
        Q8WTC7\_9CNID\_Somermeyer\_2022 & 33510 & 3.055 & 0.499 & 0.540 & 0.818 & 0.098 & -0.745 & -0.241 & 0.458 & 0.000 \\
        F7YBW8\_MESOW\_Aakre\_2015 & 9192 & 3.575 & 0.008 & -0.214 & 0.925 & 0.241 & 0.376 & -0.374 & 0.609 & 0.581 \\
        GFP\_AEQVI\_Sarkisyan\_2016 & 51714 & 3.878 & 0.724 & 0.456 & 0.741 & 0.198 & -0.879 & -0.547 & 0.407 & 0.058 \\
        CAPSD\_AAV2S\_Sinai\_2021 & 42328 & 4.728 & 0.439 & 0.269 & 0.862 & 0.189 & -0.422 & -0.223 & 0.539 & 0.005 \\
        F7YBW8\_MESOW\_Ding\_2023 & 7922 & 5.426 & 0.783 & -0.353 & 0.714 & 0.203 & -0.204 & -0.683 & 0.320 & 0.051 \\
        GCN4\_YEAST\_Staller\_2018 & 2638 & 17.068 & 0.966 & -0.292 & -0.083 & 0.295 & -0.839 & -0.146 & 0.304 & 0.031 \\

    \end{longtable}
    \endgroup
\end{landscape}

\begin{landscape}
    \centering
    \begingroup
    \small
    \renewcommand{\arraystretch}{1.1}
    \begin{longtable}{llrrrrrrrrr} 

        \caption{Selected landscape features for RNAGym tasks with mean mutation depth > 1.}
        \label{tab:rnagym} \\

        \hline
        Dataset & RNA Type & Size & $\phi_{\text{lo}}$ (\%) & $\rho_a$ (\%) & NFC & $I_d$ & $\epsilon_\text{reci}$ & FDC & $\phi_{\text{EE}}$ & $\eta$\\
        \hline
        \endfirsthead

        \multicolumn{11}{c}%
        {{\bfseries \tablename\ \thetable{} -- continued from previous page}} \\
        \hline
        Dataset & RNA Type & Size & $\phi_{\text{lo}}$ (\%) & $\rho_a$ (\%) & NFC & $I_d$ & $\epsilon_\text{reci}$ & FDC & $\phi_{\text{EE}}$ & $\eta$  \\
        \hline
        \endhead

        \hline
        \multicolumn{11}{r}{{\small\textit{Continued on next page}}} \\
        \endfoot

        \hline
        \endlastfoot

        Andreasson2020 & ribozyme & 7343 & 0.078 & 0.080 & 0.404 & 2.748 & 0.043 & -0.243 & 0.515 & 0.766 \\
        Beck2022 & ribozyme & 21321 & 0.022 & 0.498 & 0.917 & 0.335 & 0.123 & 0.081 & 0.522 & 0.092 \\
        Domingo2018 & tRNA & 4175 & 0.021 & 0.127 & 0.777 & 0.628 & 0.182 & -0.506 & 0.681 & 0.076 \\
        Guy2014 & tRNA & 25491 & 0.393 & 0.106 & 0.484 & 2.434 & 0.191 & -0.377 & 0.571 & 0.677 \\
        Janzen2022 & ribozyme & 1953 & 0.031 & 0.129 & 0.742 & 0.770 & 0.097 & -0.285 & 0.627 & 0.004 \\
        Janzen2022 & ribozyme & 1953 & 0.048 & 0.182 & 0.823 & 0.666 & 0.065 & -0.250 & 0.644 & 0.003 \\
        Janzen2022 & ribozyme & 1953 & 0.041 & 0.176 & 0.516 & 1.434 & 0.087 & -0.017 & 0.617 & 0.001 \\
        Janzen2022 & ribozyme & 1953 & 0.032 & 0.177 & 0.529 & 1.107 & 0.075 & -0.186 & 0.625 & 0.001 \\
        Janzen2022 & ribozyme & 1953 & 0.049 & 0.170 & 0.695 & 0.915 & 0.100 & -0.226 & 0.609 & 0.002 \\
        Ke2017 & mRNA & 5533 & 0.078 & 0.292 & 0.906 & 0.394 & 0.150 & -0.485 & 0.634 & 0.129 \\
        Kobori2015 & ribozyme & 255 & 0.008 & -0.174 & 0.881 & 0.550 & 0.089 & -0.317 & 0.800 & 0.135 \\
        Kobori2015 & ribozyme & 255 & 0.020 & -0.194 & 0.805 & 0.582 & 0.117 & -0.523 & 0.778 & 0.249 \\
        Kobori2015 & ribozyme & 1023 & 0.041 & -0.239 & 0.407 & 0.895 & 0.339 & -0.148 & 0.422 & 0.957 \\
        Kobori2016 & ribozyme & 10296 & 0.009 & 0.284 & 0.842 & 0.505 & 0.031 & -0.003 & 0.519 & 0.053 \\
        Kobori2018 & ribozyme & 16383 & 0.016 & -0.025 & 0.722 & 0.767 & 0.227 & 0.027 & 0.671 & 0.080 \\
        Li2016 & tRNA & 65536 & 0.356 & 0.051 & 0.357 & 0.707 & 0.095 & -0.578 & 0.411 & 0.033 \\
        McRae2024 & ribozyme & 74942 & 0.758 & 0.195 & 0.754 & 0.512 & 0.055 & -0.510 & 0.427 & 0.006 \\
        McRae2024 & ribozyme & 47503 & 0.439 & 0.363 & 0.750 & 0.304 & 0.107 & -0.387 & 0.499 & 0.006 \\
        Peri2022 & ribozyme & 16383 & 0.001 & 0.113 & 0.955 & 0.472 & 0.066 & -0.387 & 0.799 & 0.763 \\
        Roberts2023 & ribozyme & 33930 & 0.042 & 0.521 & 0.884 & 0.414 & 0.149 & 0.038 & 0.512 & 0.041 \\
        Roberts2023 & ribozyme & 21321 & 0.022 & 0.495 & 0.919 & 0.332 & 0.087 & -0.242 & 0.521 & 0.077 \\
        Roberts2023 & ribozyme & 9045 & 0.035 & 0.391 & 0.852 & 0.455 & 0.122 & -0.051 & 0.501 & 0.054 \\
        Roberts2023 & ribozyme & 22578 & 0.033 & 0.483 & 0.906 & 0.440 & 0.172 & -0.172 & 0.520 & 0.065 \\
        Roberts2023 & ribozyme & 10296 & 0.016 & 0.344 & 0.876 & 0.436 & 0.079 & -0.184 & 0.522 & 0.072 \\
        Soo2021 & ribozyme & 63430 & 0.021 & 0.039 & 0.752 & 0.725 & 0.238 & -0.015 & 0.641 & 0.008 \\
        Tome2014 & aptamer & 417 & 0.317 & 0.011 & 0.421 & 0.060 & 0.500 & 0.294 & 0.402 & 0.005 \\
        Tome2014 & aptamer & 2652 & 0.049 & 0.392 & 0.831 & 0.236 & 0.078 & -0.122 & 0.362 & 0.018 \\
        Zhang2020 & ribozyme & 111417 & 0.464 & 0.408 & 0.707 & 0.567 & 0.134 & -0.441 & 0.546 & 0.009 \\
        Zhang2024 & ribozyme & 61393 & 0.492 & 0.461 & 0.656 & 0.612 & 0.239 & -0.284 & 0.530 & 0.026 \\
        Zhang2024 & ribozyme & 69583 & 0.452 & 0.488 & 0.654 & 0.605 & 0.175 & -0.219 & 0.538 & 0.012 \\
        Zhang2024 & ribozyme & 149710 & 0.450 & 0.222 & 0.387 & 0.885 & 0.143 & -0.268 & 0.539 & 0.066 \\

    \end{longtable}
    \endgroup
\end{landscape}

\end{document}